\documentclass[journal=jctcce,manuscript=article]{achemso}

\usepackage[version=3]{mhchem} 
\usepackage{amsmath, amssymb, amsthm}
\usepackage{amsfonts}
\usepackage{braket}
\usepackage{color, colortbl}

\newtheorem{remark}{Remark}

\author{Mingyuan Zhang}
\affiliation[Zhejiang University]
{College of Life Sciences, Zhejiang University, Hangzhou 310027, China}
\author{Zhicheng Zhang}
\affiliation[Tongji University]
{School of Mathematical Sciences, Tongji University, Shanghai 200092, China}
\author{Hao Wu}
\affiliation[Shanghai Jiao Tong University]
{School of Mathematical Sciences, Institute of Natural Sciences, and MOE-LSC, Shanghai Jiao Tong University, Shanghai 200240, China}
\email{hwu81@sjtu.edu.cn}
\author{Yong Wang}
\affiliation[Zhejiang University]
{College of Life Sciences, Zhejiang University, Hangzhou 310027, China}
\email{yongwang_isb@zju.edu.cn}
\title[]{Flow Matching for Optimal Reaction Coordinates of Biomolecular System}

\begin{document}

\begin{tocentry}

\includegraphics[scale=0.24]{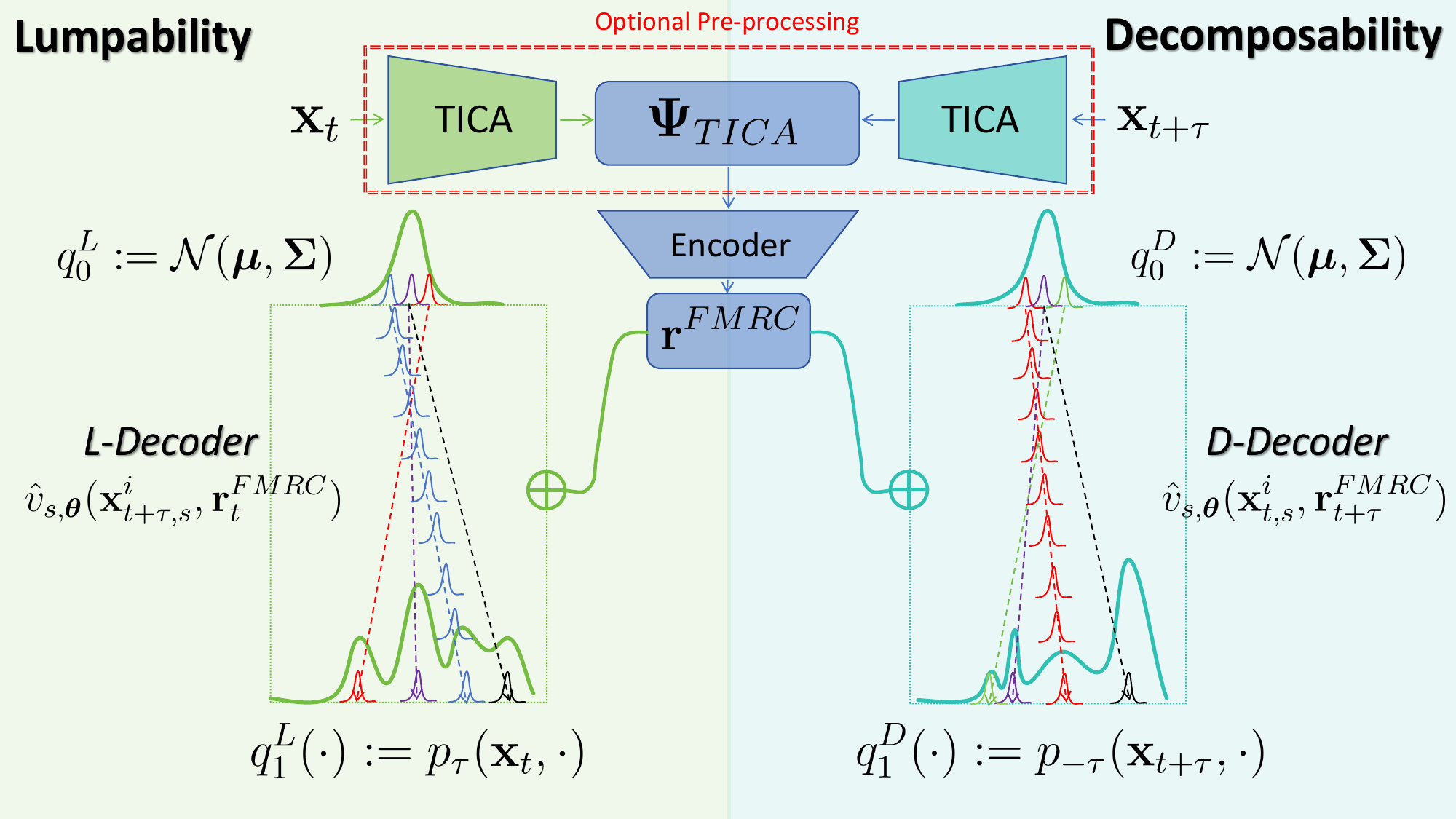}

\end{tocentry}

\begin{abstract}
We present Flow Matching for Reaction Coordinates (FMRC), a novel deep learning algorithm designed to identify optimal reaction coordinates (RC) in biomolecular reversible dynamics. FMRC is based on the mathematical principles of lumpability and decomposability, which we reformulate into a conditional probability framework for efficient data-driven optimization using deep generative models. While FMRC does not explicitly learn the well-established transfer operator or its eigenfunctions, it can effectively encode the dynamics of leading eigenfunctions of the system transfer operator into its low-dimensional RC space. We further quantitatively compare its performance with several state-of-the-art algorithms by evaluating the quality of Markov State Models (MSM) constructed in their respective RC spaces, demonstrating the superiority of FMRC in three increasingly complex biomolecular systems. In addition, we successfully demonstrated the efficacy of FMRC for bias deposition in enhanced sampling of a simple model system. Finally, we discuss its potential applications in downstream applications such as enhanced sampling methods and MSM construction.
\end{abstract}

\section{Introduction}
Reaction coordinate (RC) is the core object of biophysical studies using molecular dynamics (MD) simulation.  On the one hand, with the technical advancement in computational hardware \cite{7012191}, the MD trajectories of molecular systems with increasing length and complexity present a significant challenge for their biophysical interpretation due to their high dimensionality. By projecting these trajectories into a low-dimensional manifold parameterized by an RC, one can remarkably reduce the effort in understanding the underlying biophysical mechanism. For example, in the well-established Markov State Model (MSM) approach \cite{prinz2011markov}, the dimension reduction of the raw trajectory data by optimal RCs is essential for the assignment of different conformations into the ``microstates'' which assume local equilibrium. On the other hand, many interesting biological phenomena at a timescale of seconds or even hours, are still out of the reach of current MD simulations. To accelerate MD simulation in such a scenario, enhanced sampling techniques, such as umbrella sampling\cite{torrie1977nonphysical} or metadynamics\cite{laio2002escaping}, which biased the simulation along a given RC, are often utilized. However, in both scenarios, the efficacy of the strategy is strongly dependent on the ``optimality'' of the RC. Intuitively, optimal RC is a parameterization of a low-dimensional projection of the original system which encodes the full-dimensional system dynamics. To do so, the RC should distinguish between metastable states by assigning similar RC coordinates to kinetically close conformations. It should also accurately characterize the transition pathways and their associated transition states within limited dimensions \cite{bussi2020using}. However, the determination of such RC, in general, cannot be accurately determined based on human expertise in increasingly complex biomolecular systems. 

As a result, machine learning techniques have been gaining popularity in the determination of RC in a data-driven manner according to an optimization target, or loss function, based on a specific definition of RC in mathematical language. However, such a precise mathematical definition of optimal RC mutually agreed by the community has not yet been established \cite{wang2019past,chen2019nonlinear,perez2013identification,mcgibbon2017identification,kang2024computing,lazzeri2023molecular,bittracher2023optimal}. The earliest attempts at such a definition are based on the splitting probability, also known as the committor function \cite{onsager1938initial,du1998transition}. Given two user-defined non-overlapping metastable states, labelled A and B (e.g. folded and unfolded state of a protein), the committor function $q_A(\mathbf x_0) := 1 - q_B(\mathbf x_0)$, which defined as the probability of reaching state A before state B given the starting conformation $\mathbf x_0$, is argued as the optimal RC for the transition between A and B. Utilizing such definition, multiple machine learning algorithms\cite{kang2024computing,lazzeri2023molecular,jung2023machine} have been proposed to compute optimal RC as the committor function with success in torsional isomerization, protein folding, and protein aggregation. However, a major problem of such a definition is that it relies on the human definition of states A and B, and sometimes such prior information might not be available or reliably provided. 

Such an issue further motivates the development of unsupervised RC machine learning algorithms based on a loss function that can be purely computed from simulation data without any state labelling. One notable group of such unsupervised algorithms constructs the optimal RC under the mathematical framework provided by the variational approach to conformational dynamics (VAC) \cite{noe2013variational} and its irreversible extension, the variational approach for Markov processes (VAMP) \cite{wu2020variational}. Here, we focus solely on reversible dynamics fulfilling the detailed balance condition within the scope of VAC. Under the VAC viewpoint, the optimal RC is defined as the leading eigenfunctions of the system transfer operator (see Theory and Methods). Various VAC-based RC machine learning algorithms such as time-lagged independent component analysis (TICA)\cite{perez2013identification} and its variants Koopman-reweighted TICA\cite{wu2017variational}, kernel TICA\cite{schwantes2015modeling}, state-free reversible VAMPnets (SRV)\cite{chen2019nonlinear}, etc. have been proposed to numerically approximate these eigenfunctions from data by seeking a linear combination of the basis function which maximizes autocorrelation. These algorithms have achieved great success in a wide range of biomolecular systems including protein folding\cite{mckiernan2017modeling,sidky2019high,bonati2021deep}, protein-protein interaction \cite{plattner2017complete}, ligand binding \cite{plattner2015protein,brotzakis2018accelerating}, etc., and have been widely accepted as a routine algorithm in the pipeline of Markov State Model (MSM) construction for their generality and robustness. Nevertheless, there are several pitfalls associated with such algorithms\cite{bittracher2018data,bittracher2023optimal}: 

\begin{itemize}
	\item The number of dimensions of the leading eigenfunctions can sometimes be redundant, which makes them harder to interpret than the committor function and less applicable to enhanced sampling purposes. 
	\item The accurate approximation of eigenfunctions is expensive since it requires extensive simulation data covering the whole conformational space. 
	\item The linear regression nature of these algorithms may limit the expressitivity of the learned RC.
\end{itemize}

To address these pitfalls, Bittracher et al.\cite{bittracher2023optimal} have recently proposed a new theoretical framework for the unsupervised learning of optimal RC based on their previously developed transition manifold theory\cite{bittracher2018transition,bittracher2018data}. Instead of approximating the leading eigenfunctions directly, they aimed to construct a low-dimensional RC space projected from the original full dimensional conformational space which preserves the major system dynamics described by the original system transfer operator. The parameterization of the basis functions of this projected space can then be viewed as the optimal RC. They further formulated mathematical definitions of such optimal RC, known as lumpability and decomposability. We found these formulations particularly attractive, as they have been theoretically well-established and numerically demonstrated on simple model systems that they can be computed in theory with less sampling required in comparison to the VAC-based algorithms. Furthermore, since these formulations only involve the modelling of original and projected transition probability, they are ideal optimization targets for deep generative modelling techniques with strong expressivity and robustness. However, they have not yet proposed or implemented a ready-to-use deep learning algorithm based on such formulations that can be directly applied to biomolecular MD trajectories. As a result, the efficacy of such formulations in computing RC for real-life biomolecular systems has not been investigated. 

On the other hand, Tiwary et al.\cite{wang2019past} introduced an alternative perspective on optimal RC for unsupervised learning based on information theory \cite{tishby2000information}. They argued that an optimal RC should be a low-dimensional representation which preserves minimal information of the system conformation at the current time step, while still being able to accurately predict the future system conformation after a short lag time. The predictability of such an RC is mathematically quantified by the mutual information between the current RC representation and the future full-dimensional conformation. Building on this concept, they developed an RC deep learning algorithm called the state predictive information bottleneck (SPIB) \cite{wang2021state}, which has been successfully applied to studying biophysical processes such as protein folding and membrane permeation of small molecules \cite{mehdi2022accelerating}. They also demonstrated the potential of their algorithm in improving other sampling or modelling techniques, such as enhanced sampling \cite{mehdi2022accelerating}, MSM construction \cite{wang2024information}, and the weighted ensemble method \cite{wang2024augmenting}. Moreover, methods have been proposed to further integrate the information bottleneck principle with generative models, including variational autoencoders \cite{wang2024latent} and normalizing flows \cite{wu2024reaction}, enabling the construction of interpretable and simplified models of reduced kinetics in the RC space.

In this article, we 1) present a novel unsupervised deep learning algorithm for determining the optimal RC of biomolecular reversible dynamics, based on principles of lumpability and decomposability. We refer to this algorithm as Flow Matching for Reaction Coordinates (FMRC), following our recent mathematical analysis of the algorithm \cite{zhang2024flow}. We then 2) perform a systematic comparison of the capability of learning optimal RC of different algorithms discussed above with FMRC. Finally, we 3) performed a preliminary test on the capability of FMRC in learning RC for bias deposition in enhanced sampling simulation.

For FMRC, we first reformulate the concepts of lumpability and decomposability to an equivalent conditional probability formulation that aligns with modern deep generative learning architectures. To further streamline the training process and improve modelling accuracy, we constructed our RC learning model as a continuous normalizing flow \cite{chen2018neural} conditioned (decoder) on a simple feed-forward neural network (encoder) which represents the optimal RC. We simplified the training process further by using a simulation-free training method as in flow matching \cite{lipman2022flow,tong2023improving}. Such implementation leads to an easy-to-train, accurate RC deep learning algorithm with low variance, demonstrated across three increasingly complex protein folding systems, with minimal reliance on prior knowledge of the system.

We then evaluate the quality of the RCs generated by FMRC trained on these datasets by comparing them with those produced by a range of popular unsupervised algorithms. This evaluation is based on each algorithm's ability to encode the major system dynamics described by the original system transfer operator in their low-dimensional RC space. This is achieved by comparing the eigenvalues of the MSM constructed inside each RC space. To take account of the stochastic elements presented in these algorithms, we repeated the training process 10 times for each algorithm using consistent hyperparameters for each system, allowing us to assess the variance across different algorithms. 

Our results demonstrate a significantly higher amount of dynamics preservation by the FMRC in the RC space than all other algorithms, along with a smaller training variance than all deep learning-based algorithms for all three systems. Moreover, for all three systems, we trained FMRC with a consistent training hyperparameters setting with only minor adjustments, demonstrating its robustness in training. Additionally, by tuning the lag time, we demonstrated that the free energy surface (FES) projected on the learned RC spaces can either preserve fine details or provide macroscopic interpretation for all three systems. Finally, using a 3.2 ns unbiased simulation dataset from our previous study \cite{zhang2024enhanced} for FMRC training, we showed that the RC learned by FMRC is able to efficiently drive the transitions between the system metastable states of alanine dipeptide (Ala2) in enhanced sampling simulations and converge the FES estimates. Taken together, these results suggest FMRC is a promising tool for downstream applications such as enhanced sampling methods and MSM construction.

\section{Theory and Methods}

\subsection{Transfer Operator}

MD trajectories can be considered as a Markov process $\{\mathbf x_t\}$ which progressively evolves the conformational probability density $p_t(\mathbf x)$ at time step $t$ towards the system equilibrium distribution $\pi(\mathbf x)$  in the state space $\Omega$. This equilibrium distribution is unique and corresponds to the Boltzmann distribution $\pi(\mathbf x)=Z^{-1}\exp\left(-\beta \mathcal H(\mathbf x)\right)$, where $Z$ is the partition function, $\mathcal H$ is the system Hamiltonian and $\beta$ is the inverse temperature $\beta = 1/(k_BT)$. Such a Markov process has an associated transfer operator $\mathcal T(\tau)$, which propagates the weighted probability density at current time step $t$, $u_t(\mathbf x) = \pi(\mathbf x)^{-1}p_t(\mathbf x)$, to the weighted probability density after a short lag time $\tau$, $u_{t+\tau}(\mathbf x) = \pi(\mathbf x)^{-1}p_{t+\tau}(\mathbf x)$:
\begin{equation}
    u_{t+\tau}(\mathbf{y})=\mathcal{T}(\tau)u_{t}(\mathbf{y})=\frac{1}{\pi(\mathbf{y})}\int_{\Omega}\space p_{\tau}(\mathbf{x},\mathbf{y})\pi(\mathbf{x})u_{t}(\mathbf{x})\mathbf{d}\mathbf{x}
\end{equation}
where $p_\tau(\mathbf x,\mathbf y)$ is the transition probability of a system conformation being $\mathbf x$ at time $t$ given that the system conformation being $\mathbf y$ after a short lag time $\tau$, which due to Markovianity, is only dependent on $\mathbf x$:
\begin{equation}
    p_\tau(\mathbf x,\mathbf y) = \mathbb {P}(\mathbf x_{t+\tau} = \mathbf y\mid\mathbf  x_{t} = \mathbf x)
\end{equation}
Notice that in reversible dynamics considered in this article, $p_\tau$ satisfies the condition of detailed balance:
\begin{equation}
    \pi(\mathbf x)p_\tau(\mathbf x,\mathbf y) =\pi(\mathbf y)p_\tau(\mathbf y,\mathbf x)
\end{equation}
Under this condition, the transfer operator of a system is a self-adjoint operator and thus can undergo spectral decomposition that yields an infinite series of real eigenvalues $\lambda_i$ and orthonormal eigenfunctions $\psi_i  := \psi_i(\mathbf x)$:
\begin{equation}
    \mathcal T(\tau)\psi_i = \lambda_i\psi_i
\end{equation}
where the eigenvalues $\lambda_i$ are usually arranged in descending order, with the largest eigenvalue always being 1:
\begin{equation}
    1 \equiv \lambda_0 > \lambda_1 > \lambda_2 > ...> \lambda_\infty
\end{equation}
The eigenvalues $\lambda_i$ each corresponds to an implied timescale (ITS) $t_i$ of the orthogonal process represented by its associated eigenfunction $\psi_i$:
\begin{equation}
    t_i = -\frac{\tau}{\ln|\lambda_i|} 
\end{equation}
This indicates the first eigenfunction $\psi_0$ has an infinite timescale $t_0\to \infty$. After evolving the dynamics by $\mathcal T(\tau)$ for a very long time, all the other eigenfunctions associated with an eigenvalue $\lambda_i < 1$ will decay to zero, representing the system equilibrium:
\begin{equation}
    \lim_{\tau\to\infty}\mathcal T(\tau)u_t(\mathbf x) = \psi_0 = 1
\end{equation}
The leading eigenfunctions from $\psi_1$ are therefore the slowest modes of the system corresponding the to conformational transition processes between the metastable states of the system. From this viewpoint, if $\mathcal T(\tau)$ or $\psi_i$ is available, one can trivially define the optimal RC as the leading eigenfunctions $\psi_i$ of the system. However, neither $\mathcal T(\tau)$ nor $\psi_i$ is analytically available for complex biomolecular systems. To tackle this issue, one well-established strategy is to utilize VAC to approximate $\mathcal T(\tau)$ or $\psi_i$ in a data-driven manner (see Supporting Information).

\subsection{Lumpability and Decomposability}

While the VAC-based methods have been successfully applied to a wide range of biomolecular systems, as stated in the Introduction, computing RC using VAC-based algorithms has significant pitfalls. To address these issues, Bittracher et al.\cite{bittracher2023optimal} proposed a set of novel mathematical definitions of optimal RC, known as lumpability and decomposability, following their earlier theoretical framework of transition manifold theory \cite{bittracher2018transition,bittracher2018data}. Lumpability states that the optimal RC $\bold r^{LD}$ should satisfy the following relationship for all time-lagged pairs $\{\bold x_t,\bold x_{t+\tau}\}$:
\begin{equation}
p_{\tau}(\bold x,\bold y)\approx\mathbb{P}(\bold x_{t+\tau}=\bold y\mid\bold r^{LD}(\bold x_{t})=\bold r^{LD}(\bold x))
\end{equation}

In words, if the RC is indeed optimal, the transition probability from conformation $\bold x$ to conformation $\bold y$ after a short lag time $\tau$ is only dependent on the low-dimensional RC values of the conformation $\bold r^{LD}(\bold x)$, instead of its full-dimensional conformational coordinates $\bold x$. The name ``lumpability" comes from that the optimal RC can effectively ``lump'' all kinetically close conformational coordinates together to a point in the RC space defined by $\bold r^{LD}(\bold x)$. 

In contrast, decomposability states that the optimal RC $\bold r^{LD}(\bold x)$ should satisfy the following relationship for all time-lagged pairs $\{\bold x_t,\bold x_{t+\tau}\}$:
\begin{equation}
p_{\tau}(\bold x,\bold y)\approx\mathbb{P}(\bold r^{LD}(\bold x_{t+\tau})=\bold r^{LD}(\bold y)\mid\bold x_{t}=\bold x)\cdot\pi_{local}(\bold r^{LD}(\bold y),\bold y)
\end{equation}
where $\pi_{local}(\bold r, \bold y)$ is the local equilibrium probability distribution of the conformation $\bold y$ inside the set which shares the same value of $\bold r^{LD}(\bold y)=\bold r$. In words, this suggests that the transition from $\bold x$ to $\bold y$ can be ``decomposed'' into a two-step process:

\begin{enumerate}
	\item The transition from $\bold x$ to any conformation that belongs to the level set $\{\mathbf y'\mid \mathbf y'\in \Omega, \bold r^{LD}(\bold y')=\bold r^{LD}(\bold y)\}$.
	\item The local equilibration to the conformation $\bold y$ inside this level set.
\end{enumerate}

\begin{remark}
Our definitions of lumpability and decomposability slightly deviate from the formalism presented by Bittracher et al.\cite{bittracher2023optimal} However, it can be demonstrated that the two descriptions are equivalent (see Supporting Information). The current formulation is adopted for its conceptual clarity, and it clearly indicates the relationship between coarse-grained transition densities and conditional distributions associated with the RC. Furthermore, this formulation naturally leads to a loss function for evaluating the RC, which facilitates optimization of the RC using generative modelling techniques (see Section 2.3).
\end{remark}

Furthermore, in reversible dynamics satisfying equation 3, these two conditions have been shown to be equivalent \cite{bittracher2023optimal}. Notice although equation 8 and equation 9 do not involve modelling $\psi_i$ or $\mathcal T(\tau)$ explicitly, the modelling of transition probability $p_\tau(\bold x,\bold y)$ is closely linked to the transfer operator theory since $\mathcal T(\tau)$ is dependent on $p_\tau(\bold x,\bold y)$ as in equation 1. In fact, if we consider the projected dynamics $\{\bold r^{LD}_t\}$ inside the $\bold r^{LD}$ space, which is equivalently governed by a reduced transfer operator $\mathcal T_{\bold r}(\tau)$ in the RC space, it can be shown that $\mathcal T_{\bold r}(\tau)$ is a close approximation to the original system $\mathcal T(\tau)$ \cite{bittracher2018transition}. As a result, the leading eigenvalues $\lambda_{i}^{\bold r}$ and eigenfunctions $\psi_i^{\bold r}$ of $\mathcal T_{\bold r}(\tau)$ are also close approximations to the original leading $\lambda_i$ and $\psi_i$. Physically, this means the full-dimensional systems dynamics, in terms of the major slow modes ($\psi_i$) and their associated timescales ($\lambda_i$), are encoded in the RC space with a minimum loss, if the RC is indeed optimal.

\clearpage

\subsection{Flow Matching for Reaction Coordinates (FMRC)}

\subsubsection{Modelling Objectives}

\begin{figure}[h]
    \centering
    \includegraphics[scale=0.75]{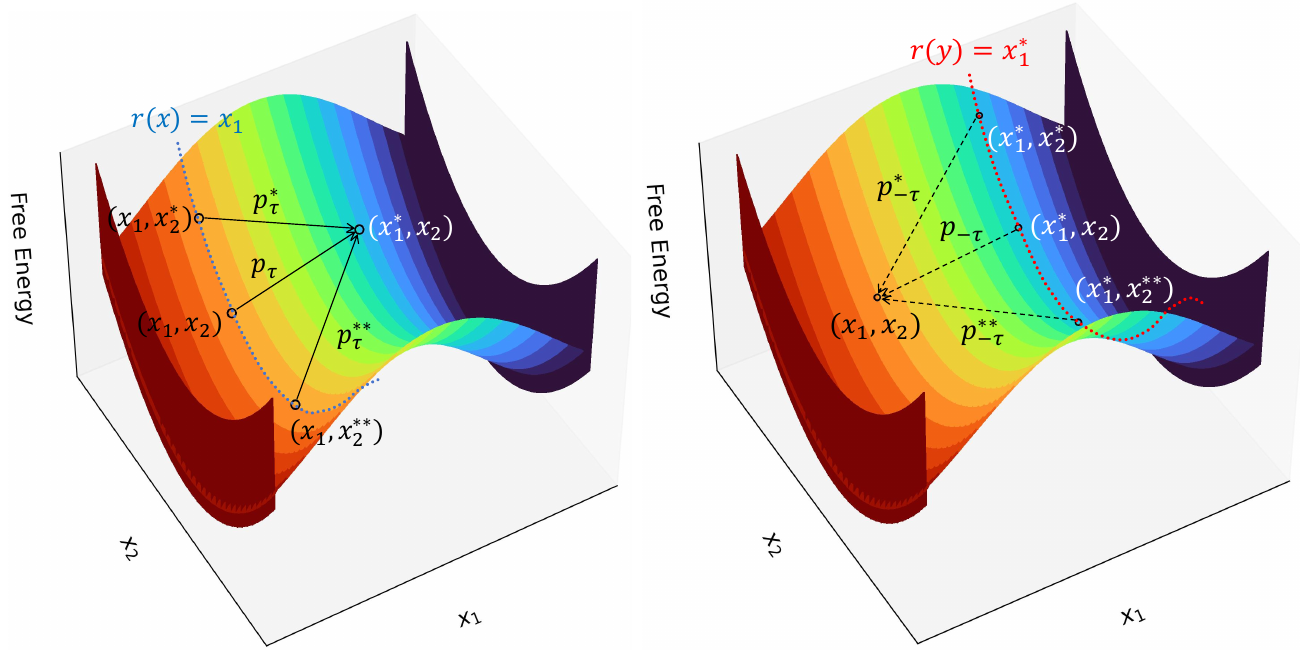}
    \caption{4D plots of an optimal RC that satisfies both lumpability (left) and decomposability (right) for a 2D model system. The x-axis and y-axis of the plot are the 2D coordinates of the system. The z-axis represents the relative free energy of each coordinate. The color bar represents the value of an optimal RC $\bold r = x_1$. The initial coordinate $\bold x_t$ of a transition with lag time $\tau$ is denoted in black and the final coordinate $\bold x_{t+\tau}$ is denoted in white. (left) An $\bold r(\bold x)$ isoline of the initial coordinates is shown as a blue dotted line. We further denote three initial coordinates on this isoline as $\mathbf x^*:=(x_1,x_2^*)$, $\mathbf x:=(x_1,x_2)$ and $\mathbf x^{**}:=(x_1,x_2^{**})$. Three Markovian transitions from these coordinates to a final coordinate $(x_1^*,x_2)$ are shown as black arrows, and the probabilities of these transitions are denoted as $p_\tau^{*}:=p_\tau(\mathbf x^*,\mathbf y)$, $p_\tau:=p_\tau(\mathbf x,\mathbf y)$ and $p_\tau^{**}:=p_\tau(\mathbf x^{**},\mathbf y)$ depending on their initial $x_2$ values. (right) An $\bold r(\bold y)$ isoline of the final coordinates is shown as a red dotted line. We further denote three final coordinates on this isoline as $\mathbf y^*:=(x_1^*,x_2^*)$, $\mathbf y:=(x_1^*,x_2)$ and $\mathbf y^{**}:=(x_1^*,x_2^{**})$. Three Markovian transitions from the initial coordinate $(x_1,x_2)$ to these coordinates are shown as backward black dotted arrows to emphasize that this is not a backward transition but the backward transition probability defined in equation 10. The probabilities of these transitions are denoted as $p_{-\tau}^{*}:=p_{-\tau}(\mathbf y^*,\mathbf x)$, $p_{-\tau}:=p_{-\tau}(\mathbf y,\mathbf x)$ and $p_{-\tau}^{**}:=p_{-\tau}(\mathbf y^{**},\mathbf x)$ depending on their final $x_2$ values.}
    \label{fig:1}
\end{figure}

We next move on to present our novel algorithm, FMRC, for the computation of optimal RC as $\bold r^{FMRC} \approx \bold r^{LD}$ based on lumpability and decomposability. For the ease of analysis, we introduce the backward transition probability
\begin{equation}
p_{-\tau}(\mathbf{y}, \mathbf{x}) = \mathbb{P}(\mathbf{x}_{t} = \mathbf{x} \mid \mathbf{x}_{t+\tau} = \mathbf{y}) 
\end{equation}
which represents the posterior distribution of the state \(\mathbf{x}_t\) given the future state \(\mathbf{x}_{t+\tau}\). Then the conditions of lumpability and decomposability (equations 8 and 9) can be both expressed in terms of conditional probabilities for all time-lagged pairs \(\{\mathbf{x}_t, \mathbf{x}_{t+\tau}\}\):
\begin{equation}
\text{Lumpability}\to p_{\tau}(\mathbf{x},\mathbf{y})\approx \mathbb{P}(\bold x_{t+\tau}=\bold y\mid\bold r^{LD}(\bold x_{t})=\bold r^{LD}(\bold x)) \tag{11A}
\end{equation}
\begin{equation}
\text{Decomposability}\to p_{-\tau}(\mathbf{y},\mathbf{x})\approx \mathbb{P}(\bold x_{t}=\bold x\mid\bold r^{LD}(\bold x_{t+\tau})=\bold r^{LD}(\bold y)) \tag{11B}
\end{equation}

The proof of equation 11B be found in the work of Zhang et al.~\cite{zhang2024flow} and is also provided in the Supporting Information. To provide a physically intuitive interpretation of equations 11A and 11B, we use a simple 2D double well potential to illustrate this idea (Figure 1). It is not difficult to see for this simple system, the x-axis coordinate, $x_1$, is the only relevant degree to the transition process, and therefore, an intuitive optimal RC could be $\bold r = x_1$. In Figure 1 (left), three Markovian transitions are initiated on the same $\bold r(\bold x) = x_1$ isoline and our lumpability formulation from equation 10 states that $p_\tau^*=p_\tau=p_\tau^{**}$ (See Figure 1 legend for their definition). This is reasonable, since the fluctuation along the $x_2$ coordinate does not involve crossing any energy barriers and therefore, should be very fast. The actual transition-limiting degree is $x_1$ and should be the major contributor to $p_\tau^*$, $p_\tau$ or $p_\tau^{**}$. Thus, the probability of any transition initiating from the blue isoline to $\mathbf y$ should be a close approximation to $p_{\tau}(\mathbf{x},\mathbf{y})$ in this case. The same logic follows for our decomposability formulation. The fluctuation along the $x_2$ coordinate is expected to be very fast and should not affect $p^{*}_{-\tau}$, $p_{-\tau}$ or $p^{**}_{-\tau}$ significantly. As a result, as our decomposability formulation stated, given that the final coordinates reside on the red isoline, the probability of finding the initial coordinate at $\mathbf x$ does not depend on the exact location of the final coordinate on the isoline. 

Therefore, if we would like to approximate $\bold r^{LD}$ as $\bold r^{FMRC}$ in a data-driven manner, we can minimize the difference between the left hand side (l.h.s) and the right hand side (r.h.s.) of equations 11A and 11B for all time-lagged pairs  $\{\bold x_t,\bold x_{t+\tau}\}$ by inserting $\bold r^{FMRC}$ as $\bold r^{LD}$. Assuming sufficient sampling, the l.h.s. of both equations could be trivially represented as the empirical distribution from the MD trajectory dataset. On the other hand, we model the r.h.s of both equations together with the optimal RC $\mathbf r^{FMRC}$ in our FMRC algorithm, which is detailed below. The graphical representations of our algorithms are depicted in Figure 2.

\begin{figure}[h]
    \centering
    \includegraphics[scale=0.5]{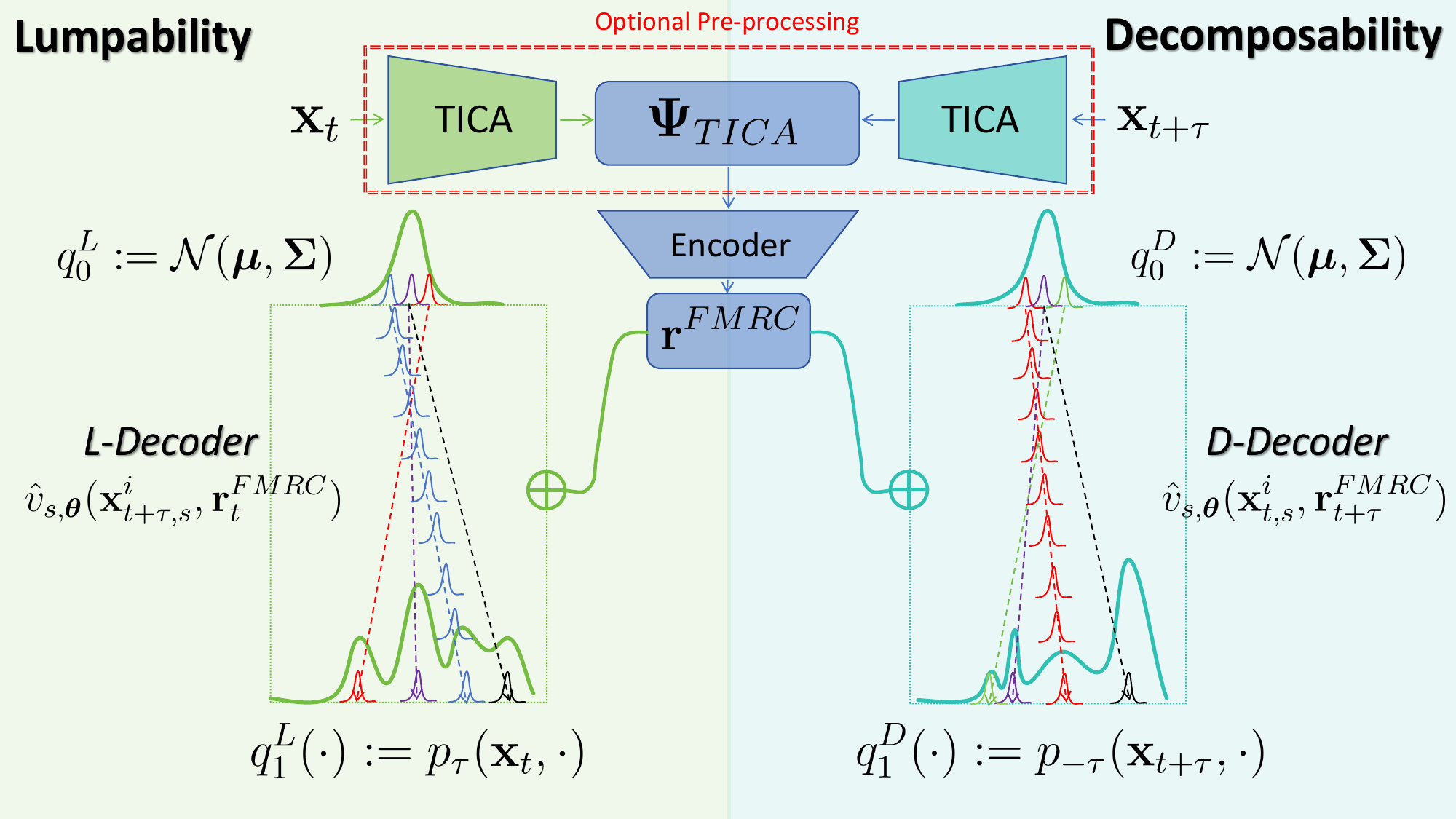}
    \caption{The overall architecture of FMRC. For each time-lagged pair $\{\bold x, \bold x_t\}$, they are pre-processed by TICA (optional) and encoded by the encoder. The L-decoder or D-decoder then optimizes the vector field $\hat v_{s,\boldsymbol \theta}(\bold x^i_{t+\tau,s},\bold r^{FMRC}_t)$ or $\hat v_{s,\boldsymbol \theta}(\bold x^i_{t,s},\bold r^{FMRC}_{t+\tau})$ conditioned on (denoted as $\oplus$) the latent variable $\bold r^{FMRC}(\bold x_t)$ or $\bold r^{FMRC}(\bold x_{t+\tau})$ of the linear interpolation Gaussian path which transforms independent samples in $q_0^L$ or $q_0^D$ into independent samples in $q_1^L$ or $q_1^D$, respectively. Please refer to the main text for a detailed explanation. }
    \label{fig:2}
\end{figure}

\subsubsection{Representing $\bold r^{LD}$ as an encoder}

We construct our $\bold r^{FMRC}$ in a two-step process:

\begin{enumerate}
	\item (Optional) Compute a TICA transformation $\boldsymbol \Psi_{TICA} = [\psi_1,\psi_2,...,\psi_{d_{int}}]^\top$ at the training lag time $\tau$ with the trajectory data and truncate the series at an intermediate number of dimensions $d_{int}$. This is not strictly necessary but we found this option is the simplest and the most physically intuitive way for easier and better $\bold r^{FMRC}$ learning for biomolecular systems we tested in this article, especially for more complex systems. Thus, all FMRC models in this article are trained with this pre-processing step. We will discuss the choice of $d_{int}$ below.
	\item Apply a neural network transformation $\boldsymbol \Xi^{Encoder}_{nn,\boldsymbol \theta}$, which we refer to as the encoder, to $\boldsymbol \Psi_{TICA}$ (or directly to $\bold x$) to yield a low-dimensional latent variable as $\bold r^{FMRC}$:
		\begin{equation}
			\bold r^{FMRC} = \boldsymbol \Xi^{Encoder}_{nn,\boldsymbol \theta}(\boldsymbol \Psi_{TICA}) \tag{12}
		\end{equation}
\end{enumerate}

\subsubsection{Flow Matching}

With $\bold r^{FMRC}$ at hand, we proceed to model $\mathbb P(\bold x_{t+\tau}=\bold y \mid \bold r^{FMRC} (\bold x_t) = \bold r^{FMRC}(\bold x)) $ and $\mathbb P(\bold x_t = \bold x \mid \bold r^{FMRC}(\bold x_{t+\tau}) = \bold r^{FMRC}(\bold y))$ using continuous normalizing flow (CNF) \cite{chen2018neural}. CNF is a generative modelling technique which allows the parameterization of a complex target probability distribution $q_1:= q_1(\bold z)$ of a random variable $\bold z$ from a simple prior probability distribution $q_0:= q_0(\bold z)$ that allows efficient sampling. This is achieved by a process which propagates the prior distribution $q_0$, through a mixture of paths $q_s:= q_s(\bold z)$, towards the target probability $q_1$ that is governed by an ordinary differential equation (ODE). Specifically, we use a novel simulation-free CNF training method, known as flow matching (FM) \cite{lipman2022flow}, for our training process. In FM, instead of running the ODE simulation explicitly, we use a time-dependent vector field $v_s$ to model the probability path $q_s$ governed by the ODE implicitly
\begin{equation}
\frac{\bold d\bold z_s}{\bold ds} = v_s(\bold z_s) \tag{13}
\end{equation}
where ${\bf z}_0\sim q_0$ and $s$ is a scalar ranging between $0$ and $1$ which can interpreted as the ``time'' of the propagation process. To distinguish it from the system's physical time \( t \), we will denote the virtual time in flow matching by \( s \) throughout the following discussion.

Interestingly, since $v_s$ does not make any assumption on the path mixture $q_s$ of the propagation \cite{lipman2022flow,tong2023improving}, we can specify an arbitrary propagation path as $q_s$. Here, we follow the basic form of FM \cite{tong2023improving}, which specifies $q_s$ in a way that individual samples $\bold z^i_0$, $\bold z^i_1$ from $q_0$ and $q_1$ are independently linearly interpolated via a Gaussian path. Under this setting, given that $\bold z^i_0$, $\bold z^i_1$ are the end points of the linear interpolation, $\bold z_s^i$ on this interpolation can be sampled as:
\begin{equation}
\bold z_s^i \sim p_s(\bold z_s^i\mid \bold z_0^i,\bold z_1^i)= \mathcal N(\boldsymbol \mu_s,\boldsymbol \sigma) \tag{14}
\end{equation}
where $\mathcal N(\boldsymbol \mu_s,\boldsymbol \sigma)$ is a Gaussian with mean $\boldsymbol \mu_s = s\bold z^i_1 + (1-s)\bold z_0^i$ and diagonal covariances $\boldsymbol \sigma$ with all entries as a small value. The value of the vector field along this specific interpolation path is:
\begin{equation}
v_s(\bold z_s^i\mid \bold z_0^i,\bold z_1^i) = \bold z_1^i-\bold z_0^i \tag{15}
\end{equation}
Tong et al.\cite{tong2023improving} further show that under such setting, we can parameterize the vector field $\hat v_{s,\boldsymbol \theta}$ for the propagation by minimizing the following loss function $\mathcal L_{FM}$, which is an expectation over all individual samples $\bold z^i_0$, $\bold z^i_1$ from $q_0$ and $q_1$:
\begin{equation}
\mathcal L_{FM}(\boldsymbol \theta) = \mathbb E[||\hat v_{s,\boldsymbol \theta}(\bold z_s^i) -(\bold z_1^i-\bold z_0^i)||^2] \tag{16}
\end{equation}
where $s$ is sampled uniformly between the interval $[0,1]$.
In the case where \(\mathcal{L}_{FM}\) is ideally minimized, the ODE defined by equation 13 can accurately transform samples from the distribution \(q_0\) into the distribution \(q_1\).

\begin{remark}
More rigorously, if \(\boldsymbol{\sigma} > \mathbf{0}\) in equation 14, the distributions governed by the ODE obtained from FM at \(s = 0\) and \(s = 1\) are given by \(q_0 * \mathcal{N}(\mathbf{0}, \boldsymbol{\sigma})\) and \(q_1 * \mathcal{N}(\mathbf{0}, \boldsymbol{\sigma})\), respectively, where \(*\) denotes the convolution operator for probability density functions. As \(\boldsymbol{\sigma} \to \mathbf{0}\), these distributions converge to \(q_0\) and \(q_1\). In our experiments, we choose a small but positive \(\boldsymbol{\sigma}\), primarily to ensure numerical stability during training.
\end{remark}

\subsubsection{Flow Matching as Decoders}

In FMRC, we define $q_0$ and $q_1$ separately for modelling lumpability and decomposability in parallel, which we denoted as $q_0^L$, $q_0^D$,$q_1^L$ and $q^D_1$, respectively. We first define the prior distribution, $q_0^L$ and $q_0^D$, as:
\begin{equation}
q_0^L=q_0^D := \mathcal N(\boldsymbol \mu,\boldsymbol \Sigma) \tag{17}
\end{equation}
where $\mathcal N(\boldsymbol \mu,\boldsymbol \Sigma)$ is a Gaussian distribution with mean and diagonal covariance being the mean and variance of $\boldsymbol \Psi_{TICA}$: $\boldsymbol \mu = [\mu(\psi_1^{TICA}),...,\mu(\psi_{d_{int}}^{TICA})]^\top$ and $\boldsymbol \Sigma = diag[\sigma^2(\psi_1^{TICA}),...,\sigma^2(\psi_{d_{int}}^{TICA})]$. We then define the target distribution, $q_1^L$ and $q_1^D$ as:
\begin{equation}
q_1^L(\cdot):= p_\tau(\bold x_t,\cdot)\tag{18A}
\end{equation}
\begin{equation}
q_1^D(\cdot):= p_{-\tau}(\bold x_{t+\tau},\cdot)\tag{18B}
\end{equation}
for given $\bold x_t$ and $\bold x_{t+\tau}$.
Following the FM training objective from equation 16, we can parameterize $q_1^L$ and $q_1^D$ by regressing the vector fields $\hat v_{s,\boldsymbol \theta}(\bold x^i_{t+\tau,s},\bold x_t)$ and $\hat v_{s,\boldsymbol \theta}(\bold x^i_{t,s},\bold x_{t+\tau})$ through minimizing the loss functions $\mathcal L^L$ and $\mathcal L^D$:
\begin{equation}
\mathcal L^L (\boldsymbol \theta)=\mathbb E[||\hat v_{s,\boldsymbol \theta}(\bold x^i_{t+\tau,s},\bold x_t) -(\bold x_{t+\tau}-\bold x^i_{t+\tau,0})||^2] \tag{19A}
\end{equation}
\begin{equation}
\mathcal L^D (\boldsymbol \theta)=\mathbb E[||\hat v_{s,\boldsymbol \theta}(\bold x^i_{t,s},\bold x_{t+\tau}) -(\bold x_{t}-\bold x^i_{t,0})||^2] \tag{19B}
\end{equation}
Similar to equation 15, $\bold x^i_{t+\tau,s}$ or $\bold x^i_{t,s}$ denotes samples from the Gaussian path at ``time'' $s$ through equation 13. $\bold x_{t+\tau}$, $\bold x_t$ are individual samples from the target distribution $q_1^L$, $q_1^D$  (hence directly from the trajectory dataset) and $\bold x^i_{t+\tau,0}$, $\bold x^i_{t,0}$ are individual samples from prior distribution $q_0^L$, $q_0^D$. Similarly, $s$ is uniformly sampled between 0 and 1. Notice that in order to model the conditional probability $p_\tau(\bold x_t,\cdot)$ and $p_{-\tau}(\bold x_{t+\tau}, \cdot)$, we let the vector fields $\hat v_{s,\boldsymbol \theta}(\bold x^i_{t+\tau,s},\bold x_t)$ and $\hat v_{s,\boldsymbol \theta}(\bold x^i_{t,0},\bold x_{t+\tau})$ condition on $\bold x_t$ and $\bold x_{t+\tau}$ as well, in contrast to the original FM training objective from equation 16. Furthermore, if we consider conditioning on the latent variables of the encoder $\bold r_t^{FMRC}:=\bold r^{FMRC}(\bold x_t)$ and $\bold r_{t+\tau}^{FMRC}:=\bold r^{FMRC}(\bold x_{t+\tau})$ and minimize the following loss functions instead:
\begin{equation}
\tilde {\mathcal L}^L (\boldsymbol \theta)=\mathbb E[||\hat v_{s,\boldsymbol \theta}(\bold x^i_{t+\tau,s},\bold r^{FMRC}_t) -(\bold x_{t+\tau}-\bold x^i_{t+\tau,0})||^2] \tag{20A}
\end{equation}
\begin{equation}
\tilde {\mathcal L}^D (\boldsymbol \theta)=\mathbb E[||\hat v_{s,\boldsymbol \theta}(\bold x^i_{t,0},\bold r^{FMRC}_{t+\tau}) -(\bold x_{t}-\bold x^i_{t,0})||^2] \tag{20B}
\end{equation}
It can be seen that the minimum of equations 20A and 20B provides an upper bound of 19A and 19B and they are equal if and only if $\bold r^{FMRC}$ satisfies lumpability and decomposability stated in equations 11A and 11B. \cite{zhang2024flow} Thus, for FMRC, we assemble the encoder with both decoders and minimize the FMRC loss function $\mathcal L^{FMRC}(\boldsymbol \theta)$ as
\begin{equation}
\mathcal L^{FMRC}(\boldsymbol \theta) = \tilde {\mathcal L}^L (\boldsymbol \theta) + \tilde {\mathcal L}^D (\boldsymbol \theta) \tag{21}
\end{equation}
to optimize all parameters $\boldsymbol \theta$ simultaneously with back propagation. Once equation 21 is properly minimized, $\bold r^{FMRC}$ will satisfy lumpability and decomposability and hence $\bold r^{FMRC} \approx \bold r^{LD}$.

\subsection{The Comparison of RC Learning Algorithms}

\begin{figure}[h]
    \centering
    \includegraphics[scale=0.5]{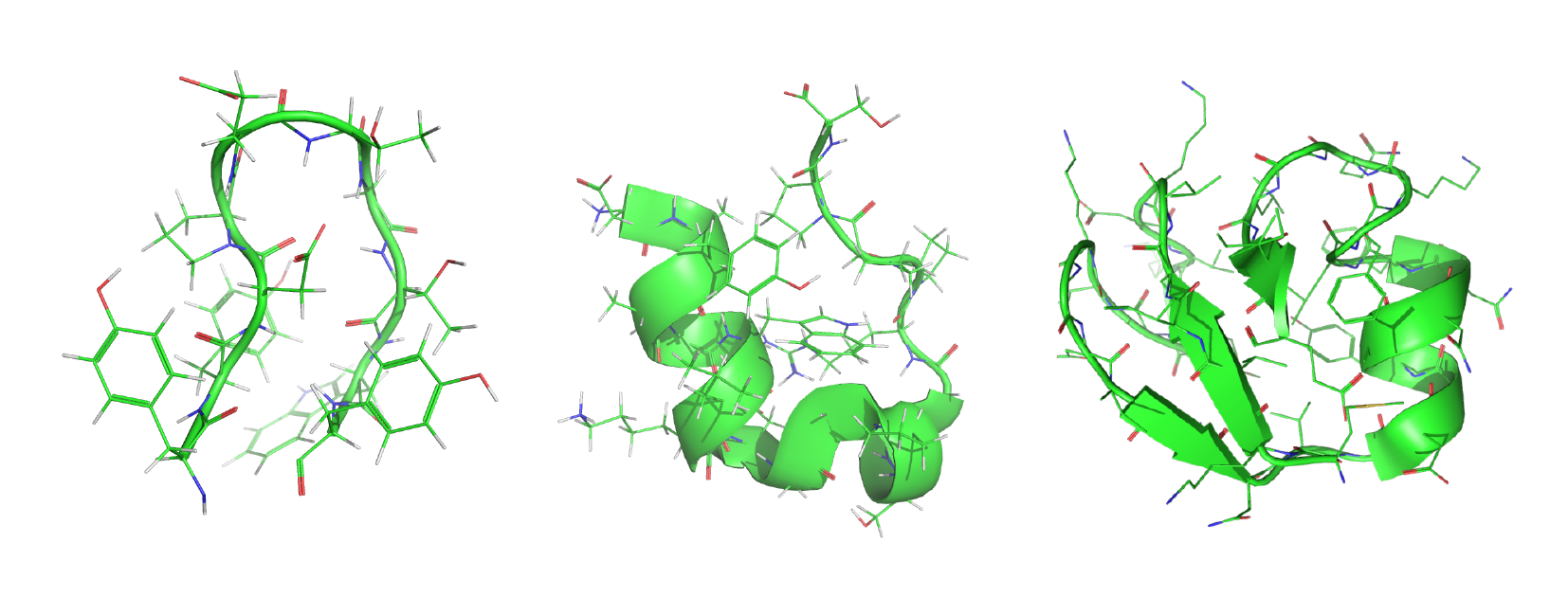}
    \caption{Extensively sampled trajectories of three biomolecular systems used for FMRC performance evaluation and RC comparison in this study: (left) chignolin variant CLN025 (PDB ID: 2RVD), (middle) Trp-Cage (PDB ID: 2JOF) and (right) NTL-9 (PDB ID: 2HBA).}
    \label{fig:3}
\end{figure}

To further test the efficacy of FMRC, we compared the optimality of $\mathbf r^{FMRC}$ with RC learned by TICA\cite{perez2013identification} (denoted as $\mathbf r^{TICA}$), SRV\cite{chen2019nonlinear} (denoted as $\mathbf r^{SRV}$) and SPIB\cite{wang2021state} (denoted as $\mathbf r^{SPIB}$) using the same dataset. Detailed descriptions of these algorithms are provided in the Supporting Information. To ensure a fair quantitative evaluation of RC optimality, we adopted a simple strategy based on VAC\cite{noe2013variational} and MSM\cite{prinz2011markov} (see Supporting Information). 
We use three extensively sampled ($\sim$ 0.1-1.1 ms) MD trajectory datasets of mini-protein folding with increased complexity, which involves the formation of a range of secondary structures such as beta-hairpins or alpha helices from D.~E.~Shaw research \cite{lindorff2011fast} (Figure 3). This data-rich scenario was assumed to maximize the performance of all methods, leaving the comparison of sampling efficiency for RC learning to future studies. We then learned $\bold r^{TICA}$, $\bold r^{SRV}$, $\bold r^{FMRC}$ and $\bold r^{SPIB}$ from the same datasets using the same $\tau$ and performed k-means clustering in these RC spaces using an equal number of clusters.
For $\bold r^{FMRC}$ and $\bold r^{SPIB}$, each dimension of the RC is normalized between 0 and 1 before clustering, whereas for $\bold r^{TICA}$ and $\bold r^{SRV}$, the eigenvectors were scaled to represent a kinetic distance \cite{noe2015kinetic} for optimal clustering. Finally, we constructed MSMs using cluster assignments from the different RC spaces with the same $\tau_{MSM}$ to ensure high-quality MSMs that obey Markovianity. 

To assess and compare the optimality of different RCs, we evaluated the eigenvalues $\hat \lambda_i^{MSM}$ of the leading eigenvectors $\hat \psi_{i}^{MSM}$ from the MSMs constructed in each RC space. Given that the same hyperparameters (see Supporting Information) were used throughout the construction process and assuming minimal performance differences in the k-means algorithm across all RC spaces, we inferred the quality of the original RCs implicitly from the quality of the constructed MSMs based on VAC: the higher the eigenvalues $\lambda_i^{MSM}$, the higher the dynamical content preserved in the RC space and the better the RC. The same strategy was also employed to determine the optimal truncation dimension $d_{int}$ for $\boldsymbol \Psi_{TICA}$ during FMRC pre-processing for all systems in this study. 

Notice that we do not intend to restrict the scope of the FMRC algorithm or its comparison solely to MSM construction. Instead, we aim to explore broader potential downstream applications such as enhanced sampling methods. In common enhanced sampling techniques such as umbrella sampling or metadynamics, the RC used to apply bias potential is typically limited a small number of dimensions (usually 2) for algorithmic efficiency. Therefore, we constrained the dimensionality of the RC to 2 across all learning algorithms. This also facilitates easier visualization and hence better human interpretation of the system dynamics. 

Moreover, the hyperparameter $\tau$ plays an essential role in filtering out fast dynamics with timescales $t_i$ below this threshold. For enhanced sampling purposes, the ideal RC should capture as many slow processes as possible. To achieve this, we used a relatively small $\tau$ across all algorithms for RC comparison. Nevertheless, we also performed a comparison of $\bold r^{FMRC}$ learned with different $\tau$ to examine the impact of this hyperparameter on RC learning by FMRC.

Another important aspect in the evaluation of the RC machine learning algorithm is the variance of performance by these algorithms due to stochasticity from e.g. random seeds. In practical applications, it would be tedious and inefficient to repeat learning with the same hyperparameters and different random seeds for better performance. Therefore, after tuning the hyperparameters for satisfactory learning performance, we repeatedly learn 10 RCs using the same hyperparameters with the whole trajectory dataset for each algorithm in each example. We then compare the mean and standard deviation of the eigenvalues $\hat \lambda_i^{MSM}$ to incorporate algorithm consistency into our comparison. 

In addition to the data-rich scenario, we also tested on the sensitivity of the FMRC algorithm to the amount of training data. To do this, for each example, we trained different FMRC models with the leading 10\%/25\%/50\% portion of the long trajectory and then transformed the whole trajectory into the RC space. We then performed the regular k-means clustering and constructed MSM using the same hyperparameters and compared the eigenvalues $\hat \lambda_i$ with the MSM we constructed using the $\bold r^{FMRC}$ we learned from the whole trajectory dataset.  

Finally, to investigate the origin of the differences in the RC learning performance between different algorithms, we projected all trajectory data points to different RC spaces and performed a PCCA+\cite{roblitz2013fuzzy} analysis to investigate the ability to distinguish different metastable states in the RC spaces learned by different algorithms. 

\subsection{Hyperparameters and Software Packages}

Another notable feature of the FMRC algorithm is that FMRC is very easy to train following our training protocol. For all three systems, we used the same feature selection strategy, only adjusted $d_{int}$ and $\tau$ and left hyperparameters such as learning rate, number of hidden layers/hidden nodes, and training epochs invariant for all three examples to achieve good performance. This should be pretty useful in practical applications since $d_{int}$ and $\tau$ are more physically intuitive than tuning the size or learning rate of the neural networks. For $\bold r^{SRV}$ and $\bold r^{SPIB}$, instead of running an exhaustive search for optimal training hyperparameters, we refer to the training protocols in previous studies \cite{sidky2019high,wang2024information} utilizing these algorithms for similar systems and apply the same set of hyperparameters for all three examples. See Supporting Information for a detailed description of feature selection and model training for $\bold r^{TICA}$,$\bold r^{SRV}$, $\bold r^{FMRC}$ and $\bold r^{SPIB}$ together with subsequent MSM construction in these spaces.

We performed TICA, k-means clustering, MSM construction, and PCCA+ using the Deeptime\cite{hoffmann2021deeptime} library. All neural network models for MSM construction were trained with PyTorch 2.2.2 with CUDA support. We trained SRV using the code from: \url{https://github.com/andrewlferguson/snrv} and trained SPIB using the code from: \url{https://github.com/wangdedi1997/spib}. The code of FMRC is available at: \url{https://github.com/Mingyuan00/Flow_Matching_for_RC} and we have uploaded all our analysis scripts and PyTorch model .pt files to Zenodo at \url{https://zenodo.org/records/13474614}.

\section{Results and Discussion}

\subsection{FMRC Application to Chignolin Variant CLN025}

\begin{figure}[h]
    \centering
    \includegraphics[scale=0.75]{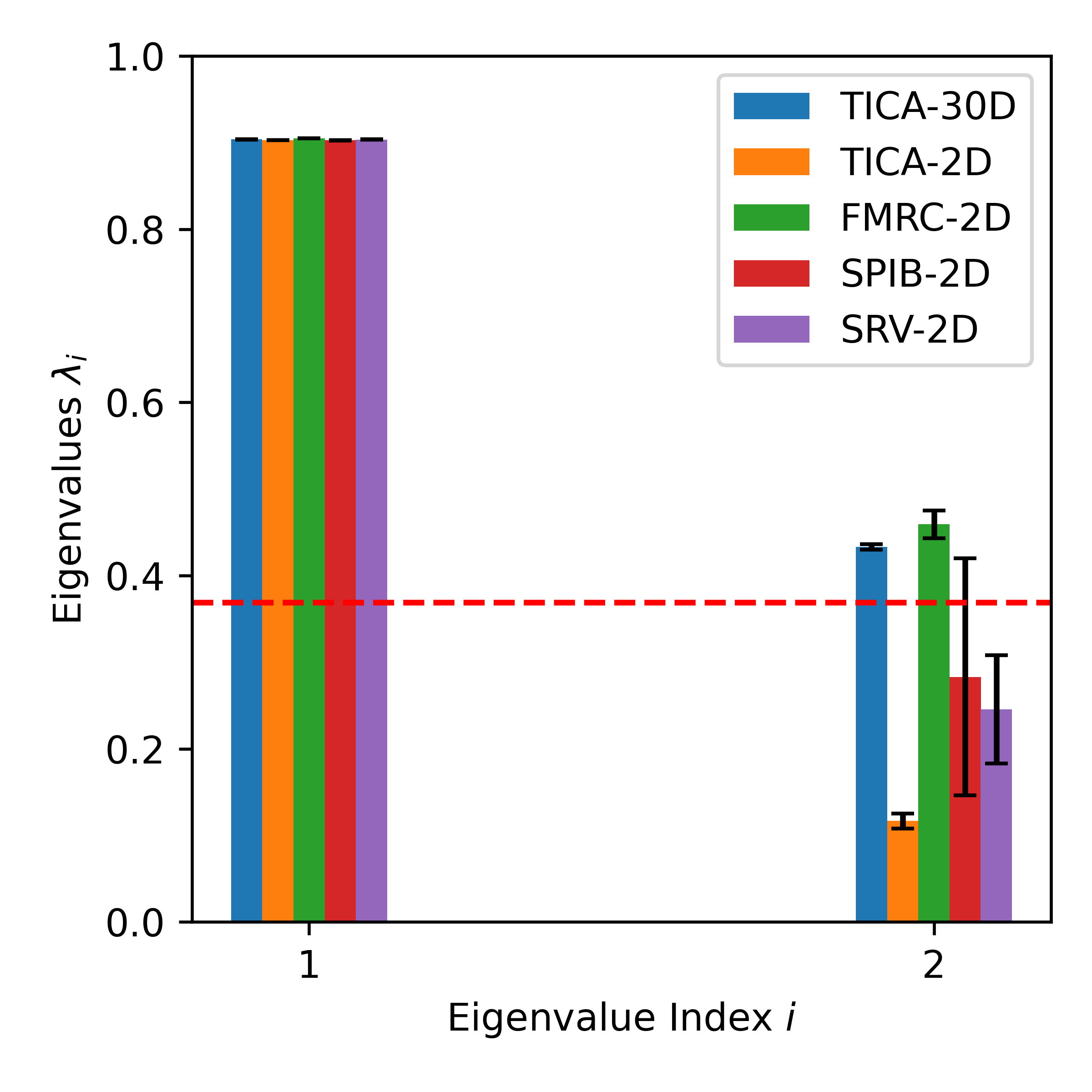}
    \caption{Comparison of $\hat \lambda_i^{MSM}$ of MSMs constructed in different RC spaces for CLN025. A red dashed line at $\lambda_i = 0.369$ has been drawn to denote a cutoff for the corresponding timescales lower than the $\tau^{MSM}$ for MSM construction. This indicates that the constructed MSM has failed to identify this slow process.}
    \label{fig:4}
\end{figure}

We first demonstrate the efficacy of our FMRC algorithm using a 106 $\mu$s trajectory of a mini-protein chignolin variant CLN025 (YYDPETGTWY, 10 amino acids, PDB ID: 2RVD) simulated at 340K. In solution, CLN025 maintains a stable beta-hairpin structure under room temperature and can interconvert between this hairpin structure (``folded'') with a fully extended conformation (``unfolded'') at its melting temperature of approximately 340K. The three-dimensional atomic coordinates were first converted into an extensive set of internal coordinates including interatomic distances and torsional angles (see Supporting Information for details) as our input features. We then selected a short lag time of $\tau = 2$ ns for RC learning. To establish an overview of the system, we performed a regular TICA+MSM analysis and identified two major slow processes in the system (Figure S1) at $\tau^{MSM}=40$ ns. Our goal here is to construct an optimal RC that is sufficient to characterize the folding process and capture these two slow modes within its RC space.

Following the protocol described in Theory and Methods, we built a series of MSM by clustering inside $\boldsymbol \Psi_{TICA}$ spaces with different dimensions. We then selected the truncation dimension as $d_{int}=30$ based on the comparison of $\hat \lambda_1^{MSM}$ and  $\hat \lambda_2^{MSM}$ and used $[\hat \psi^{TICA}_1,...,\hat \psi^{TICA}_{30}]^\top$ as the input for our FMRC neural network. We then trained 10 FMRC models with the same hyperparameters (see Supporting Information) and obtained 10 $\bold r^{FMRC}$. For comparison, we learned both a 30-dimensional $\bold r^{TICA}_{30D}$ and a 2-dimensional $\bold r^{TICA}_{2D}$. Although the algorithm is deterministic, we repeated the stochastic k-means clustering to obtain 10 MSMs for $\bold r^{TICA}_{30D}$ and $\bold r^{TICA}_{2D}$ to provide a reference to the variance introduced by k-means clustering. We also trained 10 SRV models using the input features directly and the same hyperparameters (see Supporting Information). For SPIB, we performed k-means clustering inside the 30-dimensional $\boldsymbol \Psi_{TICA}$ space which we used as our FMRC input to generate the state labels $\bold y_{t+\tau}$. These state labels were then supplied to SPIB together with the input features as inputs. The training process was repeated 10 times with the same hyperparameters (See Supporting Information) for 10 SPIB models. We then calculated the mean and standard deviation of the MSM eigenvalues $\hat \lambda_i^{MSM}$ from these RC spaces for a quantitative comparison which we summarized the results in Figure 4.

From this eigenvalue comparison, we can see that there are no significant differences between the performance of capturing the slowest mode $\psi_1$ of the system. All models successfully encoded $\psi_1$ in their RC space with the correct timescale $t_i$ and minimal variance. However, when it comes to $\psi_2$, only $\bold r^{FMRC}$ among the two-dimensional RCs consistently encodes this slow mode with minimal variance. In contrast, $\bold r^{SPIB}$, suffers from large modelling variance and can only capture this slow mode occasionally. Neither $\bold r^{TICA}_{2D}$ nor $\bold r^{SRV}$ is able to capture $\psi_2$, underpinning the limitations of methods based on spectral decomposition. Notably, $\bold r^{FMRC}$ outperforms even $\bold r^{TICA}_{30D}$, suggesting that $\bold r^{FMRC}$ can efficiently encode a substantial amount of dynamical information in a reduced representation. This is consistent with the theoretical expectation that the lumpability and decomposability principles imply that the RC space spanned by $\bold r^{FMRC} \approx \bold r^{LD}$ has a reduced transfer operator $\mathcal T_{\bold r}(\tau)$ with leading eigenvalues similar to those of the original transfer operator $\mathcal T(\tau)$. 

Next, we examine the minimum $\tau$ required for the ITS estimation to converge for MSMs constructed in different RC spaces. This property is practically important, since convergence with a shorter $\tau$ allows for shorter, parallel-sampled trajectories as input, which is known to enhance the sampling of transition events. Furthermore, MSMs constructed with shorter $\tau$ offer higher resolution and can provide more detailed biophysical insights. The ITS analysis in Figure S2 shows that MSMs constructed from $\bold r^{FMRC}$ can consistently converge with a shorter $\tau$. Moreover, MSM constructed from $\bold r^{FMRC}$ can accurately predict future macrostate populations, as demonstrated by the Chapman-Kolmogorov test (Figure S3).

\begin{figure}[h]
    \centering
    \includegraphics[scale=0.1]{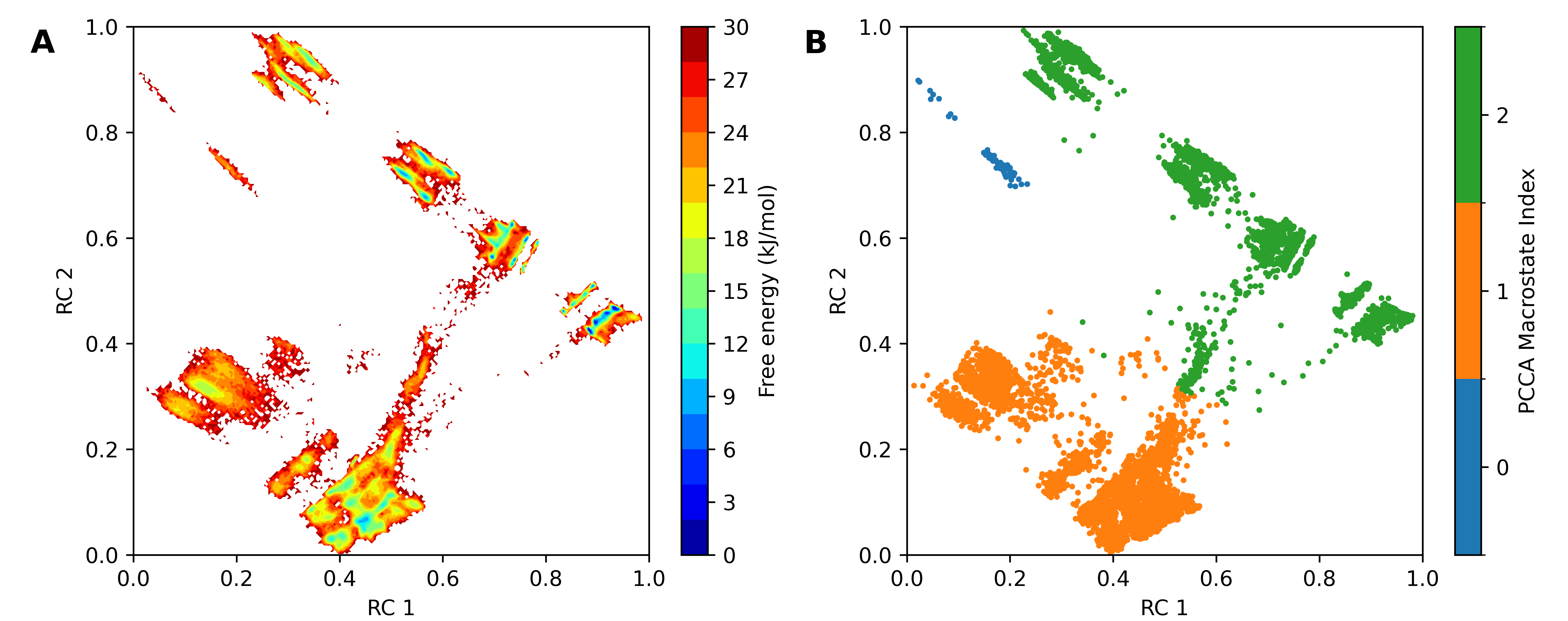}
    \caption{(A) The 2D FES projection of CLN025 and (B) the PCCA+ macrostate assignment projection for CLN025 in the normalized $\bold r^{FMRC}$ space where the best MSM was constructed.}
    \label{fig:5}
\end{figure}

We then projected all data points $\bold x$ of the trajectory to the 2D space defined by $\bold r^{TICA}_{2D}$, $\bold r^{SRV}$ , $\bold r^{FMRC}$ and $\bold r^{SPIB}$ and calculated the 2D projected FES in these spaces (Figure 5A, S4A, S5A \& S6A). Specifically, for $\bold r^{SRV}$, $\bold r^{FMRC}$ and $\bold r^{SPIB}$, we utilized both the best and worst RC models for the projections to further examine variances arising from stochasticity. Notice there are at least 10 processes slower than the lag time $\tau$ used for RC learning, so a 2D FES projection in an RC space that sufficiently resolves all dynamics slower than $\tau$ should display at least 11 metastable basins separated by energy barriers. From these projections, we can see only $\bold r^{FMRC}$ and $\bold r^{SPIB}$, which optimize RC using probabilistic modelling, meet this criterion. In contrast, $\bold r^{TICA}_{2D}$ and $\bold r^{SRV}$ fail to distinguish many actual metastable states in such low-dimensional representations. This outcome is reasonable and expected, since the approximated eigenfunctions from $\hat \psi^{TICA}_3$ or $\hat \psi_3^{SRV}$, which characterize transitions between these ignored metastable states, were truncated. These results highlight the significant pitfalls of using VAC-based $\bold r^{TICA}$ or $\bold r^{VAC}$ as RC, as discussed earlier in the Introduction section.

To further clarify the physical significance of the two slowest modes $\psi_1$ and $\psi_2$, we use PCCA+\cite{roblitz2013fuzzy} to group the microstates clustered in $\bold r^{FMRC}$ space into three macrostates (Figure 5B, S4B, S5B \& S6B). The macrostates from PCCA+ are clustered in a way that microstates with similar $\hat \psi^{MSM}_i$ values are assigned to the same macrostate. This approach allows us to interpret the physical meaning of $\psi_1$ and $\psi_2$ by comparing the conformational ensembles of different PCCA+ macrostates. As shown in Figure S7, macrostate 1 represents the unfolded ensemble, while both macrostate 0 and macrostate 2 exhibit folded beta-hairpin structures with striking structural similarities. Upon close examination of their structural details (Figure S8), we found that all conformations in macrostate 0 have an inward-facing carbonyl oxygen on residue Glu5, whereas the same oxygen in macrostate 2 tends to face outward. Although macrostate 0 is indeed metastable, it is very rare, with a population of $\sim$0.2\% at equilibrium (Figure S7). In contrast, macrostate 2 is the true stable folded state, with a population of $\sim$78.2\% at equilibrium (Figure S7). Given that macrostate 0 is distant from the major folded state, we refer to it as a ``misfolded'' state. Despite its low population and striking structural similarity to the major folded state, correctly distinguishing this misfolded state from other metastable states is essential to correctly characterize the second slow mode of the system $\psi_2$. To demonstrate this, we projected the PCCA+ assignments learned from the $\bold r^{FMRC}$ space to one best and one worst RC space from each algorithm based on $\hat \lambda_i^{MSM}$ (Figure S4, S5 and S6). In general, MSMs with low $\hat \lambda_2^{MSM}$ values are built from RC spaces that cannot effectively distinguish this misfolded state. We further observed that while two-dimensional TICA and SRV struggle to differentiate this misfolded state, SPIB can occasionally do so. However, FMRC consistently differentiates this state, even in the worst-case scenario (Figure S9B).

Furthermore, by comparing FES projected on $\bold r^{FMRC}$ and the PCCA+ macrostates assignment (Figure 5), we observed that the projection onto $\bold r^{FMRC}$ reveals a highly detailed network structure within each macrostate, with smaller substates clearly delineated by small energy barriers. However, such a microscopic view may introduce redundant information and pose challenges for macroscopic interpretation from the projected FES. Beyond the PCCA+ clustering method, we found that tuning the learning lag time $\tau$ provides an alternative effective way to generate a macroscopic (or an even more microscopic) view (Figure S10 \& S11). When training with different $\tau$ values, without altering other hyperparameters, FMRC consistently reveals more microscopic substates at shorter $\tau$ (Figure S10), while gradually merging them as $\tau$ increases (Figure S11). Notably, even as substates merge, FMRC consistently distinguishes the sparsely populated misfolded state (Figure S11). As a result, the quality of the MSM constructed in these $\bold r^{FMRC}$ spaces learned with different $\tau$ remains largely consistent across a range of $\tau$ (Figure S12), with the exception of very short $\tau$. When learning $\bold r^{FMRC}$ with less training data, it seems that a high-quality $\bold r^{FMRC}$ can already be obtained using a much smaller training dataset (Figure S13). These results demonstrate the robustness of the FMRC algorithm in downstream analysis.  

\subsection{Complexity and Transition Networks: FMRC Analysis of Trp-Cage}

We next proceeded to test FMRC on a more complex example: a 208 $\mu$s trajectory of the larger mini-protein Trp-Cage (DAYAQWLADGGPSSGRPPPS, 20 amino acids, PDB ID: 2JOF) folding simulated at 290K. We used a similar feature selection strategy to include extensive interatomic distances and dihedral angles as our input features. For this analysis, we used a slightly longer lag time of $\tau=10$ ns compared to CLN025. As with the previous example, we initially conducted a regular TICA+MSM analysis to obtain an overall view of the system (Figure S14). The results suggest that, although Trp-Cage is only twice as long in sequence as CLN025, its transition network is significantly more complex, with a greater number of slow processes present in the system. As a result, at an MSM lag time $\tau_{MSM} = 50$ ns, there was no clear spectral gap among the top 20 eigenvalues. For this comparison between different RCs, we chose to focus on the top 11 slowest processes.

\begin{figure}[h]
    \centering
    \includegraphics[scale=0.65]{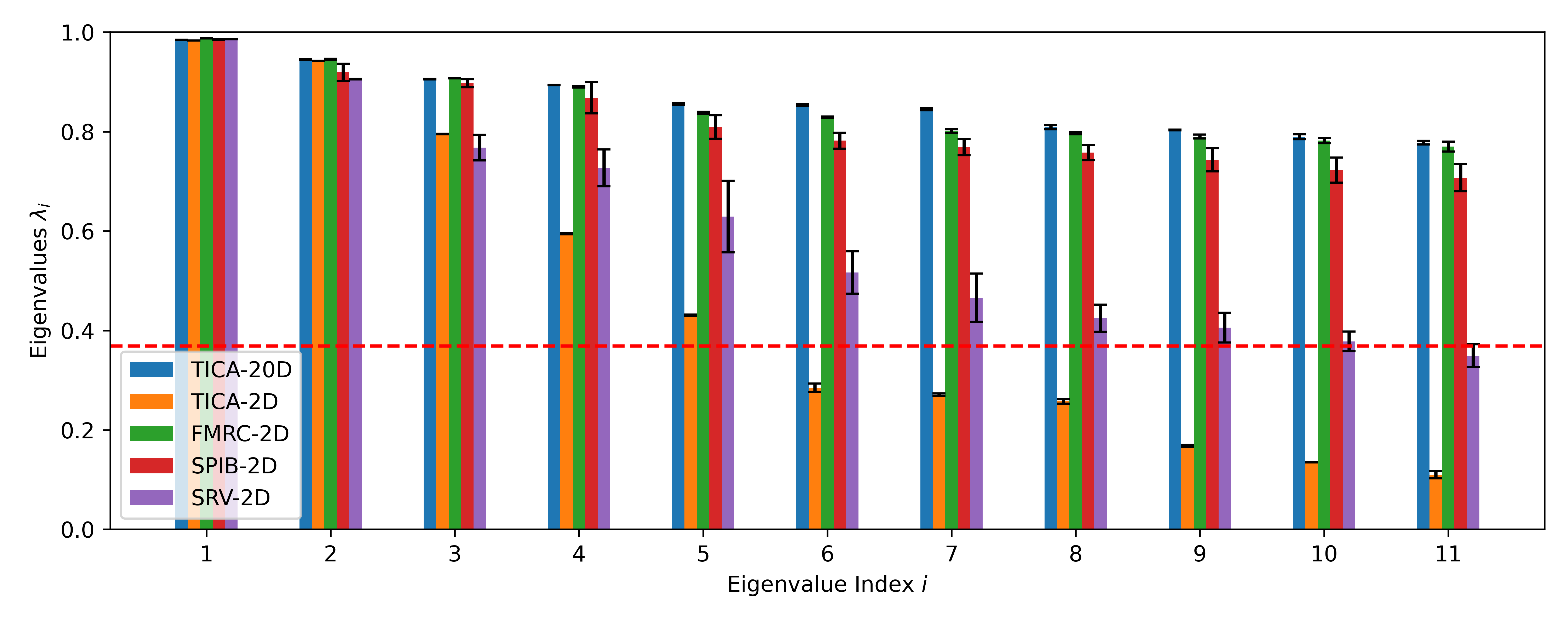}
    \caption{Comparison of $\hat \lambda_i^{MSM}$ from MSMs constructed in different RC spaces for Trp-Cage. A red dashed line at $\lambda_i = 0.369$ has been drawn to denote a cutoff for the corresponding timescales lower than the $\tau^{MSM}$ for MSM construction. This indicates that the constructed MSM has failed to identify this slow process.}
    \label{fig:6}
\end{figure}

After testing for a range of $d_{int}$, we selected $d_{int}=20$ as the truncation dimension for $\boldsymbol \Psi_{TICA}$. The resulting $\boldsymbol \Psi_{TICA} = [\hat \psi_1^{TICA},...,\hat \psi_{20}^{TICA}]^\top$ was used as the input for our FMRC neural network, following the exact same training protocol and hyperparameters as those used in the CLN025 example (see Supporting Information). Similarly, multiple models from repeated training were used to calculate mean and standard deviation of $\hat \lambda_i^{MSM}$ for the MSMs built in each RC space spanned by $\bold r^{TICA}_{20D}$, $\bold r^{TICA}_{2D}$, $\bold r^{SRV}$,$\bold r^{FMRC}$ or $\bold r^{SPIB}$ (Figure 6). 

Despite the increased complexity of the system, $\bold r^{FMRC}$ consistently encodes a similar amount of the system's major slow dynamics as a 20-dimensional $\bold r^{TICA}_{20D}$ space, with minimal training variance, showing only a slight decay in the eigenvalues of the 5th to 9th MSM eigenvectors. In contrast, the VAC-based two-dimensional RCs $\bold r^{TICA}_{2D}$ and $\bold r^{SRV}$ exhibit a significant drop in performance starting from $\hat \lambda_3^{MSM}$. Specifically, $\bold r^{TICA}_{2D}$ fails to identify the slow processes after $\hat \psi_5^{MSM}$. While introducing a neural network transformation to the input features significantly improves the performance of $\bold r^{SRV}$, the performances are still far from optimal, and the addition of the neural network introduces significant training variance. 

On the other hand, $\bold r^{SPIB}$ performs more satisfactorily, preserving the majority of the system dynamics. Yet, $\bold r^{SPIB}$ is consistently outperformed by $\bold r^{FMRC}$ in characterizating every slow mode, with greater training variance. These results are consistent with those observed in the CLN025 example, further underpinning the superior performance of $\bold r^{FMRC}$ in encoding slow dynamics within its low-dimensional RC space. 

As the system complexity increases, we observed that the ITS of MSM eigenvectors constructed in the $\bold r^{FMRC}$ space does not always converge with the shortest $\tau$ (Figure S15). Nevertheless, at $\tau_{MSM}=50$ ns, as demonstrated by the Chapman-Kolmogorov test, the MSM constructed in $\bold r^{FMRC}$ can provide very accurate predictions for almost all future macrostate populations (Figure S16).

\begin{figure}[h]
    \centering
    \includegraphics[scale=0.5]{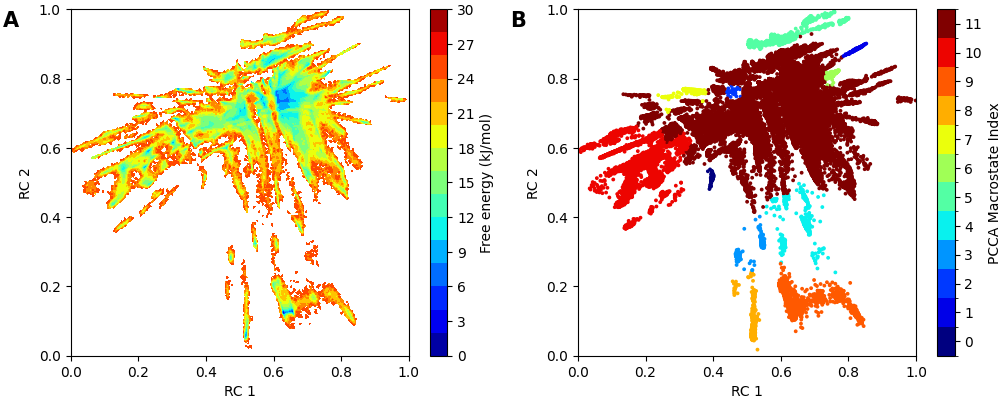}
    \caption{(A) The Trp-Cage 2D FES projection and (B) the Trp-Cage PCCA+ macrostate assignment projection on the normalized $\bold r^{FMRC}$ space where the best MSM was constructed.}
    \label{fig:7}
\end{figure}

We next examined the 2D FES and PCCA+ assignments projections in these 2D RC spaces (Figure 7, Figure S17, Figure S18 \& Figure S19). Given the smaller variances in $\hat \lambda_i^{MSM}$, we focused only on the projections in the best RC spaces with the highest $\hat \lambda_i^{MSM}$. Consistent with the observations in CLN025, the VAC-based $\bold r^{TICA}_{2D}$ and $\bold r^{SRV}$ display limited resolution between substates within the major metastable basins and are unable to distinguish several metastable states. In contrast, both $\bold r^{SPIB}$ and $\bold r^{FMRC}$ provide a more detailed description of the transition network. The PCCA+ projection on $\bold r^{FMRC}$ reveals multiple folded and misfolded states interconnected with a major unfolded state. Specifically, the native nuclear magnetic resonance (NMR) structure belongs to the macrostate 9, which contains multiple deepest basins and corresponds to the folded state (Figure S21). $\bold r^{FMRC}$ further distinguishes several structurally similar but less stable misfolded states, which are interconnected through the major unfolded state, macrostate 11 (Figure S20 \& S21). Such metastable state assignment can be distinguished by $\bold r^{SPIB}$ in a satisfactory way, although there are still some overlappings (Figure S19). This overlap might explain why the $\hat \lambda_i^{MSM}$ values for the MSMs constructed from $\bold r^{SPIB}$ are generally lower than those constructed from $\bold r^{FMRC}$.

Finally, $\bold r^{FMRC}$ can be adjusted to provide either a more microscopic or macroscopic description within its RC space by tuning the training lag time $\tau$ (Figure S22 \& S23). However, as the complexity of the system increases, we notice that the final quality of the constructed MSM becomes more sensitive to the choice of $\tau$ (Figure S24). In general, adjusting $\tau$ has a smaller impact on the quality of the leading eigenvectors. This suggests that the choice of $\tau$ for learning $\bold r^{FMRC}$ should be done with greater care, particularly if PCCA+ is used to cluster a large number of macrostates for a more detailed microscopic interpretation of the system. Similarly, the increased complexity also increases the need of a higher amount of training data for high quality $\bold r^{FMRC}$ learning (Figure S25).

\subsection{Challenging System Size: FMRC Performance on NTL-9 Folding Dynamics}

For our final example, we used a 1.11 ms trajectory of NTL-9 (MKVIFLKDVKGMGKKGEIKNVADGYANNFLFKQGLAIEA, 39 amino acids, PDB ID:2HBA) folding simulated at 355K. This system, with almost double the protein sequence size compared to Trp-Cage, presents a further challenge for our FMRC learning algorithm. To manage the increased system size, we reduced the input features by excluding all sidechain and backbone $N-O$ interatomic distances (see Supporting Information for details). Furthermore, we strided the trajectory at a frequency of 5 to fit the dataset into memory. We used a lag time of $\tau=20$ ns for all model training. As with the previous example, we began with a TICA+MSM analysis to obtain an overall picture of the system (Figure S26). Similar to Trp-Cage, NTL-9 exhibited numerous slow processes. However, our analysis suggested that a longer lag time of $\tau^{MSM}=300$ ns was necessary for MSM construction to ensure Markovianity for all MSMs constructed in different RC spaces. We found that only 8 processes were slower than $\tau^{MSM}$, and there was no clear spectral gap among the first 8 eigenvalues (Figure S26). Therefore, we focused on the top 5 slowest processes of the system for further analysis.

\begin{figure}[h]
    \centering
    \includegraphics[scale=0.75]{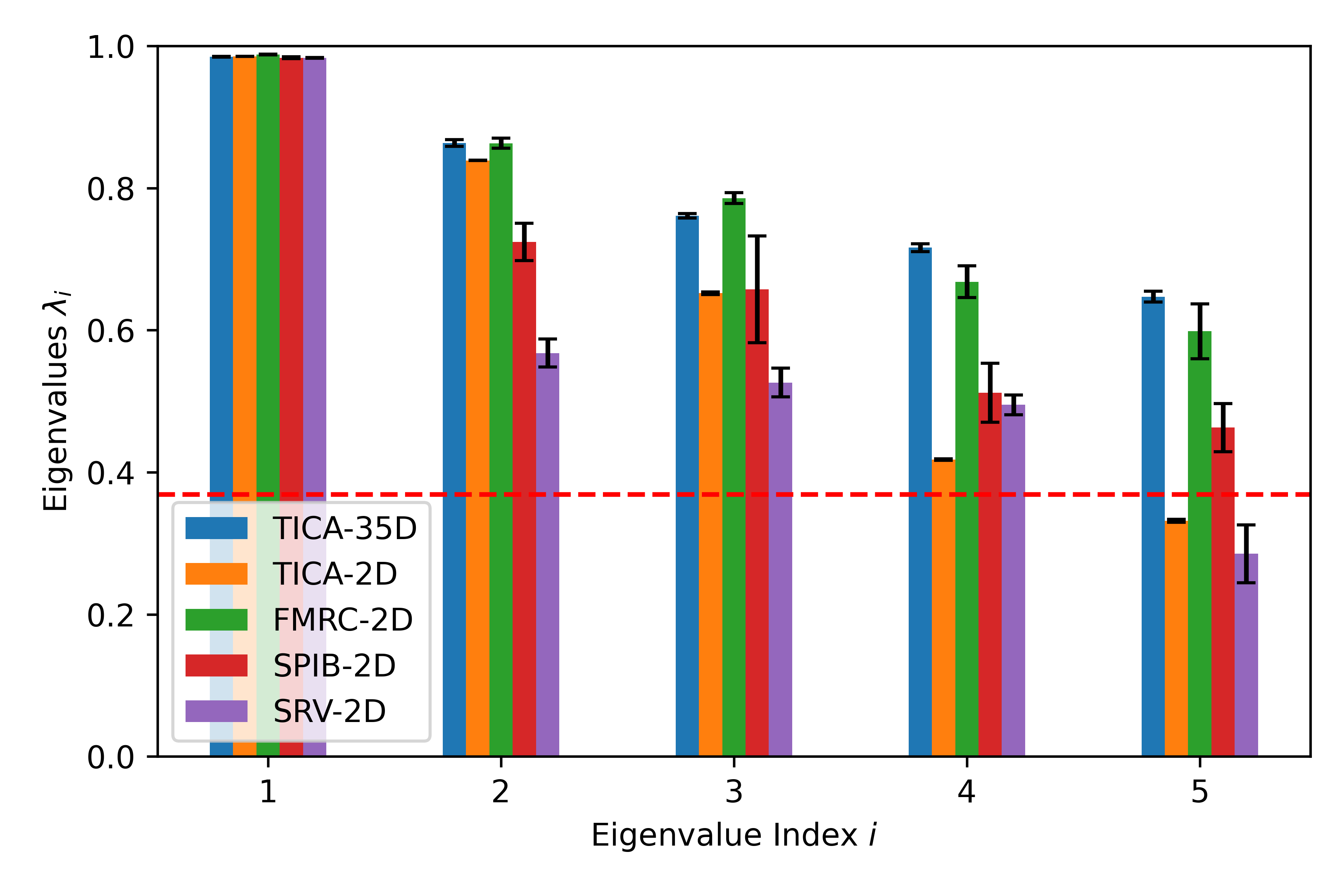}
    \caption{Comparison of $\hat \lambda_i^{MSM}$ of MSMs constructed in different RC spaces for NTL-9. A red dashed line at $\lambda_i = 0.369$ has been drawn to denote a cutoff for the corresponding timescales lower than the $\tau^{MSM}$ for MSM construction. This indicates that the constructed MSM has failed to identify this slow process.}
    \label{fig:8}
\end{figure}

We followed the same procedures as in CLN025 and Trp-Cage and selected $d_{int}=35$. $\boldsymbol \Psi_{TICA} = [\hat \psi_1^{TICA},...,\hat \psi_{35}^{TICA}]^\top$ was then used as our FMRC neural network input and we used the same training hyperparameters as in CLN025 and Trp-Cage. Means and standard deviations were also obtained in similar procedures and the results are summarized in Figure 8. We can see that $\bold r^{FMRC}$ similarly outperforms $\bold r^{TICA}_{2D}$, $\bold r^{SRV}$ and $\bold r^{SPIB}$, while being able to match the performance of a 35-dimensional $\bold r^{TICA}_{35D}$. Specifically, $\bold r^{FMRC}$ retains the system dynamics inside its RC space with only some slight decay in $\hat \lambda_4^{MSM}$ and $\hat \lambda_5^{MSM}$. MSM constructed in $\bold r^{FMRC}$ also results in an improved estimation of $\hat \psi^{MSM}_3$ in comparison to the one constructed in $\bold r^{TICA}_{35D}$. In contrast, as the system complexity grows, the performance of $\bold r^{TICA}_{2D}$,$\bold r^{SRV}$ and $\bold r^{SPIB}$ significantly decay, and none of them are able to capture the slow processes $\psi_5$ in their low-dimensional RC spaces. In this particular example, only the MSM constructed in $\bold r^{TICA}_{35D}$ converges with a shorter $\tau$ and the rest of MSMs do not have a significant difference in ITS convergence (Figure S27). Similarly, MSM constructed from $\bold r^{FMRC}$ can provide a sufficiently accurate prediction of the future macrostate population (Figure S28).

\begin{figure}[h]
    \centering
    \includegraphics[scale=0.1]{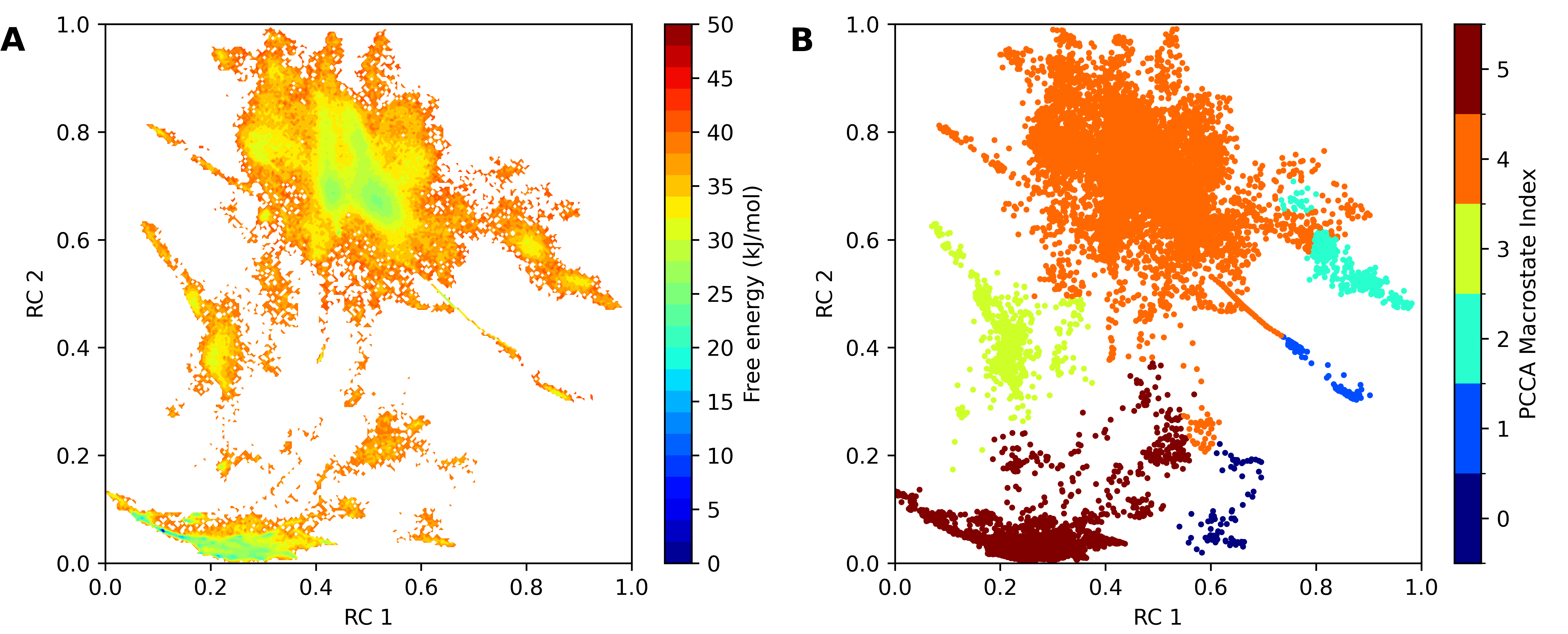}
    \caption{(A) The NTL-9 2D FES projection and (B) the NTL-9 PCCA+ macrostate assignment projection on the normalized $\bold r^{FMRC}$ space where the best MSM was constructed.}
    \label{fig:9}
\end{figure}

We further projected 2D FES and PCCA+ assignments as in Trp-Cage using the best MSM constructed in different RC spaces (Figure 9, Figure S29, Figure S30 \& Figure S31) and similar observations are also obtained. The 2D FES projection on $\bold r^{FMRC}$ consists of numerous small basins divided by small energy barriers (Figure 9A). The PCCA+ assignment projection on $\bold r^{FMRC}$ revealed a similar interconnected transition network between the major folded state (macrostate 5) and the major unfolded state (macrostate 4), accompanied by several less populated misfolded states, similar to the network structure in Trp-Cage. Representative conformations from each metastable state are presented in Figure S32. Similar trends follow that metastable states are all well distinguished in $\bold r^{FMRC}$ and $\bold r^{SPIB}$ (Figure S31, albeit to a lesser extent) and cannot be fully distinguished in VAC-based $\bold r^{TICA}_{2D}$ or $\bold r^{SRV}$ (Figure S29 \& S30). By tuning $\tau$, a more microscopic or more macroscopic representation by $\bold r^{FMRC}$ can be similarly obtained (Figure S33 \& S34). Interestingly, with the even more increased complexity, increasing $\tau$ for learning $\bold r^{FMRC}$ actually further improves the quality of the constructed MSM (Figure S35), suggesting that the optimal choice of $\tau$ for learning $\bold r^{FMRC}$ for MSM construction is very system specific and requires careful tuning. Finally, it should be noticed that with such system complexity, the requirement of the amount of training data by FMRC is even higher (Figure S36).

\subsection{Bias Deposition Using $\bold r^{FMRC}$: A Preliminary Test on Ala2}

To further validate the effectiveness of $\bold r^{FMRC}$ for enhanced sampling bias deposition, we learned the $\bold r^{FMRC}$ of the backbone torsional angle isomerization of Ala2 in solvent from a 3.2 ns unbiased MD simulation dataset from our previous study \cite{zhang2024enhanced} (See Supporting Information). We then performed multiple-walkers on-the-fly probability enhanced sampling (OPES) simulations with $\bold r^{FMRC}$ using 16 parallel replicas for 3 ns, resulting in an aggregated OPES simulation time of 16*3=48 ns. We can see OPES simulations using $\bold r^{FMRC}$ learned from a relatively small unbiased simulation dataset can efficiently converge to an almost invariant FES estimate after only 0.9 ns sampling for each replica (Figure S37A \& S37B). Moreover, at the end of the OPES simulations, both the 1D FES estimates along $\phi$ (Figure S37C) and $\psi$ (Figure S37D) and the 2D FES estimates in the $\phi$-$\psi$ spaces (Figure S37F) show almost no difference in comparison to a 250 ns reference OPES simulation using $\phi$ and $\psi$ as the biasing RC (Figure S37E). These results from this preliminary test on Ala2 demonstrate the potential of $\bold r^{FMRC}$ in enhanced sampling bias deposition for more complex systems.   

\section{Conclusion}

Obtaining a low-dimensional RC representation of the system which effectively describes the full-dimensional dynamics of biomolecular MD trajectories is one of the key problems in the field of biomolecular simulations. Current methods, either require a supervised learning scheme (e.g. committor-based methods) that can potentially be misdirected by human inputs, or suffer from the linearity of the methods (VAC-based). In this article, we introduce a novel unsupervised RC machine learning algorithm, Flow Matching for Reaction Coordinates (FMRC), which is based on a novel definition of optimal RC of lumpability and decomposability from recent theoretical studies. The novel theoretical definition has a deep relationship with the well-established transfer operator theory and our reformulation of their mathematical definitions into conditional probability enables the incorporation of such definition as an optimization target for deep learning. In contrast, this new formulation of optimal RC avoids the linearity problem presented in the VAC-based method and can be effectively learned through probabilistic modelling. Specifically, by representing the optimal RC as an encoder and using the FM architecture as decoders, we efficiently perform a search for the optimal $\bold r^{FMRC}$ constructed from input features that satisfy lumpability and decomposability. We further demonstrate that $\bold r^{FMRC}$ can effectively compress and incorporate a large amount of dynamic information in a two-dimensional space which can only be effectively described with a much higher dimensional $\bold r^{TICA}$ coordinates for all three example systems with increased complexity. In contrast, the VAC-based methods including both TICA and SRV cannot achieve such efficient system compression. Specifically, while they are able to accurately capture the slowest or the second slowest processes of the system, they are not able to correctly encode the dynamics of other important processes inside a 2D RC space. This leads to an incorrect representation of the system dynamics and these ignored processes can become significant bottlenecks if such RC is used for bias deposition in enhanced sampling simulations. Moreover, FMRC has a much smaller training variance and provides a much more consistent result in comparison to all other tested methods. We also demonstrate that the training procedure is rather simple and a minimal number of hyperparameters require tuning.

Noticeably, by providing initial state labels obtained from the classic TICA+k-means protocol to the SPIB algorithm, the learned $\bold r^{SPIB}$ is able to provide a fair estimate of the dominant implied timescales and distinguish between the major metastable states for all systems. Interestingly, a recent study\cite{federici2023latent} has demonstrated that the mutual information shared by the time-lagged pairs $\{\bold x_t, \bold x_{t+\tau}\}$ is equivalent to the autocorrelation which the VAC-based methods aim to maximize. Considering the fact that both FMRC and the VAC-based methods similarly originate from the transfer operator theory, the theoretical link between FMRC and SPIB is worth exploring for sure. We leave the theoretical investigation of their relationship for future studies.

Considering the importance of RC in simulation studies of biomolecules, we anticipate the efficacy of the FMRC algorithm will make a difference in various downstream applications. Specifically, we would like to highlight their potential usage in MSM construction and collective variable (CV)-based enhanced sampling methods. For MSM construction, we demonstrate that FMRC can efficiently reduce the dimensions of the trajectory dataset and thus allow a faster clustering step at the cost of a slight decrease in MSM model quality in comparison to the conventional TICA dimension reduction. The resulting 2D projection inside the $\bold r^{FMRC}$ space is also much richer in information in comparison to a $\bold r^{TICA}_{2D}$ representation and much easier to interpret in comparison to a high-dimensional $\bold r^{TICA}$ representation. Such 2D projection can serve as an effective complement to the visualization of the transition network between e.g. PCCA+ macrostates. By performing a preliminary test of using $\bold r^{FMRC}$ as biasing RC for enhanced sampling simulation of Ala2, we demonstrate FMRC's ability to effectively encode large amounts of dynamics from a relatively small set of training data, making it a promising candidate method to alleviate the need for human expertise. However, there are two potential problems in such an application for larger systems: 1) although the theoretical studies on lumpability and decomposability suggest that $\bold r^{FMRC}$ can be computed with sparse sampling \cite{bittracher2023optimal}, the amount of data required to learn an accurate $\bold r^{FMRC}$ seems to increase rapidly as the system complexity increases. 2) In our current study, we used hundreds to thousands of input features and computed the $\bold r^{FMRC}$ with a relatively large neural network for larger systems. As a result, the effective learning of $\bold r^{FMRC}$ only requires limited human knowledge of the system as presented in this article. However, CV-based enhanced sampling methods often require iterative bias computation on-the-fly during the simulation and computing $\bold r^{FMRC}$ can significantly slow down the simulation. We leave the investigation of the application of $\bold r^{FMRC}$ and these considerations to future studies. 

\begin{acknowledgement}

The authors thank D.~E.~Shaw research for providing the ultralong MD simulation trajectories used in this study. We acknowledge funding from the National Natural Science Foundation of China (1217011218 to H.W., 32371300 to Y.W.). Y.W. acknowledges the Fundamental Research Funds for the Zhejiang Provincial Universities (226-2024-00137), and the Zhejiang Provincial National Science Foundation of China (No. LZ24C050003).

\end{acknowledgement}

\begin{suppinfo}
Further details on the previously established theories and algorithms, including VAC, TICA, SRV, SPIB and the details of model training, MSM construction and OPES simulations as well as additional figures supporting our results, can be found in the Supporting Information. 
This information is available free of charge via the Internet at http://pubs.acs.org.
\end{suppinfo}


\bibliography{reference}

\end{document}


\renewcommand{\thetable}{S\arabic{table}} 
\renewcommand{\thefigure}{S\arabic{figure}}
\renewcommand{\theequation}{S\arabic{equation}}

\section{Theoretical Analysis of Lumpability and Decomposability}

According to the definition of the optimal reaction coordinate by Bittracher et al. \cite{bittracher2023optimal}, if $\mathbf{r}^{LD}$ strictly satisfies the lumpability and decomposability conditions, there exist functions $p_\tau^L$ and $p_\tau^D$ such that the following equations hold:
\begin{eqnarray}
\text{Lumpability} & \to & p_{\tau}(\mathbf{x},\mathbf{y})=p_{\tau}^{L}(\mathbf r^{LD}(\mathbf{x}),\mathbf{y})\label{eq:original-l}\\
\text{Decomposability} & \to & p_{\tau}(\mathbf{x},\mathbf{y})=p_{\tau}^{D}(\mathbf{x},\mathbf r^{LD}(\mathbf{y}))\pi(\mathbf{y})\label{eq:original-d}
\end{eqnarray}
For the sake of completeness, we now demonstrate that the lumpability and decomposability conditions described in the main text (see equations 8, 9, and 11) are consistent with the definitions provided above. For further theoretical details, please refer to \cite{zhang2024flow}.

In this section, for simplicity in notation and derivations, we define $\Omega_{\mathbf r}=\{\mathbf y|\mathbf r^{LD}(\mathbf y)=\mathbf r\}$, and denote by $\pi_{\mathbf r}$ the stationary distribution of $\mathbf r^{LD}$. $\pi_{\mathbf r}$ can be equivalently expressed as
\[
\pi_{\mathbf r}(\mathbf{r})=\int\pi(\mathbf{x})\mathbb{P}(\mathbf{r}^{LD}(\mathbf{x}_{t+\tau})=\mathbf{r}|\mathbf{x}_{t}=\mathbf{x})\mathrm{d}\mathbf{x}
\]
and
\begin{eqnarray*}
\pi_{\mathbf r}(\mathbf{r}) & = & \int\pi(\mathbf{x})\mathbb{P}(\mathbf{r}^{LD}(\mathbf{x}_{t})=\mathbf{r}|\mathbf{x}_{t}=\mathbf{x})\mathrm{d}\mathbf{x}\\
 & = & \int\pi(\mathbf{x})\delta(\mathbf{r}-\mathbf{r}^{LD}(\mathbf{x}))\mathrm{d}\mathbf{x},
\end{eqnarray*}
where $\delta$ denotes the Dirac function.

\subsection{Equivalence between Equations \ref{eq:original-l} and 8}

It is trivial that equation 8 is a sufficient condition for equation S1 by setting $p_{\tau}^L(\mathbf r,\mathbf y)=\mathbb P(\mathbf x_{t+\tau}=\mathbf y|\mathbf r^{LD}(\mathbf x_t)=\mathbf r)$, and here we will only prove the necessity. For an arbitrary conformation $\mathbf x$ with $\mathbf r^{LD}(\mathbf x)=\mathbf r$, if equation S1 exactly holds, we have
\begin{eqnarray*}
\mathbb{P}\left(\mathbf{x}_{t+\tau}=\mathbf{y}|\mathbf{r}^{LD}(\mathbf{x}_{t})=\mathbf{r}\right) & = & \int\mathbb{P}\left(\mathbf{x}_{t+\tau}=\mathbf{y}|\mathbf{x}_{t}=\mathbf{x}',\mathbf{r}^{LD}(\mathbf{x}_{t})=\mathbf{r}\right)\mathbb{P}\left(\mathbf{x}_{t}=\mathbf{x}'|\mathbf{r}^{LD}(\mathbf{x}_{t})=\mathbf{r}\right)\mathrm{d}\mathbf{x}'\\
 & = & \int\mathbb{P}\left(\mathbf{x}_{t+\tau}=\mathbf{y}|\mathbf{x}_{t}=\mathbf{x}'\right)\mathbb{P}\left(\mathbf{x}_{t}=\mathbf{x}'|\mathbf{r}^{LD}(\mathbf{x}_{t})=\mathbf{r}\right)\mathrm{d}\mathbf{x}'\\
 & = & \int p_{\tau}^{L}(\mathbf{r},\mathbf{y})\mathbb{P}\left(\mathbf{x}_{t}=\mathbf{x}'|\mathbf{r}^{LD}(\mathbf{x}_{t})=\mathbf{r}\right)\mathrm{d}\mathbf{x}'\\
 & = & p_{\tau}^{L}(\mathbf{r},\mathbf{y})\cdot\int\mathbb{P}\left(\mathbf{x}_{t}=\mathbf{x}'|\mathbf{r}^{LD}(\mathbf{x}_{t})=\mathbf{r}\right)\mathrm{d}\mathbf{x}'\\
 & = & p_{\tau}^{L}(\mathbf{r},\mathbf{y})\cdot1\\
 & = & p_{\tau}(\mathbf{x},\mathbf{y}).
\end{eqnarray*}

\subsection{Equivalence between Equations \ref{eq:original-d} and 9}

We first prove that equation 9 is a sufficient condition for equation S2. When equation 9 holds exactly, based on the definition of the stationary distribution \( \pi \), we have
\begin{eqnarray*}
\pi(\mathbf{y}) & = & \int\pi(\mathbf{x})p_{\tau}(\mathbf{x},\mathbf{y})\mathrm{d}\mathbf{x}\\
 & = & \pi_{local}(\mathbf{r}^{LD}(\mathbf{y}),\mathbf{y})\int\pi(\mathbf{x})\mathbb{P}(\mathbf{r}^{LD}(\mathbf{x}_{t+\tau})=\mathbf{r}^{LD}(\mathbf{y})|\mathbf{x}_{t}=\mathbf{x})\mathrm{d}\mathbf{x}\\
 & = & \pi_{local}(\mathbf{r}^{LD}(\mathbf{y}),\mathbf{y})\pi_{\mathbf r}(\mathbf{r}^{LD}(\mathbf{y})).
\end{eqnarray*}
Thus, defining
\[
p_{\tau}^{D}(\mathbf{x},\mathbf{r}) =  \frac{\mathbb{P}(\mathbf{r}^{LD}(\mathbf{x}_{t+\tau})=\mathbf{r}|\mathbf{x}_{t}=\mathbf{x})}{\pi_{\mathbf r}(\mathbf r)},
\]
we obtain
\begin{eqnarray*}
p_{\tau}^{D}(\mathbf{x},\mathbf{r})\pi(\mathbf{y}) & = & \frac{\mathbb{P}(\mathbf{r}^{LD}(\mathbf{x}_{t+\tau})=\mathbf{r}|\mathbf{x}_{t}=\mathbf{x})\pi(\mathbf{y})}{\pi_{\mathbf r}(\mathbf{r})}\\
 & = & \mathbb{P}(\mathbf{r}^{LD}(\mathbf{x}_{t+\tau})=\mathbf{r}|\mathbf{x}_{t}=\mathbf{x})\pi_{local}(\mathbf{r}^{LD}(\mathbf{y}),\mathbf{y})\\
 & = & p_{\tau}(\mathbf{x},\mathbf{y}),
\end{eqnarray*}
which implies that equation \ref{eq:original-d} is satisfied.

Next, we prove the necessity. According to equation \ref{eq:original-d},
\begin{eqnarray*}
\mathbb{P}(\mathbf{r}^{LD}(\mathbf{x}_{t+\tau})=\mathbf{r}|\mathbf{x}_{t}=\mathbf{x}) & = & \int p_{\tau}(\mathbf{x},\mathbf{y})\cdot\mathbb{P}(\mathbf{r}^{LD}(\mathbf{x}_{t+\tau})=\mathbf{r}|\mathbf{x}_{t+\tau}=\mathbf{y})\mathrm{d}\mathbf{y}\\
 & = & \int p_{\tau}^{D}(\mathbf{x},\mathbf{r}^{LD}(\mathbf{y}))\pi(\mathbf{y})\cdot\delta(\mathbf{r}-\mathbf{r}^{LD}(\mathbf{y}))\mathrm{d}\mathbf{y}.
\end{eqnarray*}
Since $p_{\tau}^{D}(\mathbf{x},\mathbf{r}^{LD}(\mathbf{y}))\equiv p_{\tau}^{D}(\mathbf{x},\mathbf{r})$ for $\mathbf y\in\Omega_{\mathbf r}$, we can get
\begin{eqnarray*}
\mathbb{P}(\mathbf{r}^{LD}(\mathbf{x}_{t+\tau})=\mathbf{r}|\mathbf{x}_{t}=\mathbf{x}) & = & p_{\tau}^{D}(\mathbf{x},\mathbf{r})\int\pi(\mathbf{y})\delta(\mathbf{r}-\mathbf{r}^{LD}(\mathbf{y}))\mathrm{d}\mathbf{y}\\
 & = & p_{\tau}^{D}(\mathbf{x},\mathbf{r})\pi_{\mathbf r}(\mathbf{r}).
\end{eqnarray*}
Therefore,
\begin{eqnarray*}
p_{\tau}(\mathbf{x},\mathbf{y}) & = & p_{\tau}^{D}(\mathbf{x},\mathbf{r}^{LD}(\mathbf{y}))\pi(\mathbf{y})\\
 & = & \mathbb{P}\left(\mathbf{r}^{LD}(\mathbf{x}_{t+\tau})=\mathbf{r}^{LD}(\mathbf{y})|\mathbf{x}_{t}=\mathbf{x}\right)\pi_{local}(\mathbf{r}^{LD}(\mathbf{y}),\mathbf{y})
\end{eqnarray*}
with $\pi_{local}(\mathbf{r},\mathbf{y})=\pi(\mathbf y)/\pi_{\mathbf r}(\mathbf r)$.

Additionally, it is worth noting that if the process \( \{\mathbf{x}_t\} \) reaches equilibrium with \( \mathbb{P}(\mathbf{x}_{t+\tau} = \mathbf{y}) = \pi(\mathbf{y}) \) and \( \mathbb{P}(\mathbf{r}^{LD}(\mathbf{x}_{t+\tau}) = \mathbf{r}) = \pi_{\mathbf r}(\mathbf{r}) \), it follows that
\begin{eqnarray*}
\mathbb{P}\left(\mathbf{x}_{t+\tau} = \mathbf{y} \mid \mathbf{r}^{LD}(\mathbf{x}_{t+\tau}) = \mathbf{r}\right) & = & \frac{\mathbb{P}\left(\mathbf{r}^{LD}(\mathbf{x}_{t+\tau}) = \mathbf{r} \mid \mathbf{x}_{t+\tau} = \mathbf{y}\right) \mathbb{P}(\mathbf{x}_{t+\tau} = \mathbf{y})}{\mathbb{P}(\mathbf{r}^{LD}(\mathbf{x}_{t+\tau}) = \mathbf{r})} \\
& = & \frac{\delta(\mathbf{r} - \mathbf{r}^{LD}(\mathbf{y})) \pi(\mathbf{y})}{\pi_{\mathbf r}(\mathbf{r})} \\
& = & \delta(\mathbf{r} - \mathbf{r}^{LD}(\mathbf{y})) \pi_{\text{local}}(\mathbf{r}, \mathbf{y}),
\end{eqnarray*}
which demonstrates that \( \pi_{\text{local}} \) characterizes the local stationary distribution within the level set \( \Omega_{\mathbf r} \).

\subsection{Equivalence between Equations 11B and 9}
If equation 9 holds, we can derive the following
\begin{eqnarray*}
p_{-\tau}(\mathbf{y},\mathbf{x}) & = & \frac{\mathbb{P}(\mathbf{x}_{t}=\mathbf{x})p_{\tau}(\mathbf{x},\mathbf{y})}{\int\mathbb{P}(\mathbf{x}_{t}=\mathbf{x}')p_{\tau}(\mathbf{x}',\mathbf{y})\mathrm{d}\mathbf{x}'}\\
 & = & \frac{\mathbb{P}(\mathbf{x}_{t}=\mathbf{x})\mathbb{P}\left(\mathbf{r}^{LD}(\mathbf{x}_{t+\tau})=\mathbf{r}^{LD}(\mathbf{y})|\mathbf{x}_{t}=\mathbf{x}\right)\pi_{local}(\mathbf{r}^{LD}(\mathbf{y}),\mathbf{y})}{\int\mathbb{P}(\mathbf{x}_{t}=\mathbf{x}')\mathbb{P}\left(\mathbf{r}^{LD}(\mathbf{x}_{t+\tau})=\mathbf{r}^{LD}(\mathbf{y})|\mathbf{x}_{t}=\mathbf{x}'\right)\pi_{local}(\mathbf{r}^{LD}(\mathbf{y}),\mathbf{y})\mathrm{d}\mathbf{x}'}\\
 & = & \frac{\mathbb{P}(\mathbf{x}_{t}=\mathbf{x})\mathbb{P}\left(\mathbf{r}^{LD}(\mathbf{x}_{t+\tau})=\mathbf{r}^{LD}(\mathbf{y})|\mathbf{x}_{t}=\mathbf{x}\right)}{\int\mathbb{P}(\mathbf{x}_{t}=\mathbf{x}')\mathbb{P}\left(\mathbf{r}^{LD}(\mathbf{x}_{t+\tau})=\mathbf{r}^{LD}(\mathbf{y})|\mathbf{x}_{t}=\mathbf{x}'\right)\mathrm{d}\mathbf{x}'}\\
 & = & \frac{\mathbb{P}\left(\mathbf{x}_{t}=\mathbf{x},\mathbf{r}^{LD}(\mathbf{x}_{t+\tau})=\mathbf{r}^{LD}(\mathbf{y})\right)}{\int\mathbb{P}\left(\mathbf{x}_{t}=\mathbf{x},\mathbf{r}^{LD}(\mathbf{x}_{t+\tau})=\mathbf{r}^{LD}(\mathbf{y})\right)\mathrm{d}\mathbf{x}'}\\
 & = & \mathbb{P}\left(\mathbf{x}_{t}=\mathbf{x}|\mathbf{r}^{LD}(\mathbf{x}_{t+\tau})=\mathbf{r}^{LD}(\mathbf{y})\right).
\end{eqnarray*}
So equation 9 is the sufficient condition for equation 11A.

Assuming that 11A holds, we have
\begin{eqnarray*}
p_{\tau}(\mathbf{x},\mathbf{y}) & = & \frac{\mathbb{P}(\mathbf{x}_{t+\tau}=\mathbf{y})\mathbb{P}\left(\mathbf{x}_{t}=\mathbf{x}|\mathbf{x}_{t+\tau}=\mathbf{y}\right)}{\mathbb{P}(\mathbf{x}_{t}=\mathbf{x})}\\
 & = & \frac{\mathbb{P}(\mathbf{x}_{t+\tau}=\mathbf{y})\mathbb{P}\left(\mathbf{x}_{t}=\mathbf{x}|\mathbf{r}^{LD}(\mathbf{x}_{t+\tau})=\mathbf{r}^{LD}(\mathbf{y})\right)}{\mathbb{P}(\mathbf{x}_{t}=\mathbf{x})}\\
 & = & \frac{\mathbb{P}(\mathbf{x}_{t+\tau}=\mathbf{y})\mathbb{P}\left(\mathbf{x}_{t}=\mathbf{x},\mathbf{r}^{LD}(\mathbf{x}_{t+\tau})=\mathbf{r}^{LD}(\mathbf{y})\right)}{\mathbb{P}(\mathbf{x}_{t}=\mathbf{x})\mathbb{P}\left(\mathbf{r}^{LD}(\mathbf{x}_{t+\tau})=\mathbf{r}^{LD}(\mathbf{y})\right)}\\
 & = & \mathbb{P}\left(\mathbf{r}^{LD}(\mathbf{x}_{t+\tau})=\mathbf{r}^{LD}(\mathbf{y})|\mathbf{x}_{t}=\mathbf{x}\right)\cdot\frac{\mathbb{P}(\mathbf{x}_{t+\tau}=\mathbf{y})}{\mathbb{P}\left(\mathbf{r}^{LD}(\mathbf{x}_{t+\tau})=\mathbf{r}^{LD}(\mathbf{y})\right)},
\end{eqnarray*}
which implies that equation 9 is satisfied with $\pi_{local}(\mathbf r,\mathbf y)=\frac{\mathbb{P}(\mathbf{x}_{t+\tau}=\mathbf{y})}{\mathbb{P}\left(\mathbf{r}^{LD}(\mathbf{x}_{t+\tau})=\mathbf{r}\right)}$. Hence, equation 9 is also a necessary condition for equation 11A.

\section{Variational Approach to Conformational Dynamics (VAC): Data-driven Approximation of Eigenfunctions $\psi_i$}

Following the transfer operator theory in the main text, a low-dimensional representation of the system dynamics can be obtained through a finite-rank approximation of the $\mathcal T(\tau)$. Notice that from main text equations 4, 6 and 7 we can re-write equation 1 into:
\begin{equation}\label{eq:spectral-decomposition}
    \mathcal T(\tau)u_t(\mathbf y)= 1 + \sum^\infty_{i=1}\exp(-\frac{\tau}{t_i})\braket{u_t,\psi_i}_\mu\psi_i
\end{equation}
where $\braket{u_t,\psi_i}_\mu = \int_\Omega \pi(\mathbf x)u_t(\mathbf x)\psi_i(\mathbf x) \mathbf d\mathbf x$. In general, reversible biomolecular dynamics only has a small finite number of slow orthogonal processes. Therefore, the infinite summation in the second component of equation S3 can be further divided into two parts, representing the slow and fast processes of the system. Since $\lambda_i$ for fast processes are small, we can truncate the series by keeping only the slow part to obtain a low-dimensional representation of the original system:
\begin{equation} 
\begin{split}
\mathcal T(\tau)u_t(\mathbf y) &=1 +\sum_{i=1}^d\exp(-\frac{\tau}{t_i})\braket{u_t,\psi_i}_\mu\psi_i+\mathcal T_{fast}(\tau)u_t(\mathbf y) \\ &\approx1 + \sum_{i=1}^d\exp(-\frac{\tau}{t_i})\braket{u_t,\psi_i}_\mu\psi_i
\end{split}
\end{equation}
where $d$ is the number of slow orthogonal processes present in the system. This can be chosen as the number of processes before a ``spectral gap'' where the eigenvalue before the gap is significantly larger than the eigenvalue after the gap. Thus, if $\mathcal T(\tau)$ of the system is available, to parameterize an optimal RC for this low-dimensional representation, one can trivially define an $d$-dimensional RC $\mathbf r^{VAC}:=\mathbf r^{VAC}(\mathbf x)$ as the leading eigenfunctions:
\begin{equation} 
\mathbf r^{VAC} = [\psi_1,\psi_2,...,\psi_d]^\top 
\end{equation}
However, neither $\mathcal T(\tau)$ nor $\psi_i$ are analytically tractable for complex systems such as all-atom explicit solvent biomolecular simulations and one currently can only obtain $\mathbf r^{VAC}$ by data-driven approximation, based on VAC\cite{noe2013variational}.

VAC suggests that although $\psi_i$ cannot be analytically obtained, one can use a variational approach to approximate $\psi_i$ in a data-driven manner. Suppose the exact solution of $\psi_i$ is given, it can be shown that the autocorrelation function $acf(\psi_i;\tau)$ of this eigenfunction is its associated eigenvalue:
\begin{equation} 
acf(\psi_i;\tau) = \mathbb E[\psi_i(\mathbf x_t)\psi_i(\mathbf x_{t+\tau})] = \lambda_i
\end{equation}
To approximate the first non-stationary eigenfunction $\psi_1$, we can propose a trial solution $\hat \psi_1 \approx \psi_1$. VAC shows that if $\hat \psi_1$ is normalized so that $\braket{\hat \psi_1,\hat \psi_1}_\mu = 1$ and orthogonal to the stationary $\psi_0$, its autocorrelation function $acf(\hat \psi_i;\tau)$ or associated eigenvalue $\hat \lambda_i$ is bounded above from the eigenvalue of the actual eigenfunction $\psi_1$:
\begin{equation} 
acf(\hat \psi_1;\tau) = \mathbb E[\hat \psi_1(\mathbf x_t)\hat \psi_1( \mathbf x_{t+\tau})]
= \hat \lambda_1\leq\lambda_1
\end{equation}
The final inequality is only satisfied when $\hat \psi_1 = \psi_1$. VAC also shows that a similar relationship can be equivalently derived for all subsequent trial eigenfunctions $\hat \psi_i$ as long as the trial functions $\hat \psi_i$ is similarly normalized and orthogonal to its previous eigenfunction $\hat \psi_{i-1}$:
\begin{equation} 
acf(\hat \psi_i;\tau) = \mathbb E[\hat \psi_i(\mathbf x_t)\hat \psi_i( \mathbf x_{t+\tau})]
= \hat \lambda_i\leq\lambda_i
\end{equation}
Such a relationship reaches an important conclusion: the higher the eigenvalues of the approximate eigenfunctions $\hat \psi_i$ we propose, the better it represents the system actual dynamics. Thus, one can solve the following optimization problem to numerically approximate any eigenfunction $\psi_i$ of the system, given sufficient MD data of the system:
\begin{equation} 
\psi_i \approx \arg\max_{\hat \psi_i}\mathbb E[\hat \psi_i(\mathbf x_t)\hat \psi_i( \mathbf x_{t+\tau})]
\end{equation}
subjects to the normalization condition $\braket{\hat \psi_i,\hat \psi_i}_\mu = 1$ and orthogonality $\braket{\hat \psi_i,\hat \psi_{i-1}} = 0$. Since $\lambda_i$ is related to the implied timescale of the process through main text equation 6, the physical intuition of this autocorrelation/eigenvalue maximization is that we are seeking the slowest modes of the system. For a comprehensive derivation of the algorithm, we refer readers to the original article \cite{noe2013variational}.

Over the years, multiple optimization algorithms have been proposed and successfully applied to analyze biomolecular simulations. In the following, we would like to briefly introduce two representative algorithms from this class, known as TICA\cite{perez2013identification} and SRV\cite{chen2019nonlinear}, which are compared to our algorithm FMRC for their capability in computing optimal RC. We also briefly explain the MSM\cite{prinz2011markov} approach as a special case of VAC.

\subsection{Time-lagged Independent Component Analysis (TICA) and State-Free Reversible VAMPnets (SRV)}

Given an MD trajectory $[\boldsymbol \chi_{t=0},\boldsymbol \chi_{t=1},...,\boldsymbol \chi_{t=T}]$, TICA seeks the eigenfunction approximation $\hat \psi^{TICA}_i$ as a linear combination of mean-free basis functions $\boldsymbol \chi$ spanned by this feature space according to equation S7:
\begin{equation} 
\hat \psi_i^{TICA} = \mathbf b_i^\top\boldsymbol \chi = \sum^l_{i=1}b_i\chi_i
\end{equation}
where $\mathbf b_i$ are the combination coefficients for the $i$th TICA eigenvector $\hat \psi_i^{TICA}$. This is achieved by the a two-step process given a lag time $\tau$:

\begin{enumerate}
    \item Slice the featurized trajectory into time-lagged pairs $\{\boldsymbol \chi_t,\boldsymbol \chi_{t+\tau}\}$ and compute the covariance matrix $\mathbf C(0)$ along with the autocorrelation matrix $\mathbf C(\tau)$:
    		\begin{equation}
    			\mathbf C(0) = \boldsymbol \chi_t^\top\boldsymbol \chi_t
    		\end{equation}
    		\begin{equation}
    			\mathbf C(\tau) = \boldsymbol \chi_t^\top \boldsymbol \chi_{t+\tau}
    		\end{equation}
    	\item Solve the generalized eigenvalue problem
    		\begin{equation}
    			\mathbf C(\tau)\mathbf B=\mathbf C(0)\mathbf B\mathbf \Lambda
    		\end{equation}
    		where $\mathbf B = [\mathbf b_1,...,\mathbf b_l]^\top$ is a matrix containing all combination coefficients and $\boldsymbol \Lambda$ is a diagonal matrix of eigenvalues $\boldsymbol \Lambda = diag(\hat \lambda_1^{TICA},...,\hat \lambda_l^{TICA})$. Notice that performing TICA in the mean-free space is equivalent to removing the stationary component $\hat \psi_0^{TICA},\hat \lambda_0^{TICA}$. 
\end{enumerate}

Theoretically, both $\mathbf C(0)$ and $\mathbf C(\tau)$ computed from reversible dynamics satisfying main text equation 3 should always be symmetric. However, in practice, this cannot be always fulfilled due to statistical noise or insufficient sampling. As a result, $\lambda_i$ in $\boldsymbol \Lambda$ may contain imaginary eigenvalues that are not physically well-defined. To address this issue, a force-symmetry procedure is commonly applied for $\mathbf C(0)$ and $\mathbf C(\tau)$ in equations S9 and S10:
\begin{equation}
\mathbf C_{sym}(0) = \frac{1}{2}(\boldsymbol \chi_t^\top\boldsymbol \chi_t + {\boldsymbol \chi_{t+\tau}}^\top \boldsymbol \chi_{t+\tau})
\end{equation}
\begin{equation}
\mathbf C_{sym}(\tau) = \frac{1}{2}(\boldsymbol \chi_t^\top \boldsymbol \chi_{t+\tau} +{\boldsymbol \chi_{t+\tau}}^\top \boldsymbol \chi_{t})
\end{equation}
This will introduce a bias when short off-equilibrium trajectory data are used \cite{wu2017variational}. Nevertheless, since we only use ultra long trajectories for analysis, we use this strategy for TICA throughout the paper, assuming the bias is limited.

The efficacy of TICA in approximating $\psi_i$ is heavily reliant on human expertise in the initial feature selection. To alleviate this reliance, Ferguson et al.\cite{chen2019nonlinear} proposed SRV, which can be seen as a neural network-based version of TICA. Instead of seeking the linear combination inside the feature space directly, SRV utilizes a neural network to learn a transformation $\boldsymbol \Xi^{SRV}_{nn,\boldsymbol \theta}(\boldsymbol \chi)$ of the input feature space where a TICA model with the highest eigenvalues $\lambda_i$ can be constructed. Mathematically, the eigenfunction $\hat \psi_i^{SRV}$ approximated by SRV is
\begin{equation}
\hat \psi_i^{SRV} = \mathbf b_i^\top\boldsymbol \Xi^{SRV}_{nn,\boldsymbol \theta}(\boldsymbol \chi) =\sum^d_{i=1}b_i \boldsymbol \Xi^{SRV}_{nn,\boldsymbol \theta}(\chi_i)
\end{equation}
where $d$ is the size of the final output layer of the neural network transformation $\boldsymbol \Xi^{SRV}_{nn,\boldsymbol \theta}(\boldsymbol \chi)$, representing the number of eigenfunctions to approximate. The same eigenvalue decomposition is applied to the neural network transformed basis sets as in equations S9, S10 and S11 to yield the combining coefficients $\mathbf b_i$. The transformation, on the other hand, is optimized together with the final eigenvalue decomposition by minimizing the following loss function:
\begin{equation}
\mathcal L_{SRV}(\boldsymbol \theta) = -\sum^d_{i=0}\hat \lambda_{i}^r
\end{equation}
where $\hat \lambda_i$ are the eigenvalues associated $\hat \psi_i^{SRV}$ with and $r = 2$ is commonly used, as well as in this article, so that the distance in the eigenfunction space of final SRV coordinates represents a physically meaningful kinetic distance \cite{noe2015kinetic}. Notice that this loss function is equivalent to equation S7 based on the relationship between autocorrelation and eigenvalues from equation S4. For algorithmic details, we refer the interested readers to the original article \cite{chen2019nonlinear}. 

Following equation S3, the TICA RC $\mathbf r^{TICA} := \mathbf r^{TICA}(\mathbf x)$ and SRV RC $\mathbf r^{SRV}:= \mathbf r^{SRV}(\mathbf x)$ referred in this article are thus similarly defined as $d$ leading approximated eigenfunctions by TICA or SRV:
\begin{equation}
\mathbf r^{TICA} = [\hat \psi_1^{TICA},...,\hat \psi_d^{TICA}]^\top 
\end{equation}
\begin{equation}
\mathbf r^{SRV} = [\hat \psi_1^{SRV},...,\hat \psi_d^{SRV}]^\top
\end{equation}
where $d$ is the number of dimensions after truncation as in equation S2.

\subsection{Markov State Model is a Special Case under VAC}

MSM\cite{prinz2011markov} is a well-established modelling technique for molecular kinetics and thermodynamics analysis. In brief, the goal of MSM is to construct a transition probability matrix $\hat {\mathbf T}(\tau)$ that is a direct approximation to the system transfer operator $\mathcal T(\tau)$. This is usually achieved by a Voronoi tessellation of the state space $\Omega$ into conformational ``microstates'' $S_i$, followed by estimating the probability of transition from a microstate to another microstate after a lag time of $\tau$. Notice that $\hat {\mathbf T}(\tau)$, similar to $\mathcal T(\tau)$, can be decomposed into eigenvalues $\hat \lambda_i^{MSM}$ and eigenvectors $\hat \psi_i^{MSM}$ which are approximations to the real eigenfunctions $\psi_i$. Here, the computed MSM eigenvectors $\hat \psi_i^{MSM}$ can be seen as a model constructed from a basis function set of indicator functions $\boldsymbol \chi^{MSM} (\mathbf x)$\cite{noe2013variational} with its elements
\begin{equation}
\chi_i^{MSM}(\mathbf x) = \begin{cases}
1, & \text{if } \mathbf x \in S_i\\
0, & \text{if } \mathbf x \notin S_i
\end{cases}
\end{equation}
Therefore, VAC can be equivalently applied to judge the quality of the MSM based on $\hat \lambda_i^{MSM}$.

\section{State Predictive Information Bottleneck (SPIB)}

Another aspect following the principles of information bottleneck from information theory states that optimal RC should use minimum information from the past to make accurate predictions of the future. Based on this viewpoint, Tiwary et al.\cite{wang2021state} developed an RC deep learning algorithm, known as SPIB, which aims to learn the optimal RC $\mathbf r^{SPIB}$ as the latent variable of a variational autoencoder\cite{kingma2013auto} (VAE) by minimizing the following loss function:
\begin{equation}
\mathcal L^{SPIB}(\boldsymbol \theta) = - \mathbb E[\log q(\mathbf y_{t+\tau}\mid\mathbf r_t^{SPIB}) - \beta\log \frac{p(\mathbf r_t^{SPIB}\mid \mathbf x_t)}{p(\mathbf r_t^{SPIB})}]
\end{equation}
Similarly, we use the notation $\mathbf r_t^{SPIB}:= \mathbf r^{SPIB} (\mathbf x_t)$ to denote the RC coordinate at time $t$. Here, the first $\log$ term is a measure of the accuracy of prediction by $\mathbf r^{SPIB}$, where instead of maximizing the likelihood of observing the conformational coordinates $\mathbf x_{t+\tau}$, SPIB refines the $\mathbf r_t^{SPIB}$ to predict the state label $\mathbf y_{t+\tau}$ of the future conformation $\mathbf x_{t+\tau}$. Therefore, an initial set of state labels for the trajectory is required for SPIB. This can be either done in a supervised way by supplying state labels based on human intuition or in an unsupervised way by performing clustering algorithms in an RC space from an unsupervised machine learning algorithm such as TICA. The initial state labels can be imperfect, since the SPIB algorithm also provides an iterative state label improving scheme to refine the state labels. The second $\log$ term is a regularization term to restrain the complexity of latent space $\mathbf r_t^{SPIB}$, associated with a hyperparameter $\beta$ to control the strength of regularization. We refer interested readers to the original article\cite{wang2021state} for technical details of the algorithm. 

\section{Model Training for Markov State Model Construction}

\subsection{Input Feature Selection}

We used featurized trajectories for all RC learnings in all three systems. For CLN025 and Trp-Cage, we used a similar feature selection strategy. We enumerated all $C_\alpha$-$C_\alpha$, $C_\beta$-$C_\beta$, $C_\gamma$-$C_\gamma$, backbone $N-O$ interatomic distances and included all of them in our input features. In addition, we enumerate all sine/cosine functions of $\phi$,$\psi$ and $\chi_{1-4}$ torsional angles and include them as well. For NTL-9, due to the increase in system size, we reduce the number of input features by including only $C_\alpha$-$C_\alpha$ interatomic distances and sine/cosine functions of $\phi$,$\psi$ and $\chi_{1-4}$ torsional angles.

Moreover, to compensate TICA as a pure linear regression algorithm without any neural network non-linearity, we learned our $\mathbf r^{TICA}_{2D}$ or $\mathbf r^{TICA}_{d_{int}}$ with even more extensive input feature set including all the features above, plus $e^{-d_i}$ functions for all interatomic distances $d_i$ included for each system. Notice we did use the extensive input feature set for getting an overall picture for all systems, but we did not include these additional $e^{-d_i}$ functions as input features for the TICA pre-processing step for FMRC learning or in the training of any other neural network based algorithms. We expect neural network-based algorithms are able to approximate such input features if they are indeed beneficial in the construction of the RC.

As a result, for $\mathbf r^{TICA}_{2D}$ and $\mathbf r^{TICA}_{d_{int}}$ learning, we used 484 input features for CLN025, 1694 input features for Trp-Cage and 1786 input features for NTL-9. For $\mathbf r^{FMRC}$, $\mathbf r^{SRV}$ and $\mathbf r^{SPIB}$, we used 277 input features for CLN025, 913 input features for Trp-Cage and 1045 features for NTL-9.

\subsection{FMRC Neural Network Architecture and Training}

The featurized trajectory was first transformed by TICA at the training lag time $\tau$. We then truncate the TICA transformation and use only the first 30 eigenvectors (CLN025)/20 eigenvectors (Trp-Cage)/35 eigenvectors (NTL-9) as the input for the encoder. We used three hidden layers with 256 hidden nodes plus a linear layer with 2 output nodes as our encoder for all three systems. For D-decoder and L-decoder, we used a neural network of three hidden layers with 256 hidden nodes to approximate the flow matching vector fields $\hat v_{s,\boldsymbol \theta}(\mathbf x^i_{t+\tau,s},\mathbf r^{FMRC}_t)$ and $\hat v_{s,\boldsymbol \theta}(\mathbf x^i_{t,s},\mathbf r^{FMRC}_{t+\tau})$. We used RELU \cite{agarap2018deep} as activation functions for all hidden layers. We set all entries in $\boldsymbol \sigma$ to be 0.001 as the standard deviation of the Gaussian path in main text equation 14. 

For training, after TICA transformation, the featurized trajectory was sliced into time-lagged pairs $\{\mathbf x_t, \mathbf x_{t+\tau}\}$. 90\% of these time-lagged pairs were randomly selected and used for each model training, while the rest of them were used for validation. All models were trained with a batch size of 512 for 100 epochs using the Adam optimizer\cite{kingma2014adam} at a learning rate of 0.001 for the first 50 epochs and 0.0001 for the final 50 epochs. After training, the output of the encoder was min-max normalized, which serves as $\mathbf r^{FMRC}$ for all subsequent analyses and illustrations. 

\subsection{SRV Neural Network Architecture and Training}
We used the same architecture and training protocol for SRV for all three systems. The featurized trajectory was directly used as the input for the neural network. We followed the architecture and training protocol for SRV training in \cite{sidky2019high} except that we used a fixed number of 100 epochs for training. Although a hidden layer of only 100 nodes seems to be small for more than a thousand input features, we decided to follow this training protocol because we found increasing neural network size/depth often leads to problems such as overfitting, training instability, or negative eigenvalues in our preliminary tests. Using their settings, we found the training is much more stable without significant overfitting.

\subsection{SPIB Neural Network Architecture Training}
We used the same architecture and training protocol for SPIB for all three systems. The featurized trajectory was directly used as the trajectory input for the neural network. In addition, SPIB requires initial state labels for each time frame as input. Since we focus on unsupervised RC learning in this article, we used a conventional TICA+k-means procedure to generate the initial labels. Specifically, we apply the TICA transformation to the input features without the extra $e^{-d_i}$ features using the training lag time $\tau$. We then similarly truncate the TICA eigenvector series and only keep $d_{int}$ leading eigenvectors for clustering. 200 microstates (same as the number of microstates we used for MSM construction) were then clustered in this TICA space using minibatch k-means \cite{sculley2010web}. The resulting microstate assignments were then used as input state labels.

We then followed the exact same architecture, hyperparameters and training protocol as those used in the Trp-Cage example in \cite{wang2024information} for all three systems. After training, the latent variable of the VAE was min-max normalized, which serves as $\mathbf r^{SPIB}$ for all subsequent analyses and illustrations. 

\section{$\bold r^{FMRC}$ for Ala2 and Multiple-Walkers OPES Simulations}

We utilized an adaptive sampling dataset of Ala2 from our previous study \cite{zhang2024enhanced} as the training dataset for FMRC. This dataset consists of 80 short trajectories with a simulation length of 40 ps each. For details of the generation of this dataset, please refer to the previous study \cite{zhang2024enhanced}. We used all interatomic distances between all heavy atoms as input features, resulting in a total of 45 features. The input features are first transformed with TICA and truncated at $d_{int}=10$. Considering the small system size, we only used 32 hidden nodes for both the encoder and the decoders of FMRC. We trained the FMRC at $\tau = 1$ ps. We kept the rest of the hyperparameters the same as those used for MSM construction.

The trained FMRC encoder was then converted to a LibTorch script and utilized by the \texttt{PYTORCH\_MODEL} functionality of PLUMED 2.9.0 \cite{tribello2014plumed,bonomi2019promoting} for bias deposition with the OPES \cite{invernizzi2020rethinking} enhanced sampling algorithm. The model is trained with the cpu version of Pytorch 1.13.1 to match the Pytorch version requirement of PLUMED. Specifically, we set up a multiple-walkers OPES simulation using the same procedure as in our previous study \cite{zhang2024enhanced} except using $\bold r^{FMRC}$ as the biasing RC, a \texttt{BIASFACTOR} of 2 and a \texttt{SIGMA} automatically estimated by PLUMED. The multiple-walkers OPES simulations are performed with Gromacs 2023 \cite{mark_abraham_2023_7588711} patched with PLUMED 2.9.0. We ran 3 ns simulation for each of the 16 replicas, resulting in an aggregated OPES simulation time of 48 ns. The convergence was checked and statistical errors were estimated using block analysis \cite{flyvbjerg1989error,invernizzi2020unified} as described in our previous study \cite{zhang2024enhanced}.

\section{Markov State Model Construction}
For all MSM construction, we first performed minibatch k-means \cite{sculley2010web} clustering in each RC space for 200 microstates. We used a maximum likelihood estimator with detailed balance restraint \cite{prinz2011markov} for the estimation of all MSM transition matrices. We then searched for a satisfactory $\tau^{MSM}$ for all MSMs from different RC spaces. Specifically, we asked all MSMs constructed at this $\tau^{MSM}$ must pass the Chapman-Kolmogorov test. For CLN025, $\tau^{MSM} = 40$ ns. For Trp-Cage, $\tau^{MSM} = 50$ ns. For NTL-9, $\tau^{MSM} = 300$ ns.

\clearpage
\newpage
\section{Supplementary Figures}

\begin{figure}[htbp]
    \centering
    \includegraphics[scale=0.7]{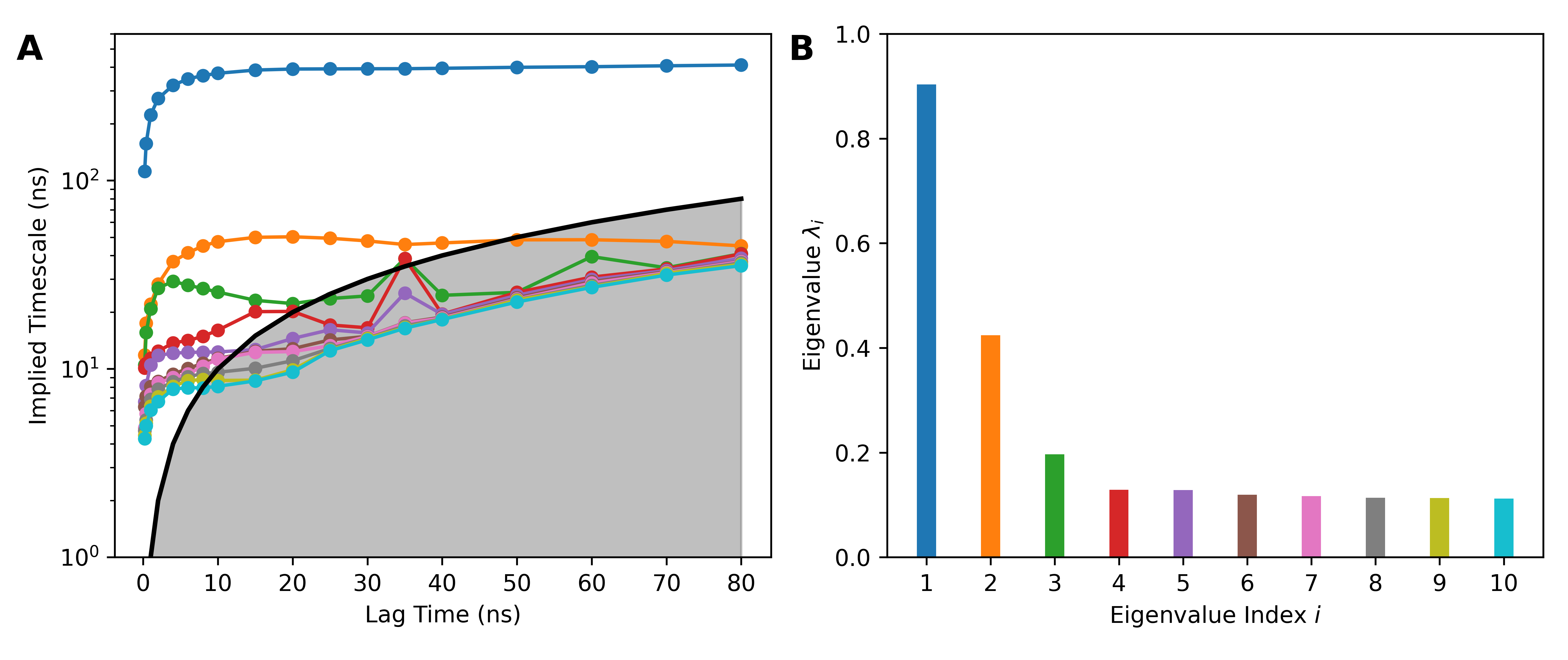}
    \caption{An overview of the dynamic processes present in the CLN025 system. (A) The ITS plot of the 10 leading eigenvectors $\hat \psi^{MSM}_i$ of the MSM constructed in the $\mathbf r^{TICA}_{30D}$ space. It can be seen there are two major slow processes. (B) The 10 highest eigenvalues $\hat \lambda_i^{MSM}$ of the MSM constructed in the $\mathbf r^{TICA}_{30D}$ space at $\tau = 40$ ns. A spectral gap can be seen between $\hat \lambda_2^{MSM}$ and $\hat \lambda_3^{MSM}$.}
    \label{fig:S1}
\end{figure}

\begin{figure}[htbp]
    \centering
    \includegraphics[scale=0.7]{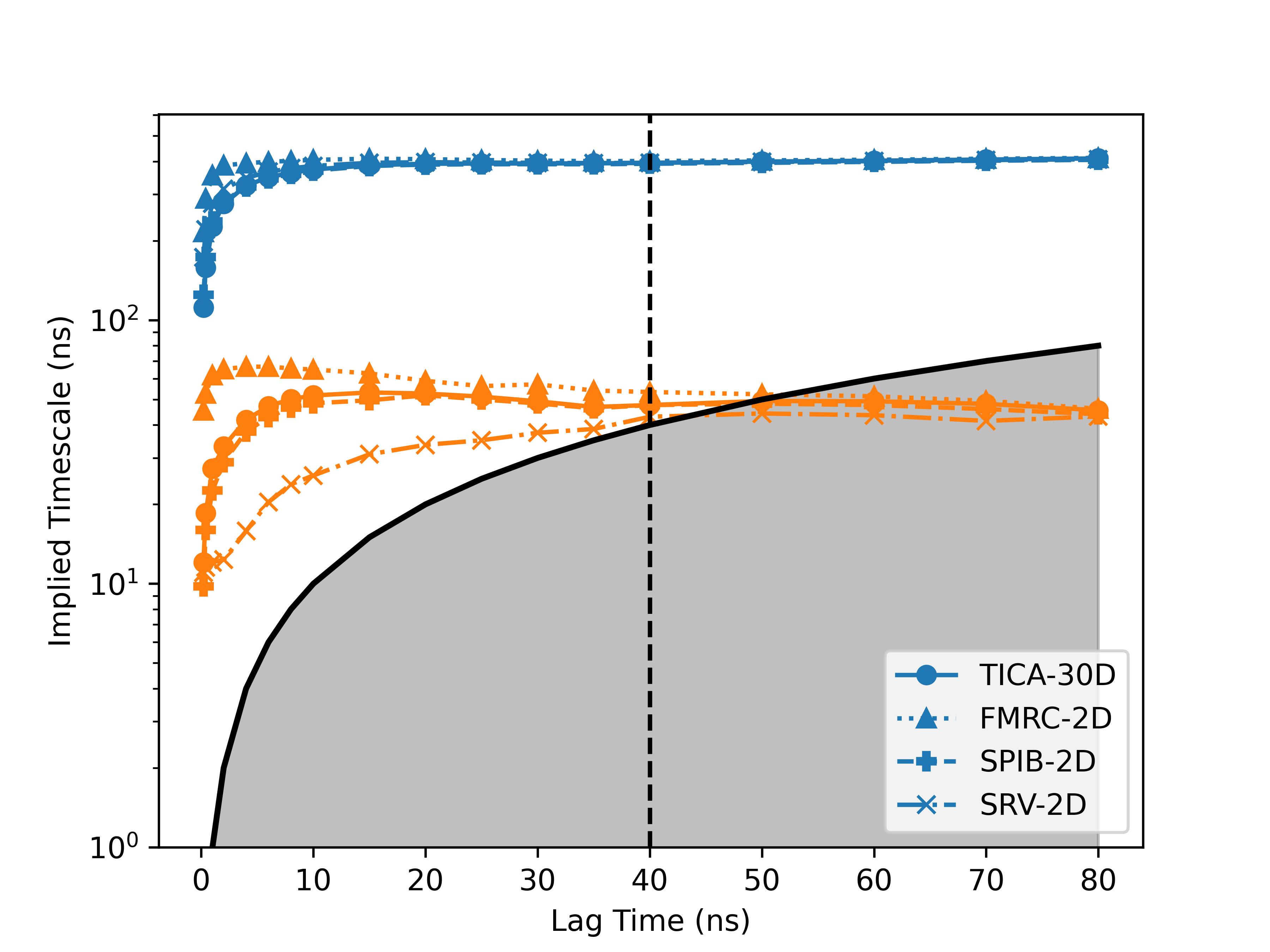}
    \caption{The comparison of ITS plot of the two leading eigenvectors $\hat \psi_i^{MSM}$ of the MSM constructed in different RC spaces for CLN025. A black dashed line is used to denote the lag time $\tau^{MSM}=40$ ns we used to construct MSM.}
    \label{fig:S2}
\end{figure}

\begin{figure}[htbp]
    \centering
    \includegraphics[scale=0.7]{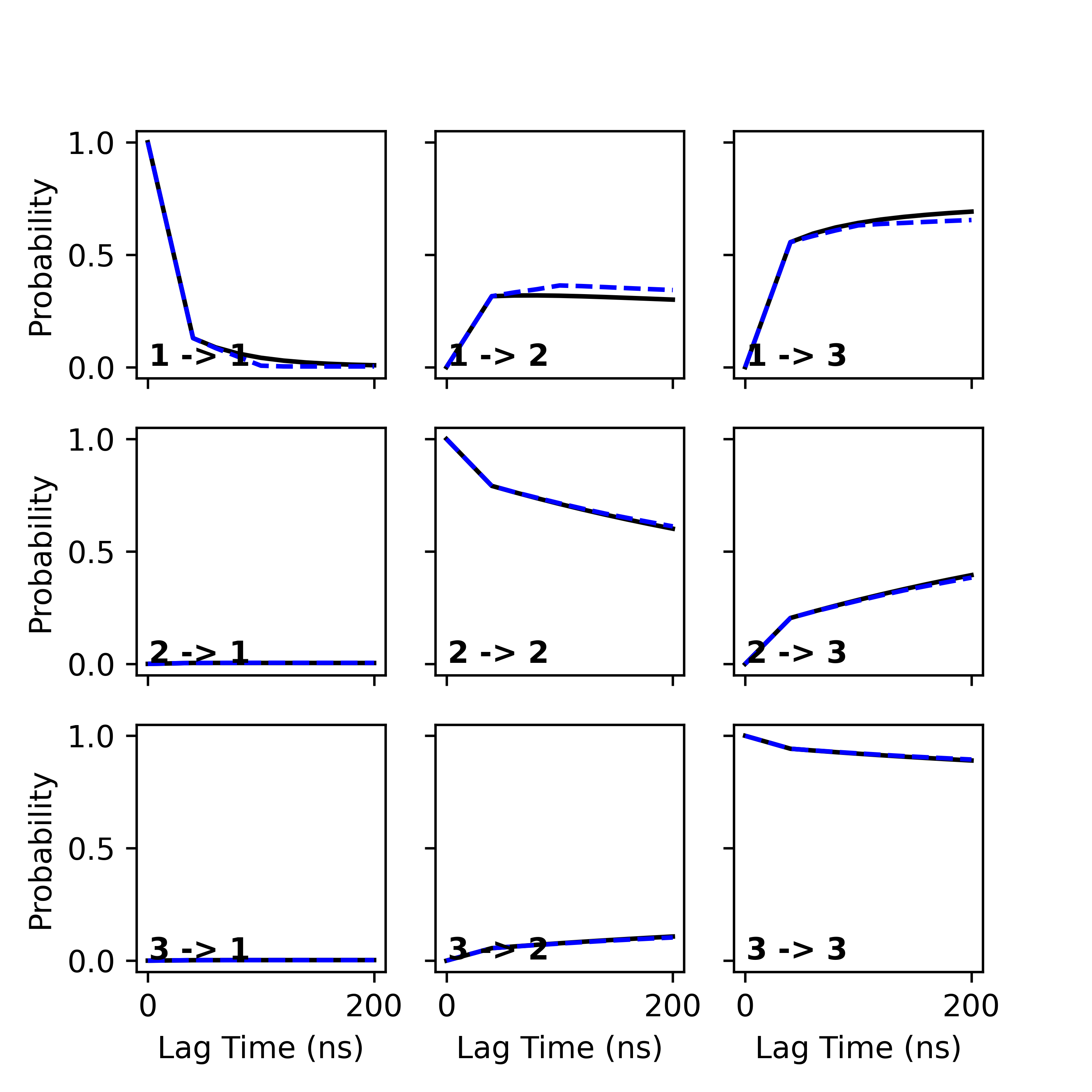}
    \caption{The Chapman-Kolmogorov test validation for the MSM constructed in the $\mathbf r^{FMRC}$ space for CLN025.}
    \label{fig:S3}
\end{figure}

\begin{figure}[htbp]
    \centering
    \includegraphics[scale=0.1]{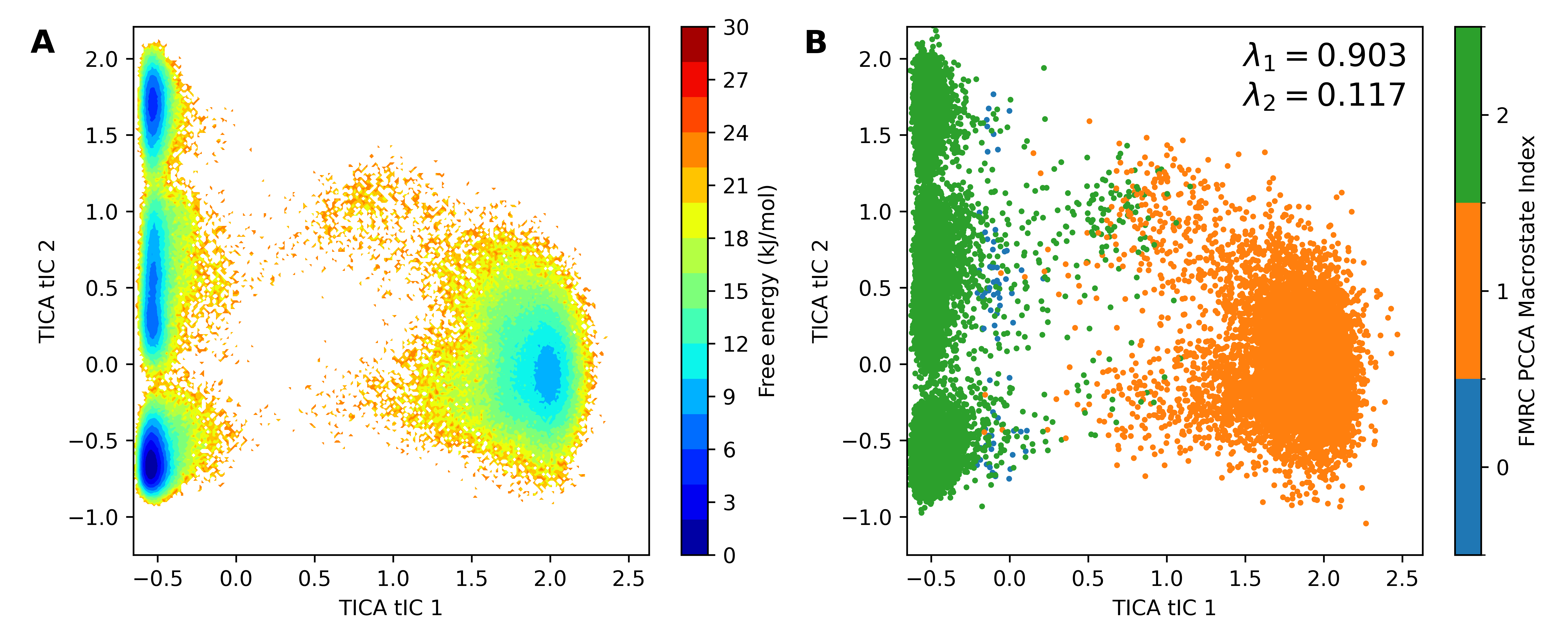}
    \caption{(A) The CLN025 2D FES projection and (B) the CLN025 PCCA+ macrostate assignment (learned from the best MSM constructed in the $\mathbf r^{FMRC}$ space) projection on the $\mathbf r^{TICA}_{2D}$ space where the MSM was constructed. The values of $\hat \lambda_1^{MSM}$ and $\hat \lambda_2^{MSM}$ are shown in the upper right corner of the figure as well.}
    \label{fig:S4}
\end{figure}

\begin{figure}[htbp]
    \centering
    \includegraphics[scale=0.1]{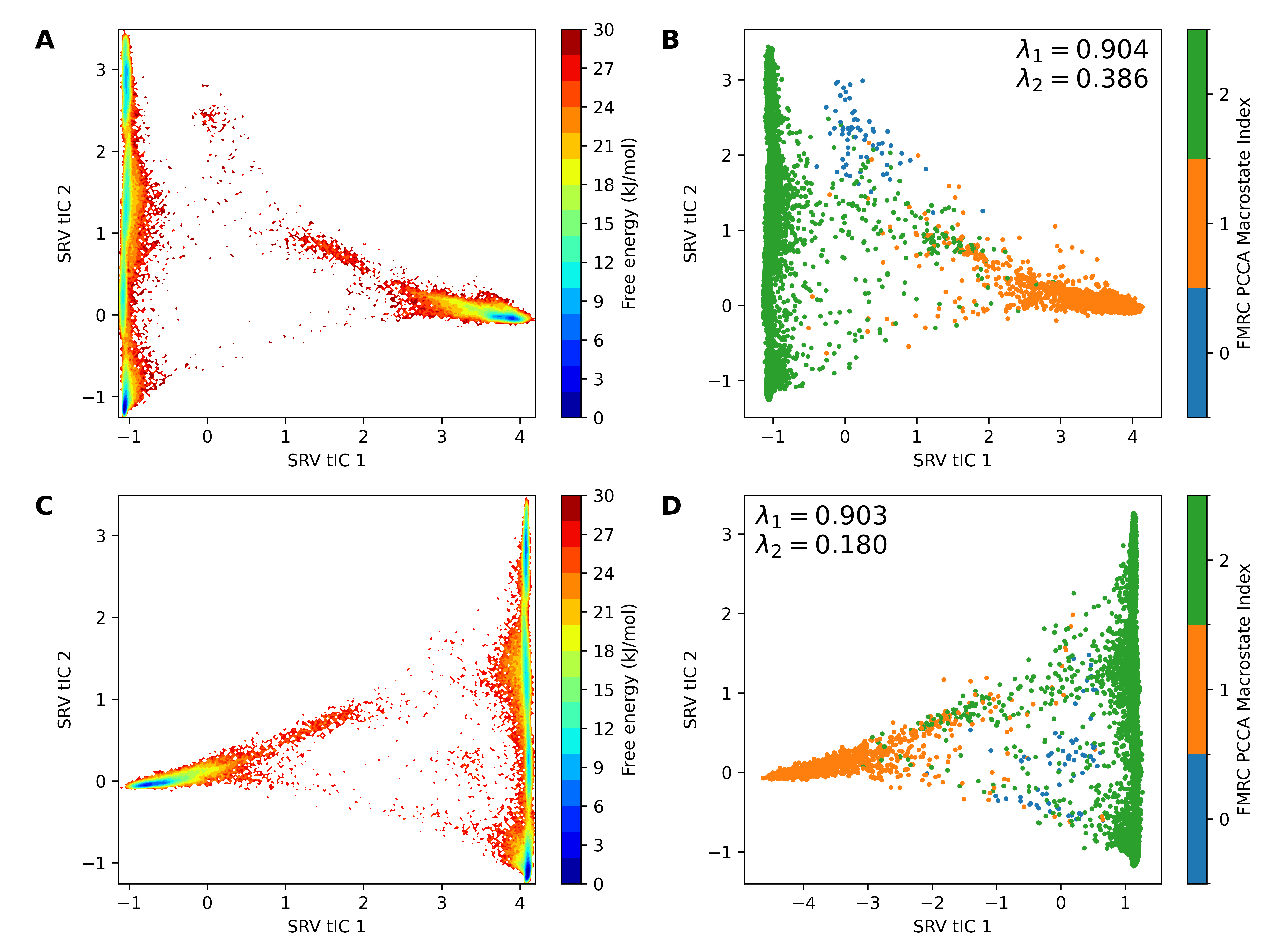}
    \caption{The CLN025 2D FES projection and the CLN025 PCCA+ macrostate assignment (learned from the best MSM constructed in the $\mathbf r^{FMRC}$ space) projection on the $\mathbf r^{SRV}$ space where the best (A,B) or the worst (C,D) MSM was constructed. The best or the worst values of $\hat \lambda_1^{MSM}$ and $\hat \lambda_2^{MSM}$ are shown in B or D as well.}
    \label{fig:S5}
\end{figure}

\begin{figure}[htbp]
    \centering
    \includegraphics[scale=0.1]{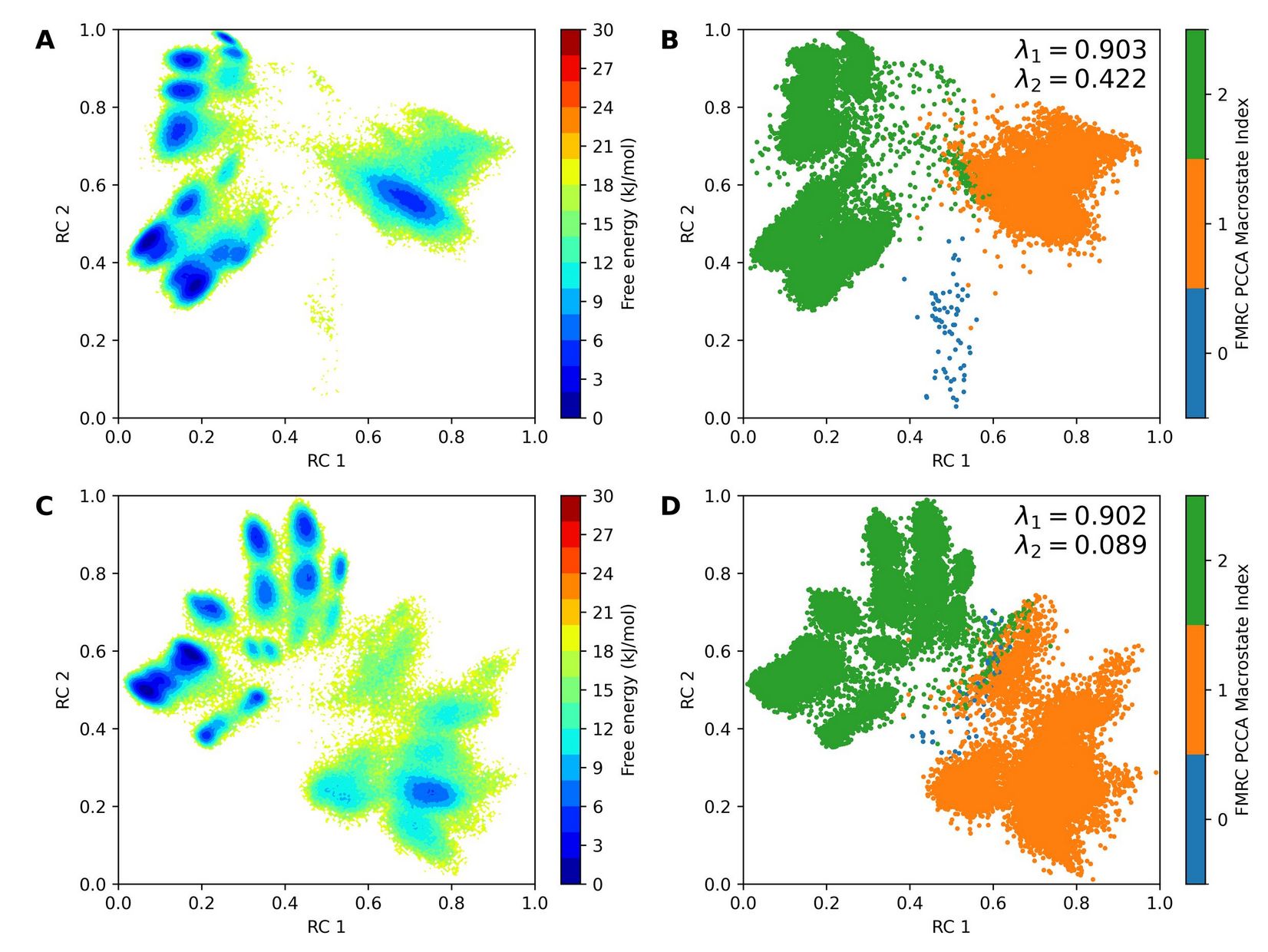}
    \caption{The CLN025 2D FES projection and the CLN025 PCCA+ macrostate assignment (learned from the best MSM constructed in the $\mathbf r^{FMRC}$ space) projection on the normalized $\mathbf r^{SPIB}$ space where the best (A,B) or the worst (C,D) MSM was constructed. The best or the worst values of $\hat \lambda_1^{MSM}$ and $\hat \lambda_2^{MSM}$ are shown in B or D as well.}
    \label{fig:S6}
\end{figure}

\begin{figure}[htbp]
    \centering
    \includegraphics[scale=0.75]{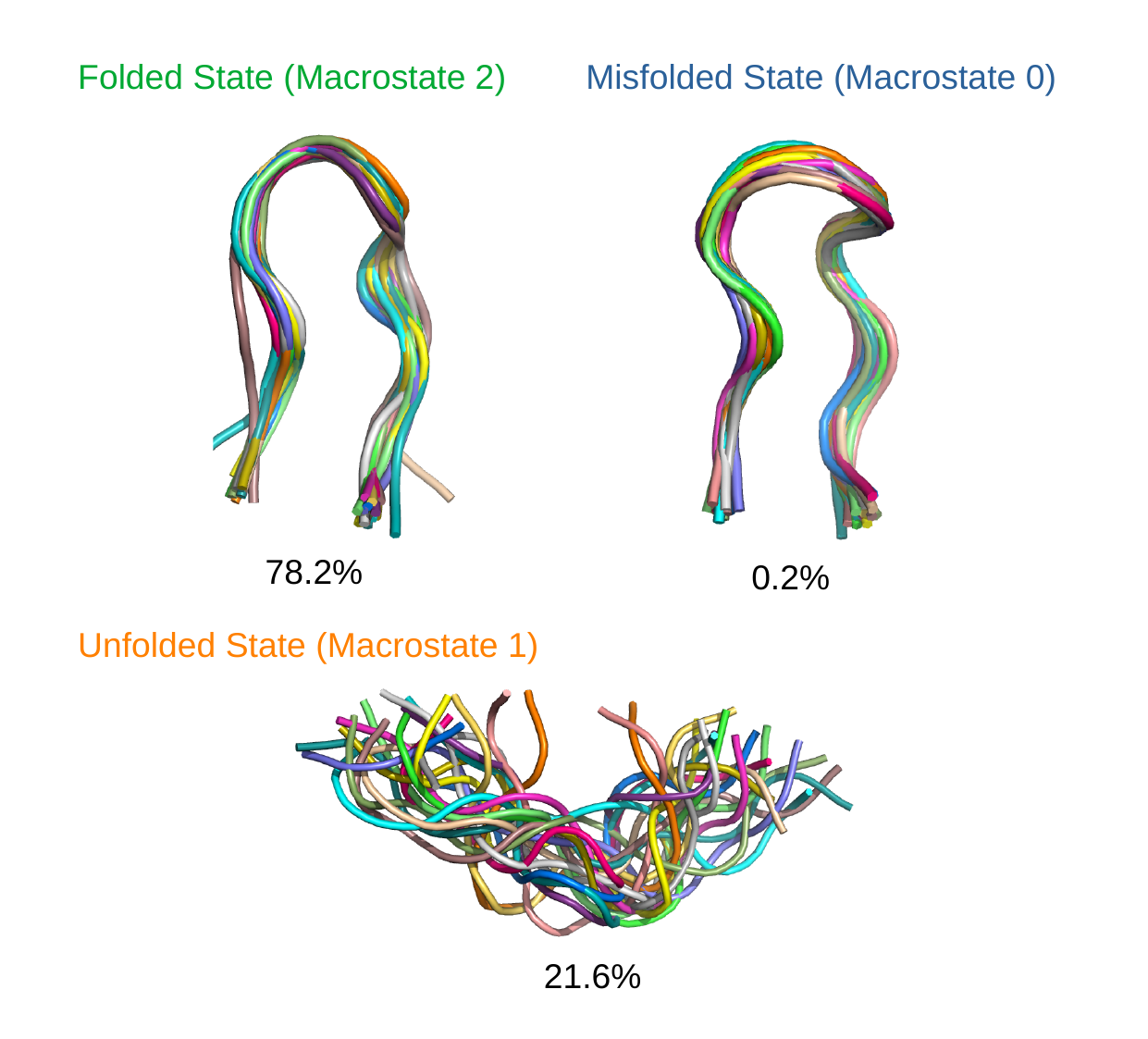}
    \caption{Representative conformations from each macrostate clustered by the PCCA+ algorithm for CLN025. We used the same color for fonts to denote each macrostate as the colors in our PCCA+ projections. Macrostate population at equilibrium estimated from the PCCA+ coarse-grained transition matrix is shown below the conformational ensemble.}
    \label{fig:S7}
\end{figure}

\begin{figure}[htbp]
    \centering
    \includegraphics[scale=1]{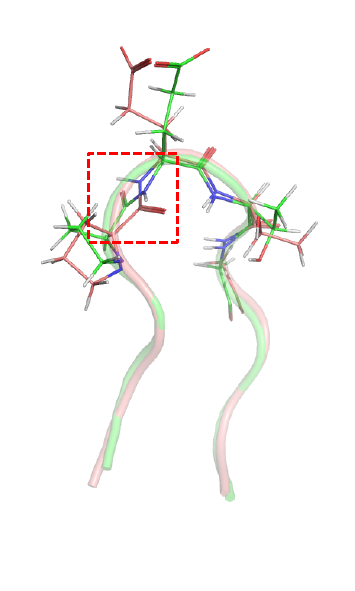}
    \caption{The structural alignment of the misfolded state (macrostate 0) and the folded state (macrostate 2). We only show one structure from each ensemble but this difference is consistently found across the whole ensembles. The major difference is a flipped carbonyl oxygen atom of Glu5, which is highlighted with a red dashed box.}
    \label{fig:S8}
\end{figure}

\clearpage
\begin{figure}[htbp]
    \centering
    \includegraphics[scale=0.1]{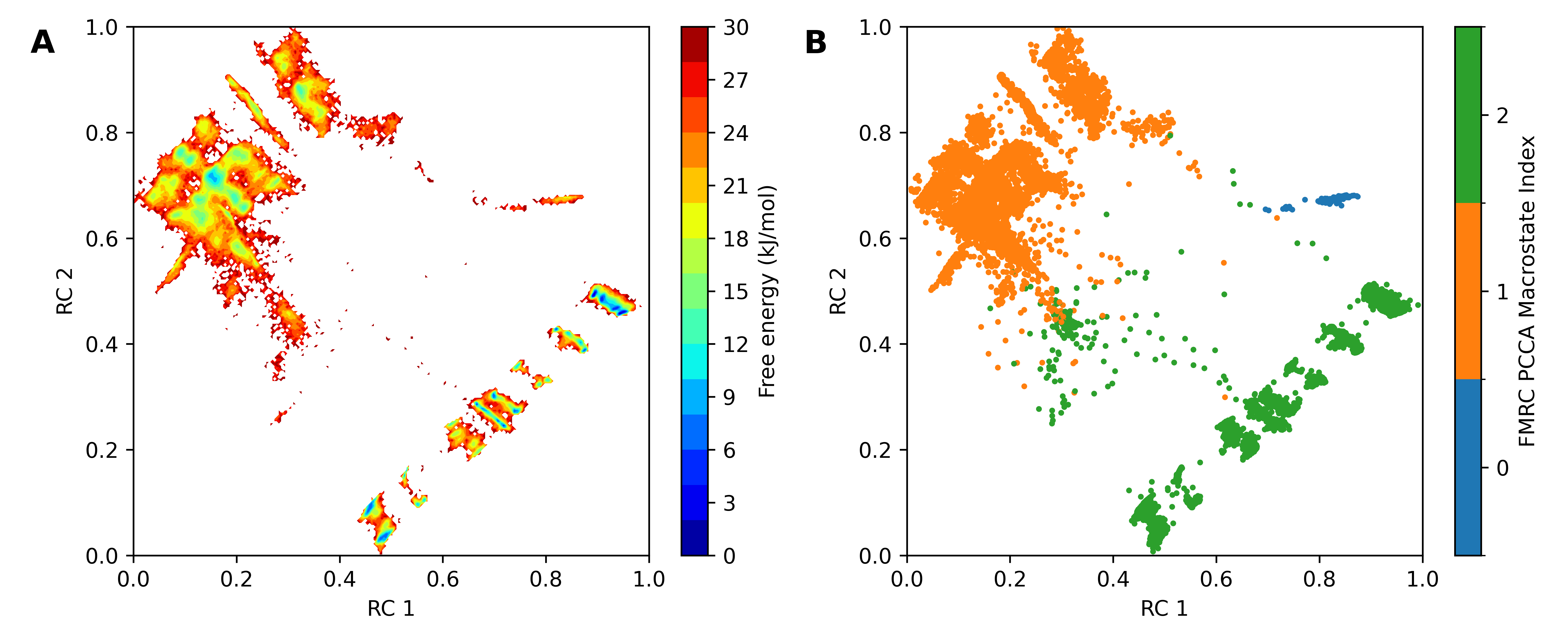}
    \caption{The CLN025 2D FES projection (A) and the CLN025 PCCA+ macrostate assignment (learned from the best MSM constructed in the $\mathbf r^{FMRC}$ space) projection (B) on the normalized $\mathbf r^{FMRC}$ space where the worst MSM was constructed.}
    \label{fig:S9}
\end{figure}

\begin{figure}[htbp]
    \centering
    \includegraphics[scale=0.1]{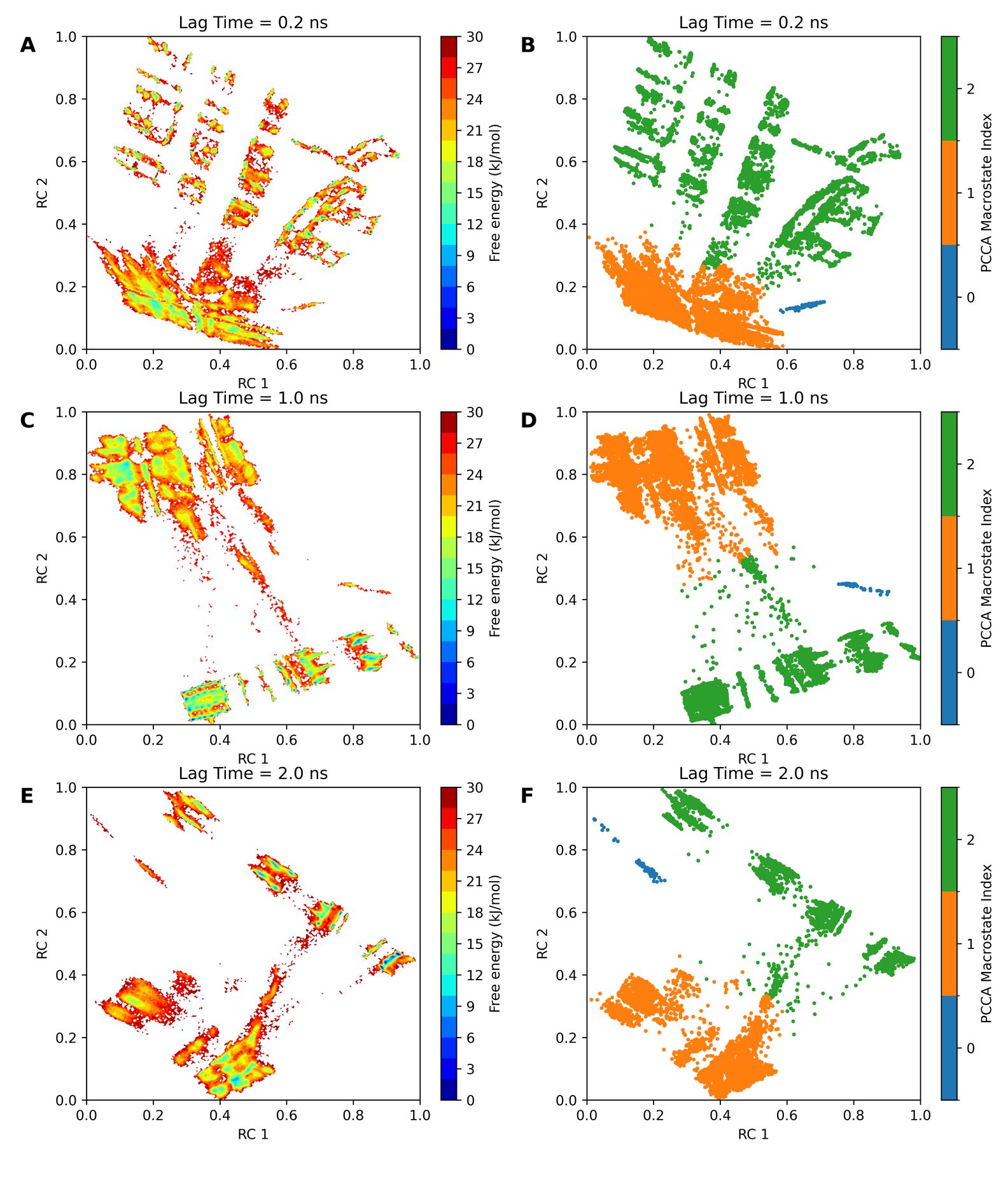}
    \caption{The CLN025 2D FES projection and the CLN025 PCCA+ macrostate assignment projection on the normalized $\mathbf r^{FMRC}$ space learned at $\tau=0.2$ ns (A,B), $\tau=1$ ns (C,D) and $\tau=2$ ns (E,F, same as main text Figure 5).}
    \label{fig:S10}
\end{figure}

\begin{figure}[htbp]
    \centering
    \includegraphics[scale=0.1]{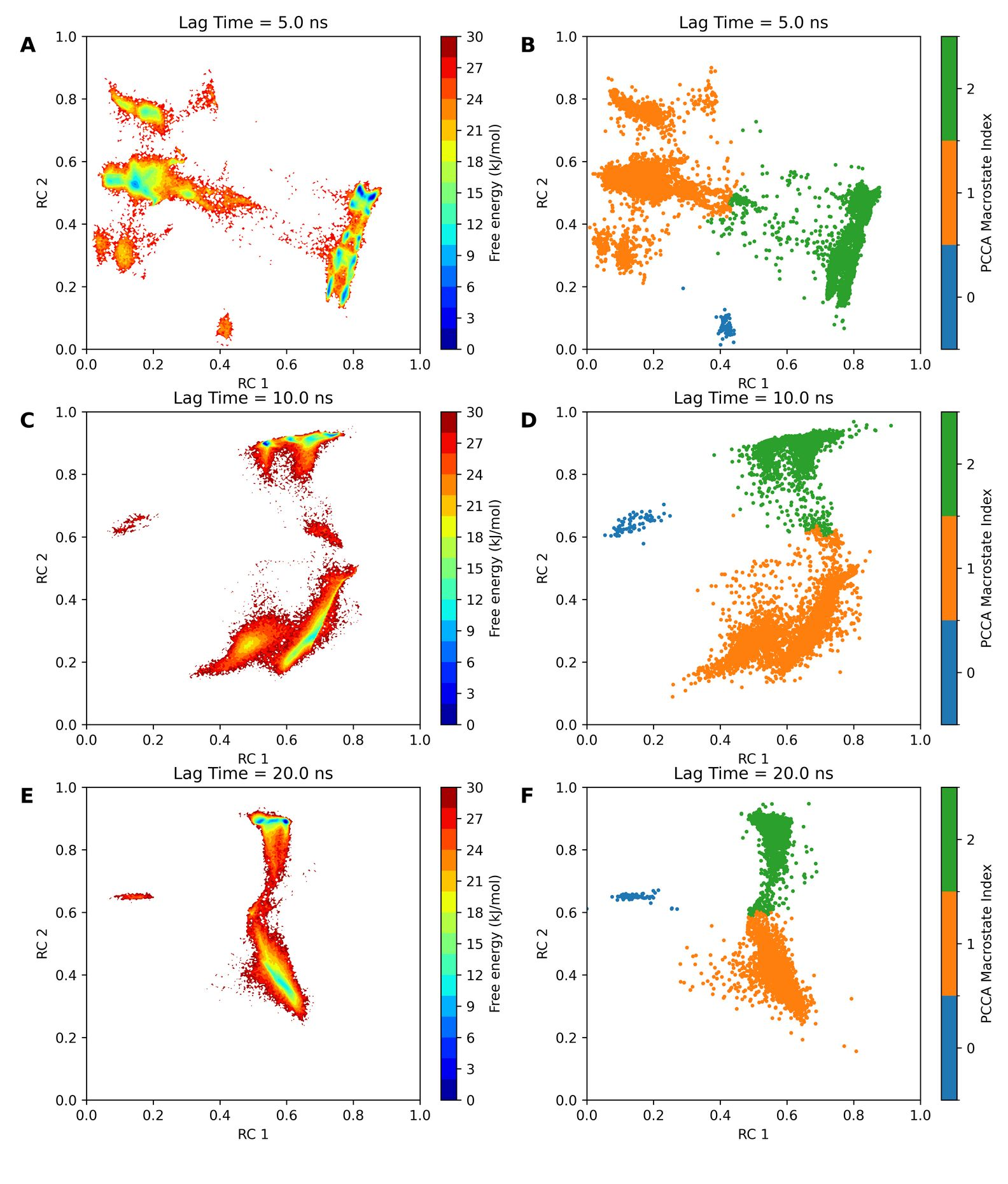}
    \caption{The CLN025 2D FES projection and the CLN025 PCCA+ macrostate assignment projection on the normalized $\mathbf r^{FMRC}$ space learned at $\tau=5$ ns (A,B), $\tau=10$ ns (C,D) and $\tau=20$ ns (E,F).}
    \label{fig:S11}
\end{figure}

\begin{figure}[htbp]
    \centering
    \includegraphics[scale=0.75]{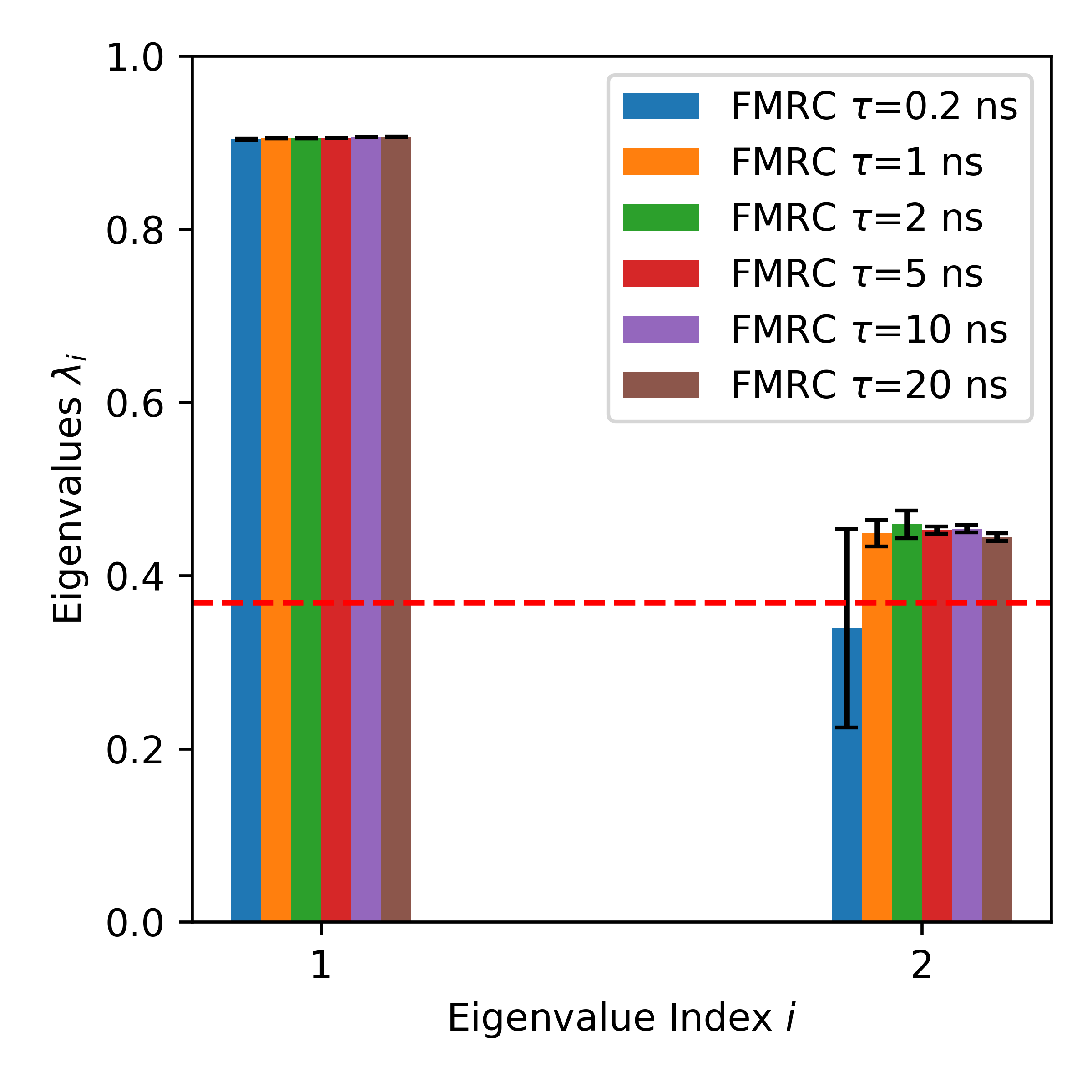}
    \caption{The comparison of $\hat \lambda_i^{MSM}$ of MSMs constructed in $\mathbf r^{FMRC}$ learned at different $\tau$ for CLN025. A red strided line at $\lambda_i = 0.369$ has been drawn to denote a cutoff for the corresponding timescales lower than the $\tau^{MSM}$ for MSM construction. This indicates that the constructed MSM has failed to identify this slow process.}
    \label{fig:S12}
\end{figure}

\begin{figure}[htbp]
    \centering
    \includegraphics[scale=0.75]{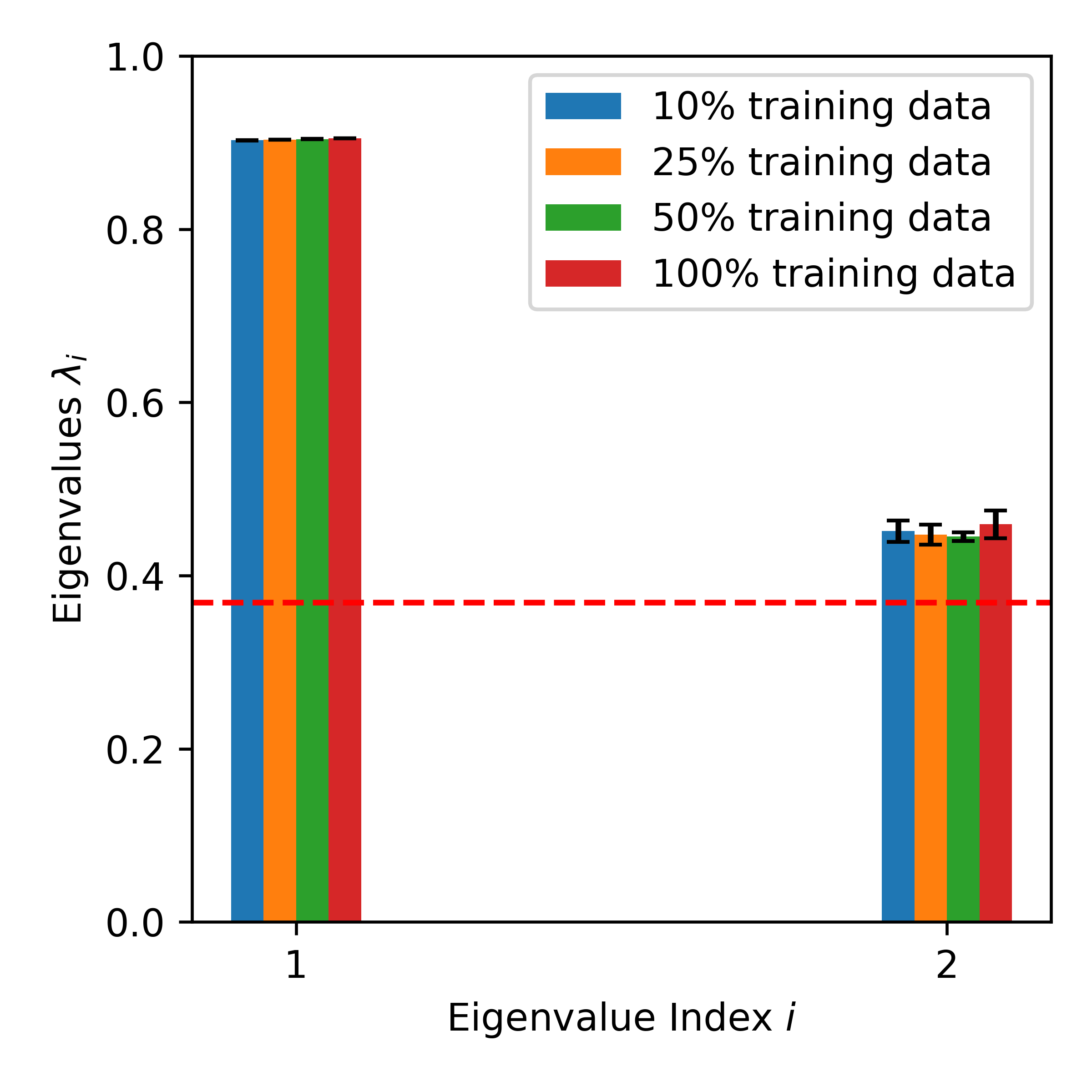}
    \caption{The comparison of $\hat \lambda_i^{MSM}$ of MSMs constructed in $\mathbf r^{FMRC}$ learned from a different amount of training data for CLN025. A red strided line at $\lambda_i = 0.369$ has been drawn to denote a cutoff for the corresponding timescales lower than the $\tau^{MSM}$ for MSM construction. This indicates that the constructed MSM has failed to identify this slow process.}
    \label{fig:S13}
\end{figure}

\clearpage
\begin{figure}[htbp]
    \centering
    \includegraphics[scale=0.7]{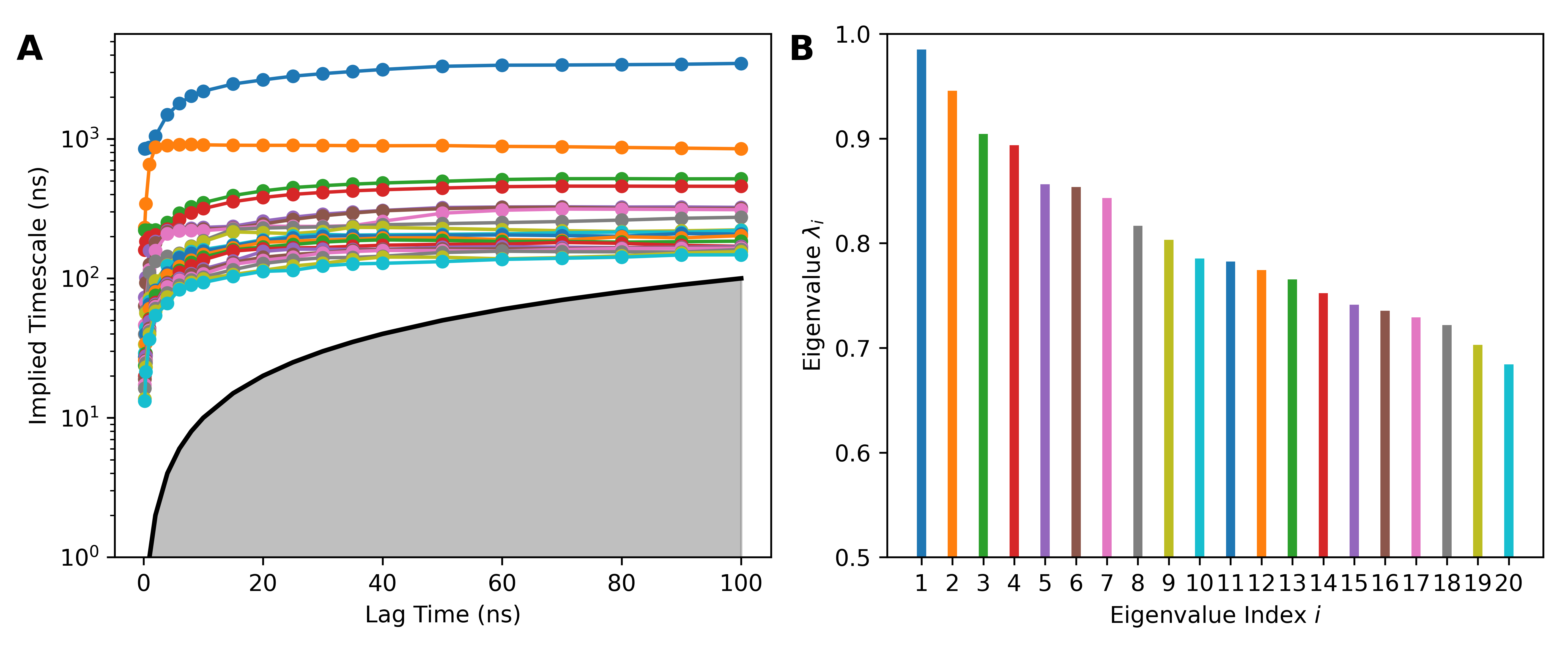}
    \caption{An overview of the dynamic processes present in the Trp-Cage system. (A) The ITS plot of the 20 leading eigenvectors $\hat \psi^{MSM}_i$ of the MSM constructed in the $\mathbf r^{TICA}_{20D}$ space. It can be seen there are multiple major slow processes. (B) The 20 highest eigenvalues $\hat \lambda_i^{MSM}$ of the MSM constructed in the $\mathbf r^{TICA}_{20D}$ space at $\tau = 50$ ns.}
    \label{fig:S14}
\end{figure}

\begin{figure}[htbp]
    \centering
    \includegraphics[scale=0.7]{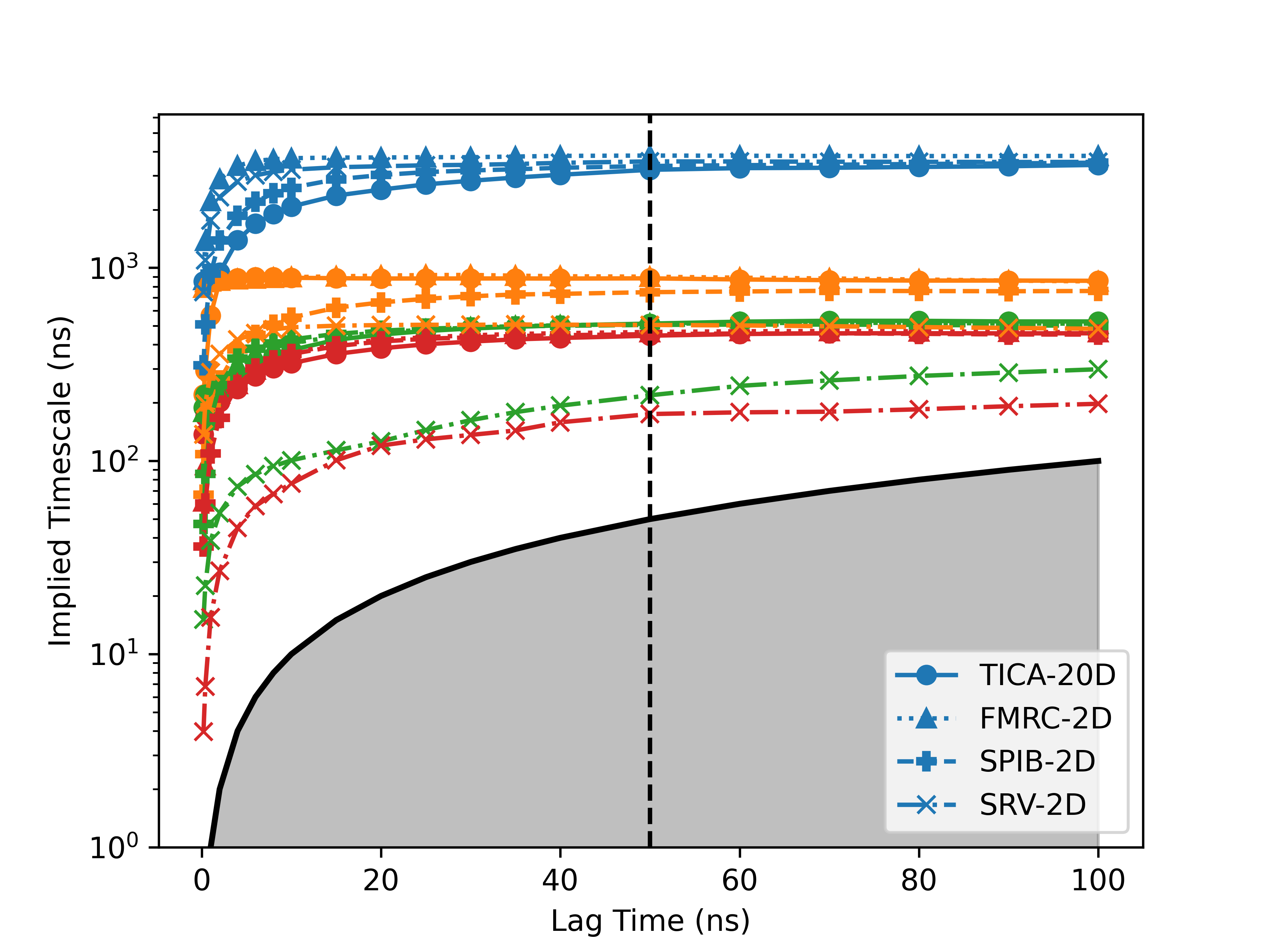}
    \caption{The comparison of ITS plot of the four leading eigenvectors $\hat \psi_i^{MSM}$ of the MSM constructed in different RC spaces for Trp-Cage. A black dashed line is used to denote the lag time $\tau^{MSM}=50$ ns we used to construct MSM. Notice we did not show all the timescales for all 11 eigenvectors for a better visualization.}
    \label{fig:S15}
\end{figure}

\begin{figure}[htbp]
    \centering
    \includegraphics[scale=0.5]{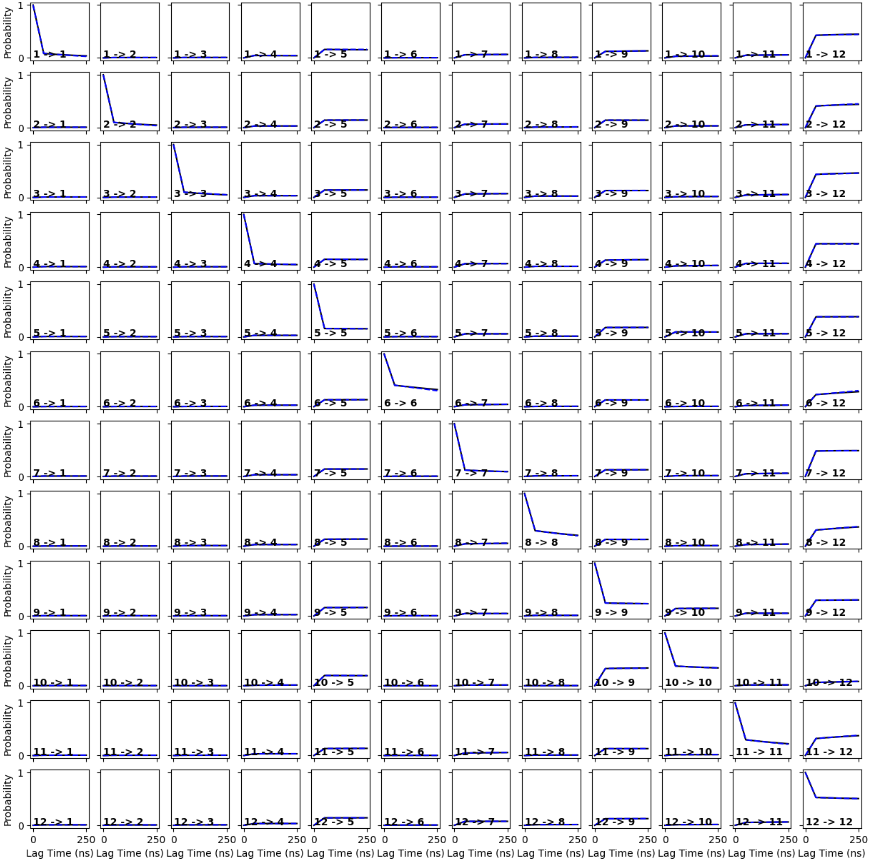}
    \caption{The Chapman-Kolmogorov test validation for the MSM constructed in the $\mathbf r^{FMRC}$ space for Trp-Cage.}
    \label{fig:S16}
\end{figure}

\begin{figure}[htbp]
    \centering
    \includegraphics[scale=0.1]{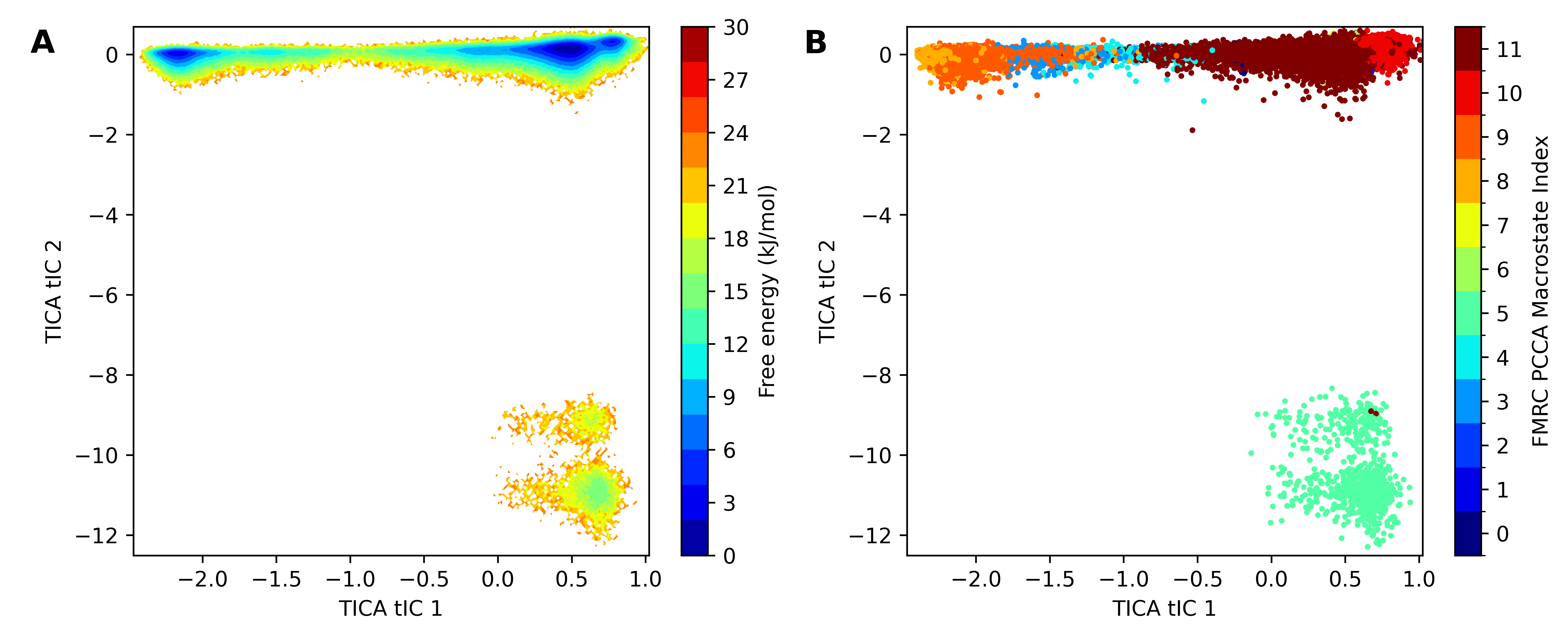}
    \caption{(A) The Trp-Cage 2D FES projection and (B) the Trp-Cage PCCA+ macrostate assignment (learned from the best MSM constructed in the $\mathbf r^{FMRC}$ space) projection on the $\mathbf r^{TICA}_{2D}$ space where the MSM was constructed.}
    \label{fig:S17}
\end{figure}

\begin{figure}[htbp]
    \centering
    \includegraphics[scale=0.1]{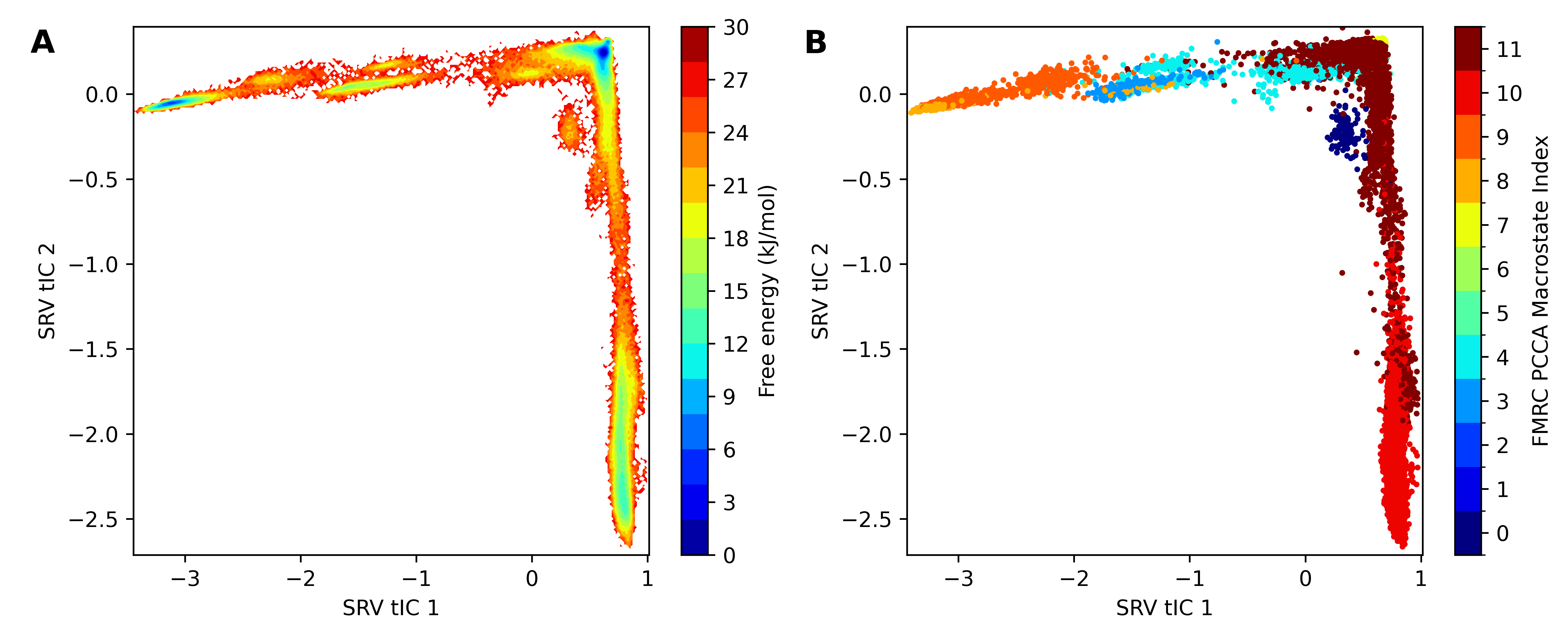}
    \caption{(A) The Trp-Cage 2D FES projection and (B) the Trp-Cage PCCA+ macrostate assignment (learned from the best MSM constructed in the $\mathbf r^{FMRC}$ space) projection on the $\mathbf r^{SRV}$ space where the MSM was constructed.}
    \label{fig:S18}
\end{figure}

\begin{figure}[htbp]
    \centering
    \includegraphics[scale=0.1]{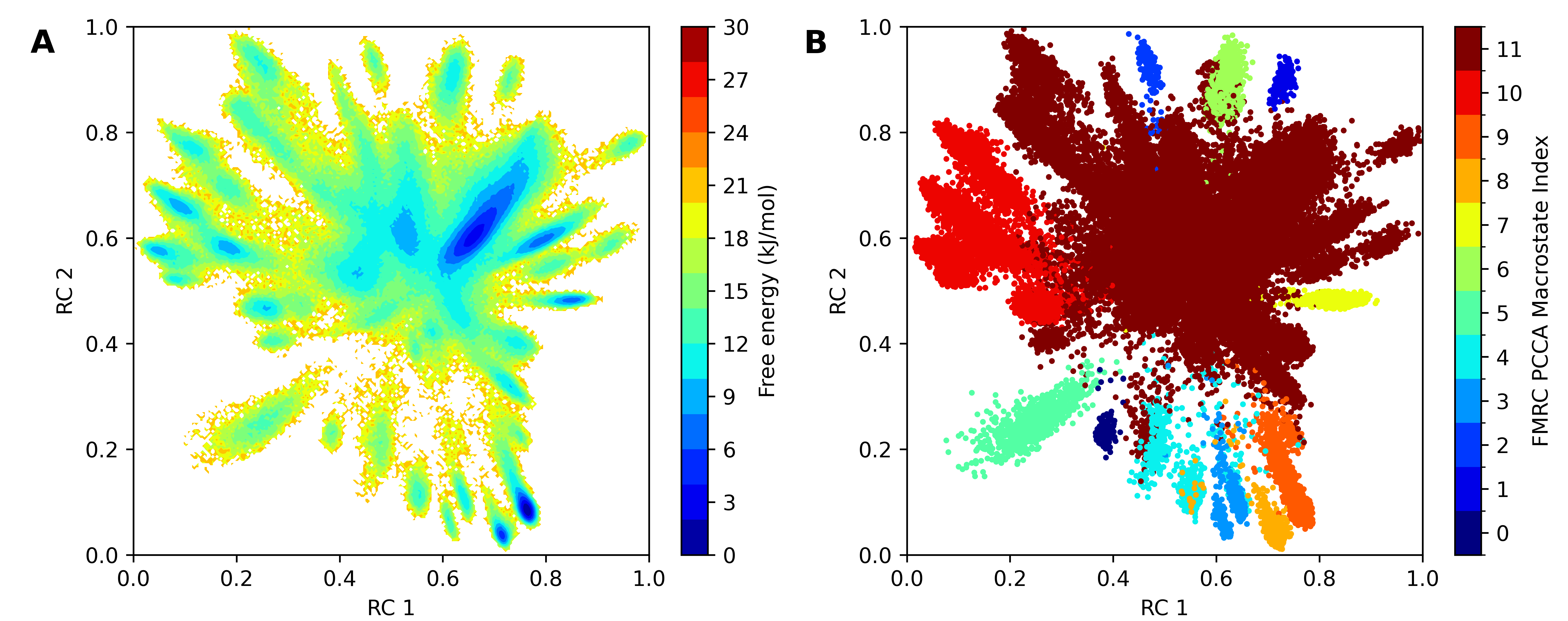}
    \caption{(A) The Trp-Cage 2D FES projection and (B) the Trp-Cage PCCA+ macrostate assignment (learned from the best MSM constructed in the $\mathbf r^{FMRC}$ space) projection on the normalized $\mathbf r^{SPIB}$ space where the MSM was constructed.}
    \label{fig:S19}
\end{figure}

\clearpage
\begin{figure}[htbp]
    \centering
    \includegraphics[scale=0.75]{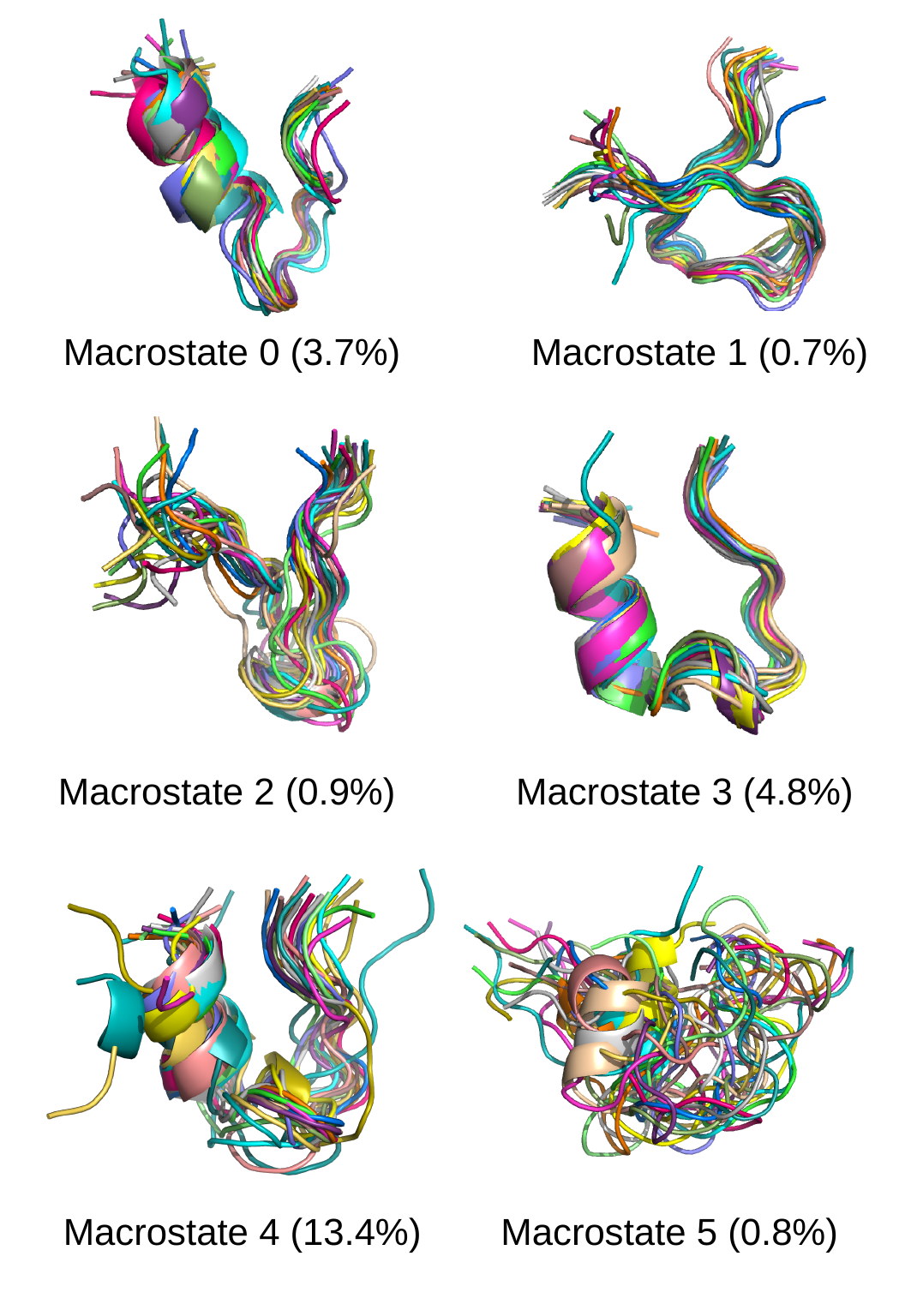}
    \caption{Representative conformations from macrostate 0-5 clustered by the PCCA+ algorithm for Trp-Cage. Macrostate population at equilibrium estimated from the PCCA+ coarse-grained transition matrix is shown below the conformational ensemble.}
    \label{fig:S20}
\end{figure}

\clearpage
\begin{figure}[htbp]
    \centering
    \includegraphics[scale=0.75]{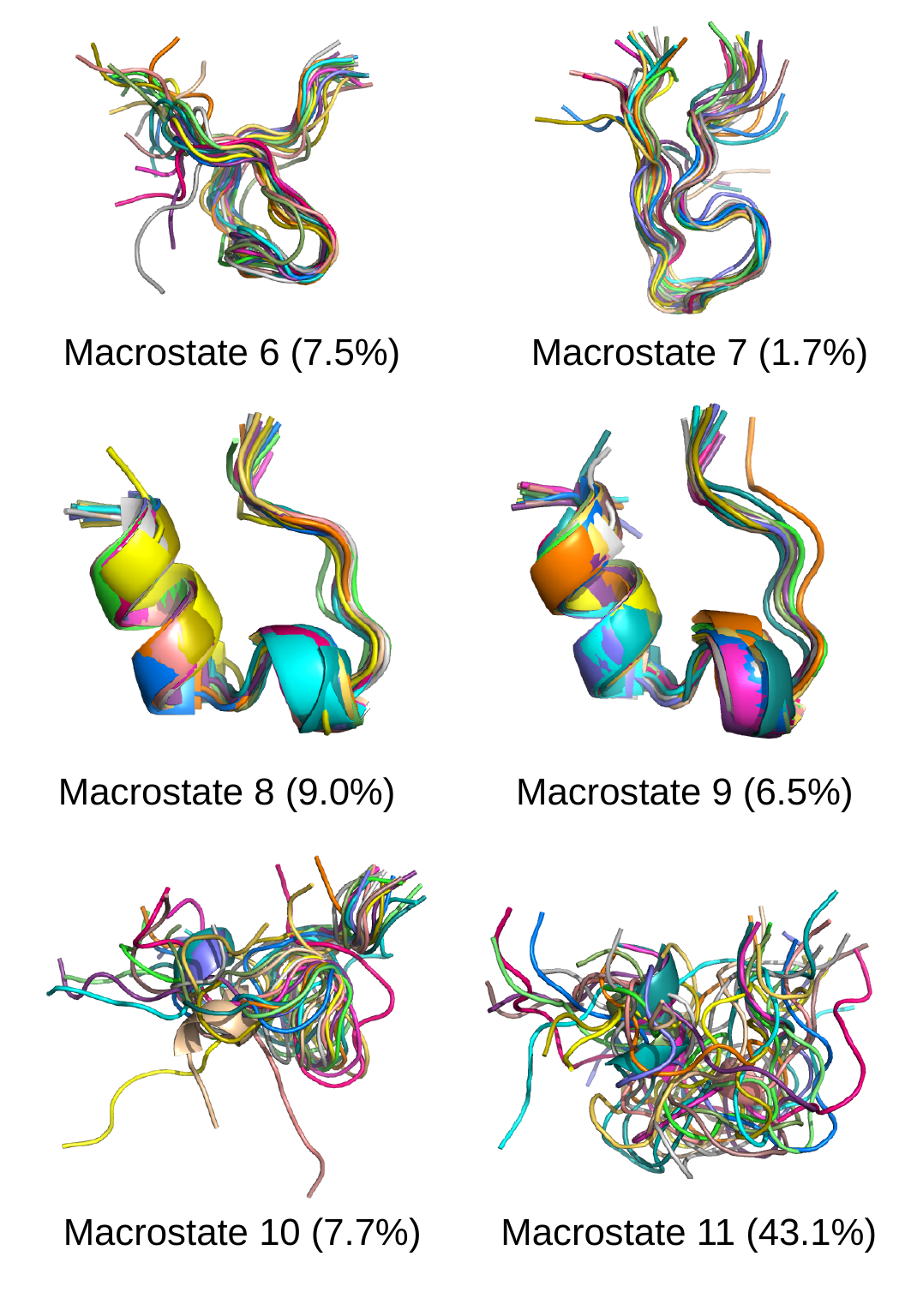}
    \caption{Representative conformations from macrostate 6-11 clustered by the PCCA+ algorithm for Trp-Cage. Macrostate population at equilibrium estimated from the PCCA+ coarse-grained transition matrix is shown below the conformational ensemble.}
    \label{fig:S21}
\end{figure}

\clearpage
\begin{figure}[htbp]
    \centering
    \includegraphics[scale=0.6]{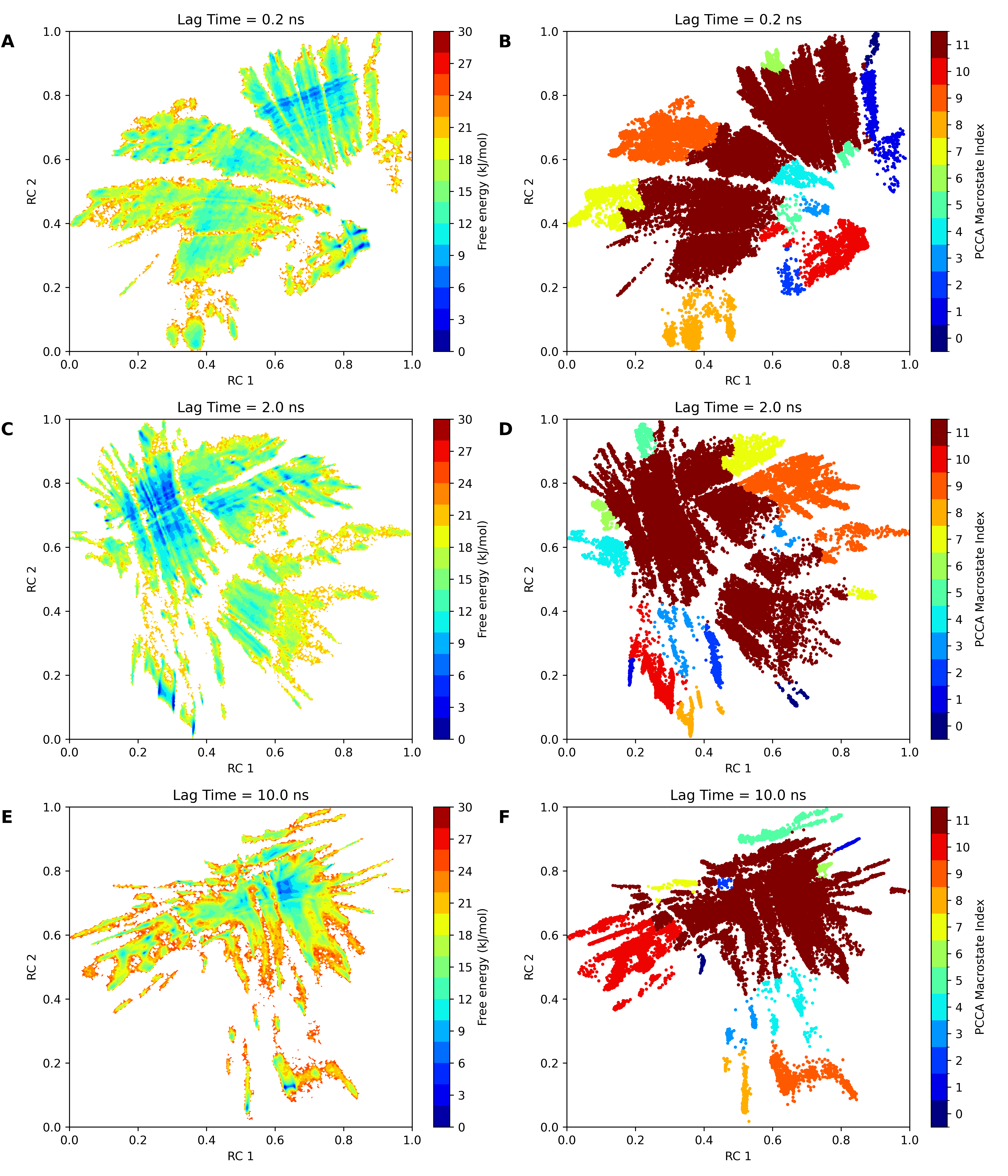}
    \caption{The Trp-Cage 2D FES projection and the Trp-Cage PCCA+ macrostate assignment projection on the normalized $\mathbf r^{FMRC}$ space learned at $\tau=0.2$ ns (A, B), $\tau=2$ ns (C, D) and $\tau=10$ ns (E, F, same as main text Figure 7). Notice the macrostate assignment indices may not correspond to the same macrostate across plots for different $\tau$.}
    \label{fig:S22}
\end{figure}

\clearpage
\begin{figure}[htbp]
    \centering
    \includegraphics[scale=0.6]{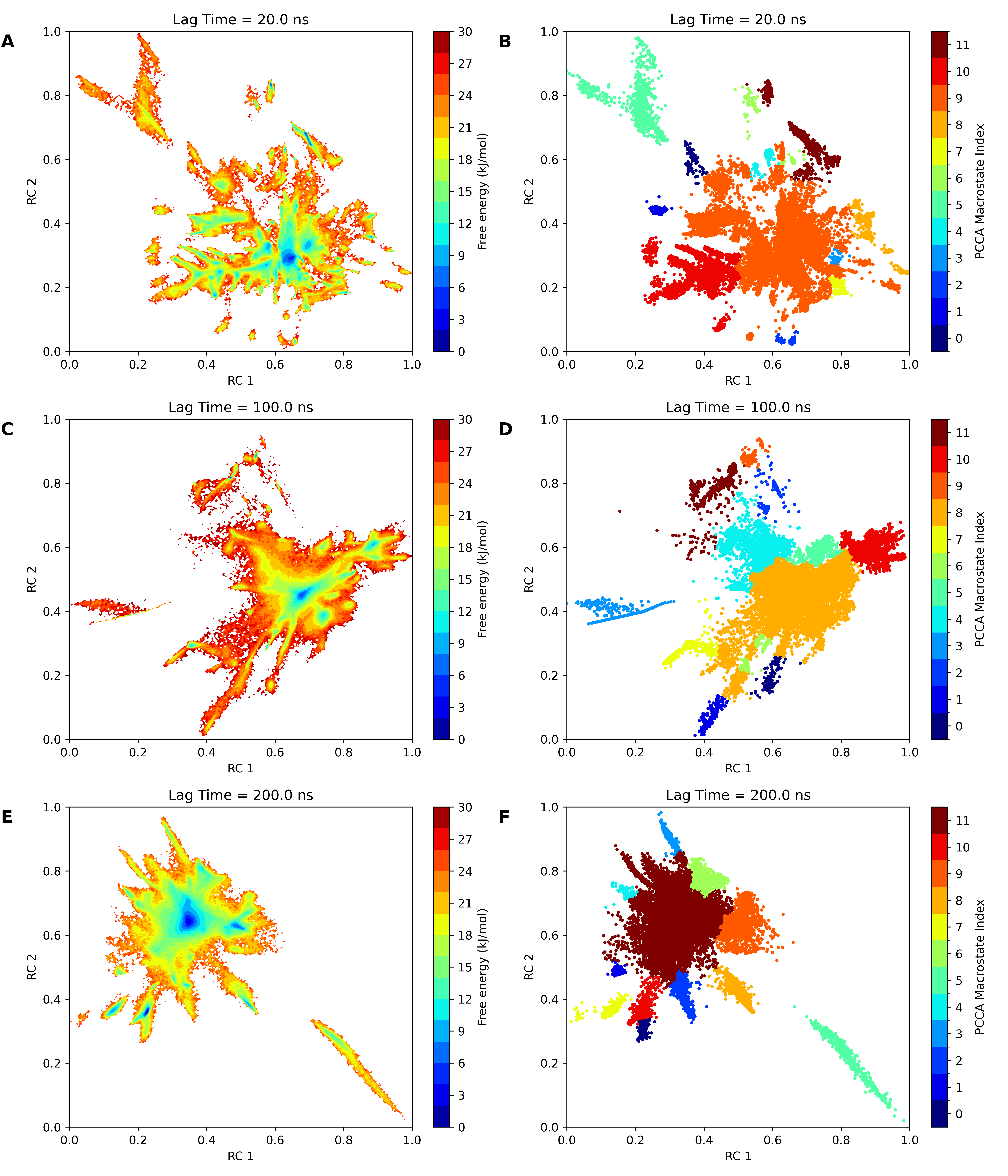}
    \caption{The Trp-Cage 2D FES projection and the Trp-Cage PCCA+ macrostate assignment projection on the normalized $\mathbf r^{FMRC}$ space learned at $\tau=20$ ns (A,B), $\tau=100$ ns (C,D) and $\tau=200$ ns (E,F). Notice the macrostate assignment indices may not correspond to the same macrostate across plots for different $\tau$.}
    \label{fig:S23}
\end{figure}

\clearpage
\begin{figure}[htbp]
    \centering
    \includegraphics[scale=0.5]{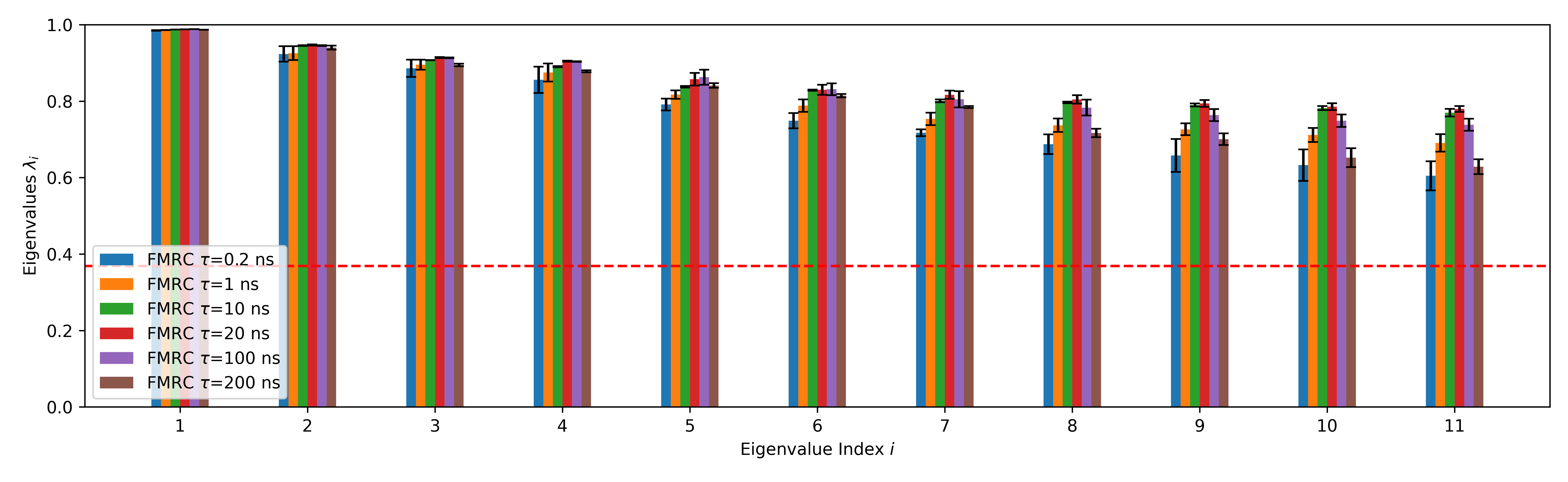}
    \caption{The comparison of $\hat \lambda_i^{MSM}$ of MSMs constructed in $\mathbf r^{FMRC}$ learned at different $\tau$ for Trp-Cage. A red strided line at $\lambda_i = 0.369$ has been drawn to denote a cutoff for the corresponding timescales lower than the $\tau^{MSM}$ for MSM construction. This indicates that the constructed MSM has failed to identify this slow process.}
    \label{fig:S24}
\end{figure}

\clearpage
\begin{figure}[htbp]
    \centering
    \includegraphics[scale=0.5]{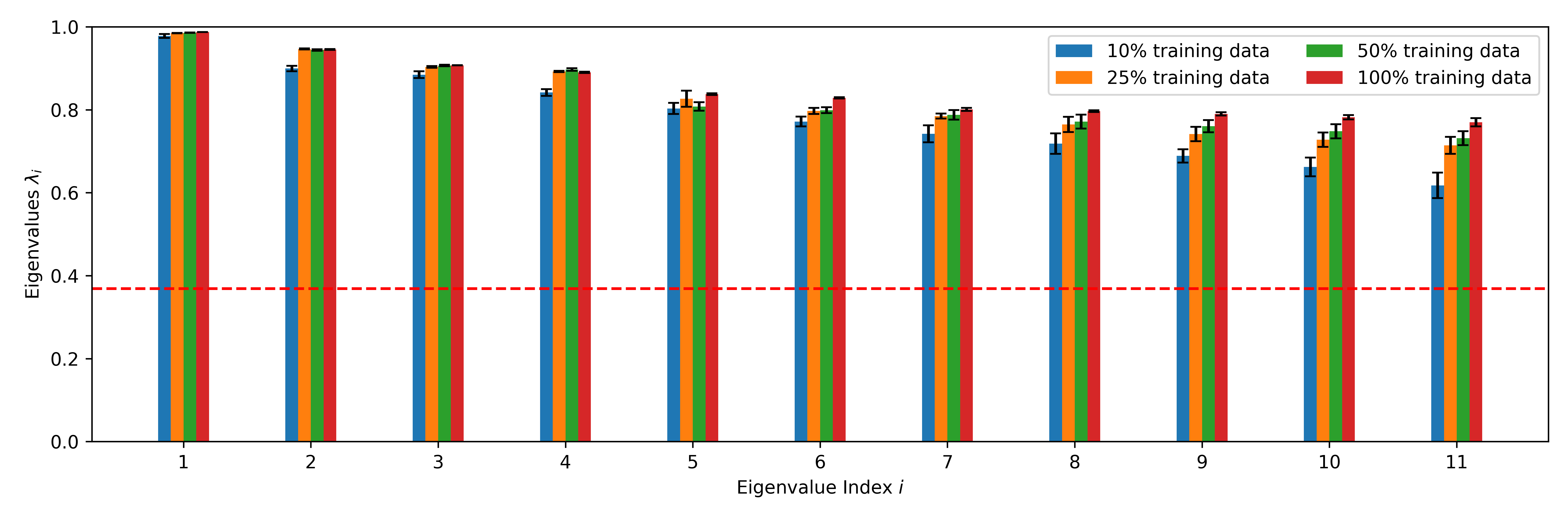}
    \caption{The comparison of $\hat \lambda_i^{MSM}$ of MSMs constructed in $\mathbf r^{FMRC}$ learned from a different amount of training data for Trp-Cage. A red strided line at $\lambda_i = 0.369$ has been drawn to denote a cutoff for the corresponding timescales lower than the $\tau^{MSM}$ for MSM construction. This indicates that the constructed MSM has failed to identify this slow process.}
    \label{fig:S25}
\end{figure}

\clearpage
\begin{figure}[htbp]
    \centering
    \includegraphics[scale=0.7]{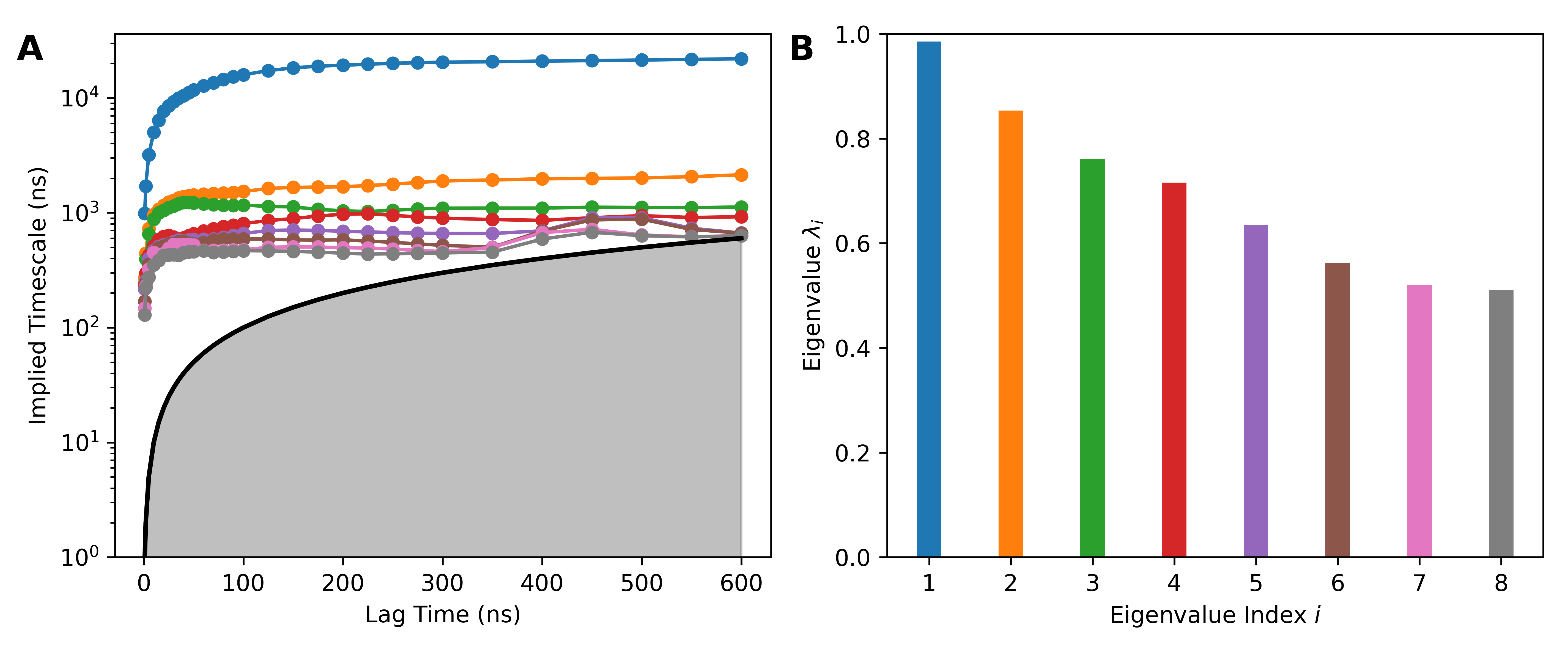}
    \caption{An overview of the dynamic processes present in the NTL-9 system. (A) The ITS plot of the 8 leading eigenvectors $\hat \psi^{MSM}_i$ of the MSM constructed in the $\mathbf r^{TICA}_{35D}$ space. It can be seen there are multiple major slow processes. (B) The 8 highest eigenvalues $\hat \lambda_i^{MSM}$ of the MSM constructed in the $\mathbf r^{TICA}_{35D}$ space at $\tau = 300$ ns.}
    \label{fig:S26}
\end{figure}

\clearpage
\begin{figure}[htbp]
    \centering
    \includegraphics[scale=0.7]{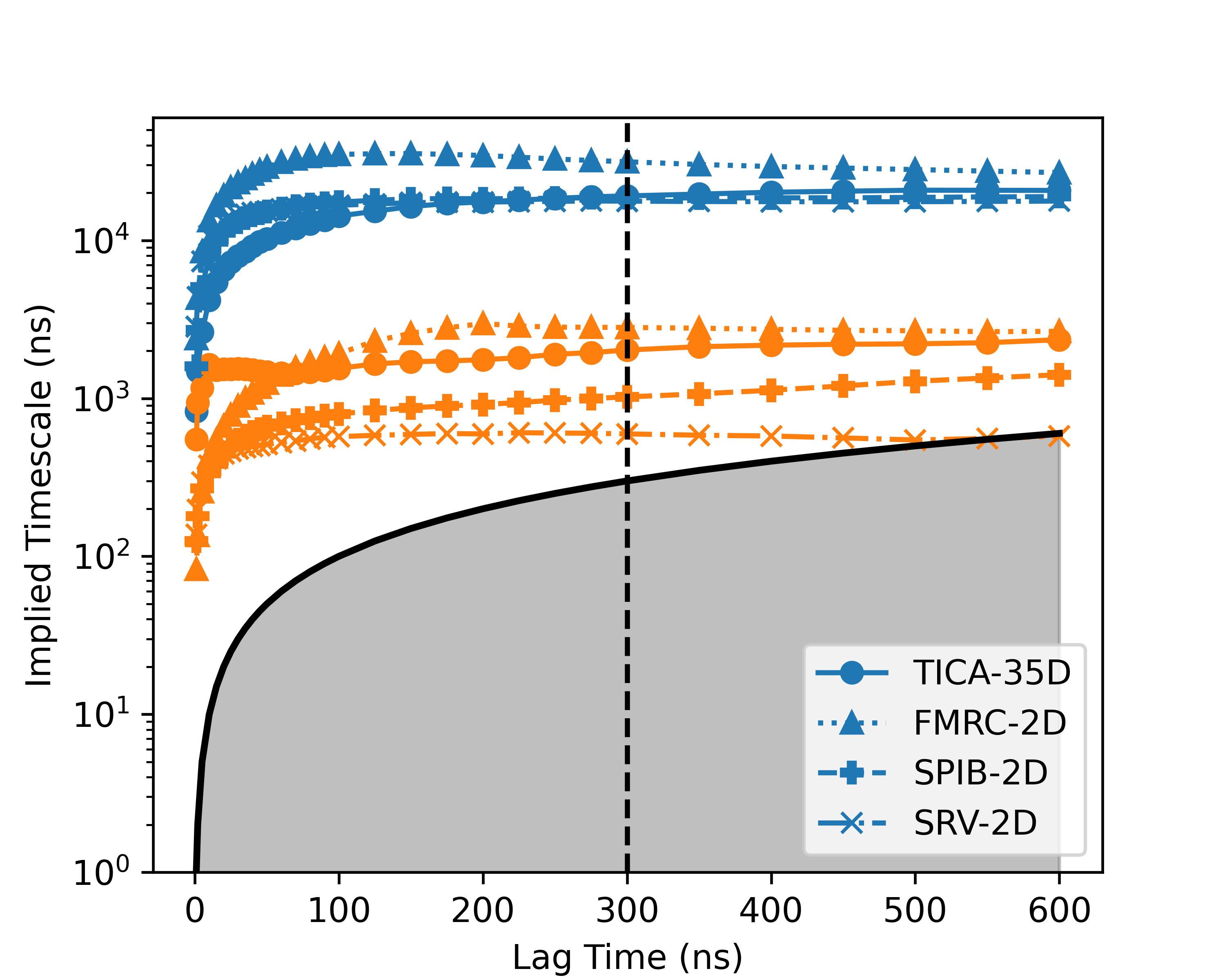}
    \caption{The comparison of ITS plot of the four leading eigenvectors $\hat \psi_i^{MSM}$ of the MSM constructed in different RC spaces for NTL-9. A black dashed line is used to denote the lag time $\tau^{MSM}=300$ ns we used to construct MSM. Notice we did not show all the timescales for all 5 eigenvectors for a better visualization.}
    \label{fig:S27}
\end{figure}

\clearpage
\begin{figure}[htbp]
    \centering
    \includegraphics[scale=0.85]{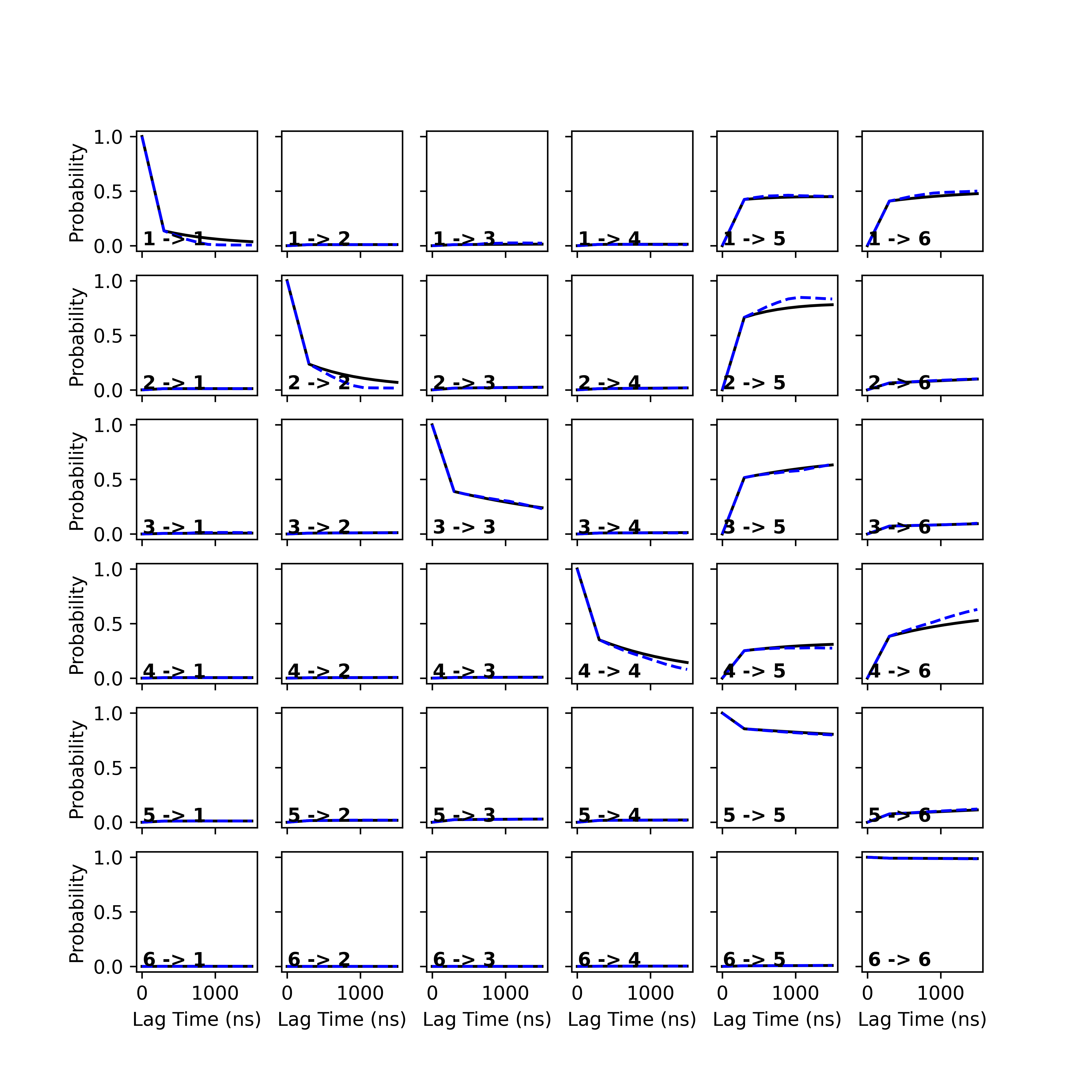}
    \caption{The Chapman-Kolmogorov test validation for the MSM constructed in the $\mathbf r^{FMRC}$ space for NTL-9.}
    \label{fig:S28}
\end{figure}

\clearpage
\begin{figure}[htbp]
    \centering
    \includegraphics[scale=0.1]{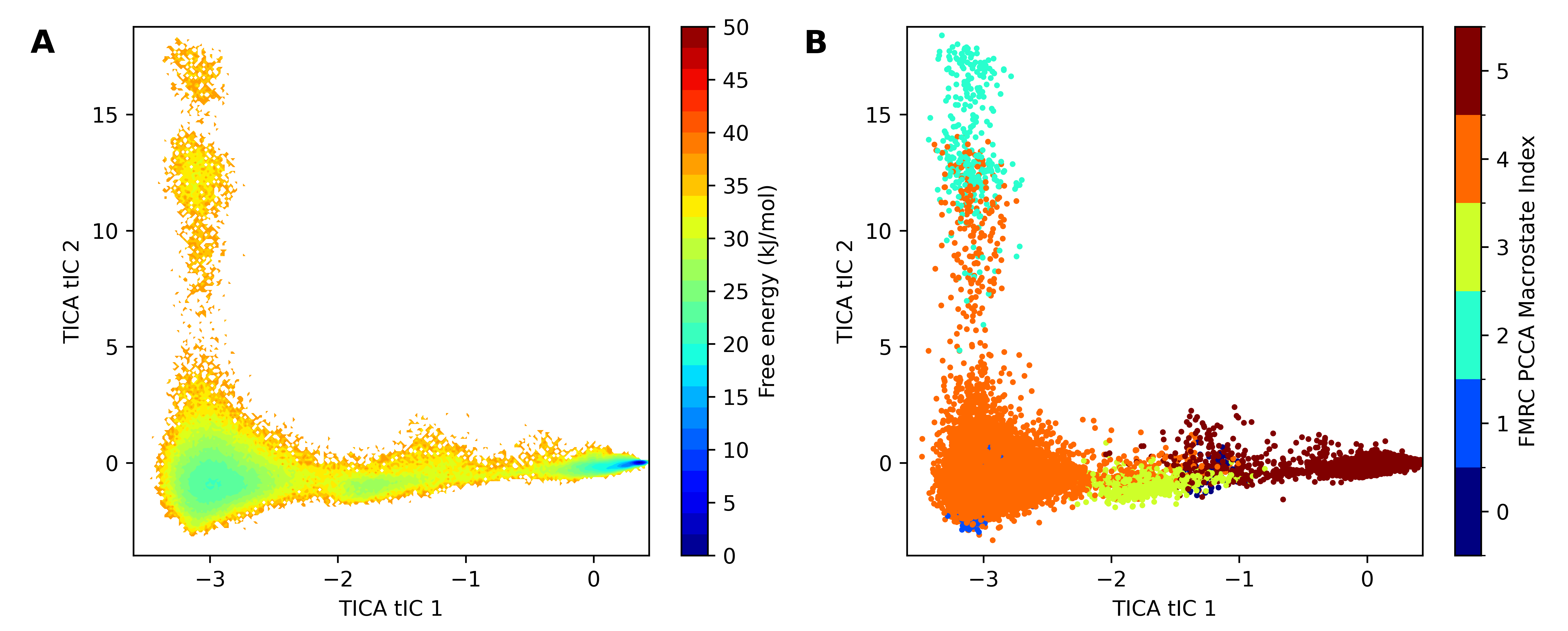}
    \caption{(A) The NTL-9 2D FES projection and (B) the NTL-9 PCCA+ macrostate assignment (learned from the best MSM constructed in the $\mathbf r^{FMRC}$ space) projection on the $\mathbf r^{TICA}_{2D}$ space where the MSM was constructed.}
    \label{fig:S29}
\end{figure}

\clearpage
\begin{figure}[htbp]
    \centering
    \includegraphics[scale=0.1]{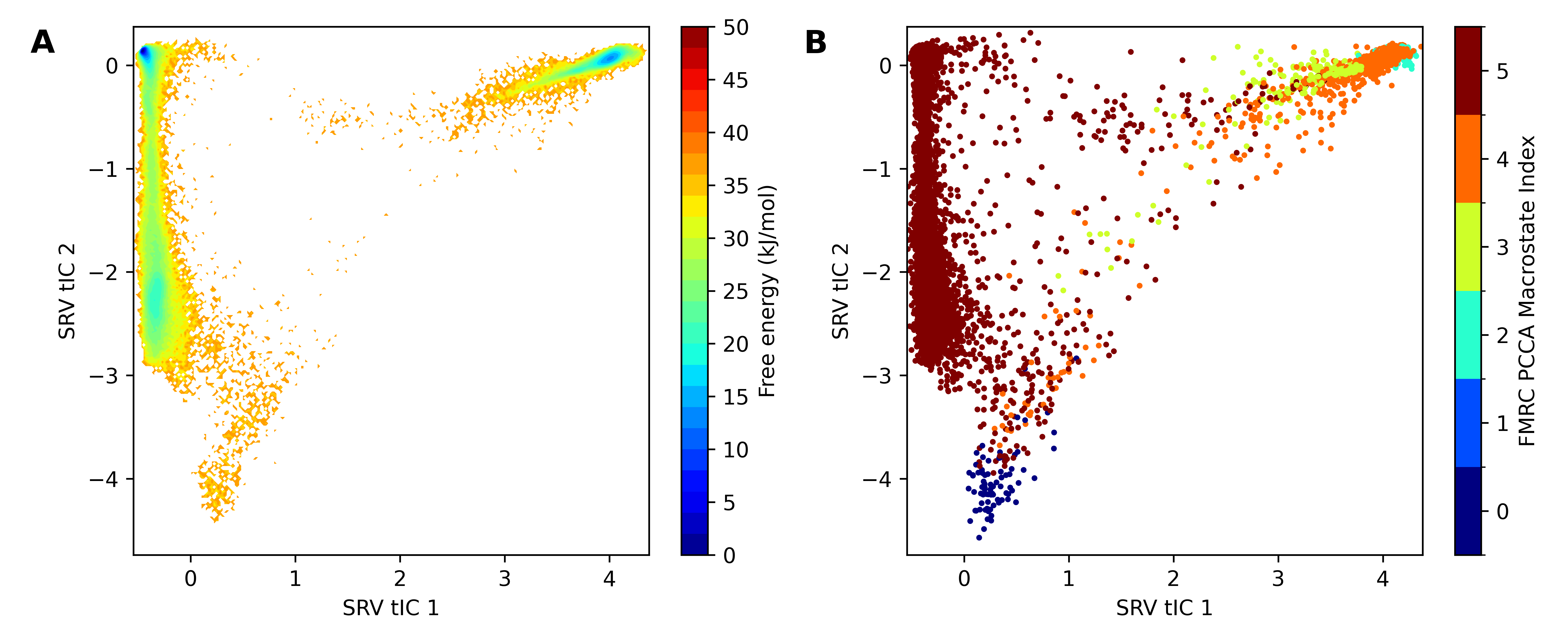}
    \caption{(A) The NTL-9 2D FES projection and (B) the NTL-9 PCCA+ macrostate assignment (learned from the best MSM constructed in the $\mathbf r^{FMRC}$ space) projection on the $\mathbf r^{SRV}$ space where the MSM was constructed.}
    \label{fig:S30}
\end{figure}

\clearpage
\begin{figure}[htbp]
    \centering
    \includegraphics[scale=0.1]{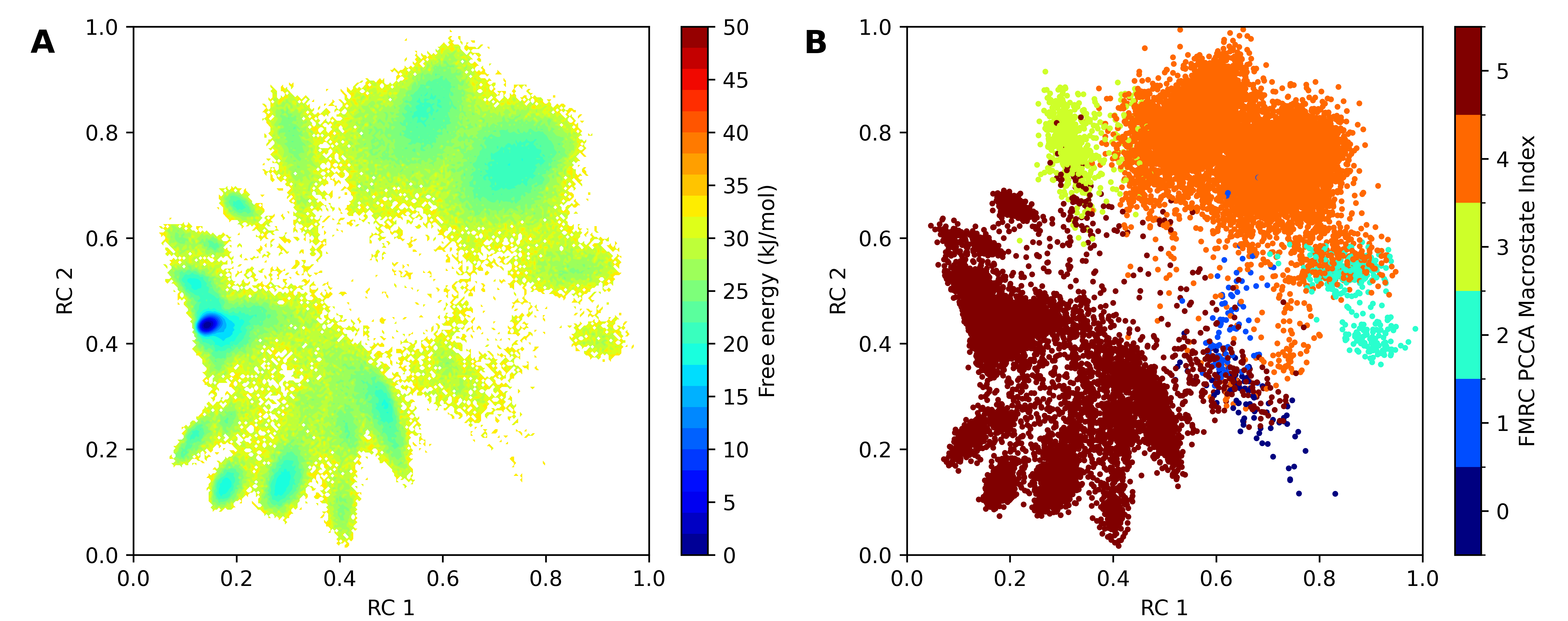}
    \caption{(A) The NTL-9 2D FES projection and (B) the NTL-9 PCCA+ macrostate assignment (learned from the best MSM constructed in the $\mathbf r^{FMRC}$ space) projection on the normalized $\mathbf r^{SPIB}$ space where the MSM was constructed.}
    \label{fig:S31}
\end{figure}

\clearpage
\begin{figure}[htbp]
    \centering
    \includegraphics[scale=0.75]{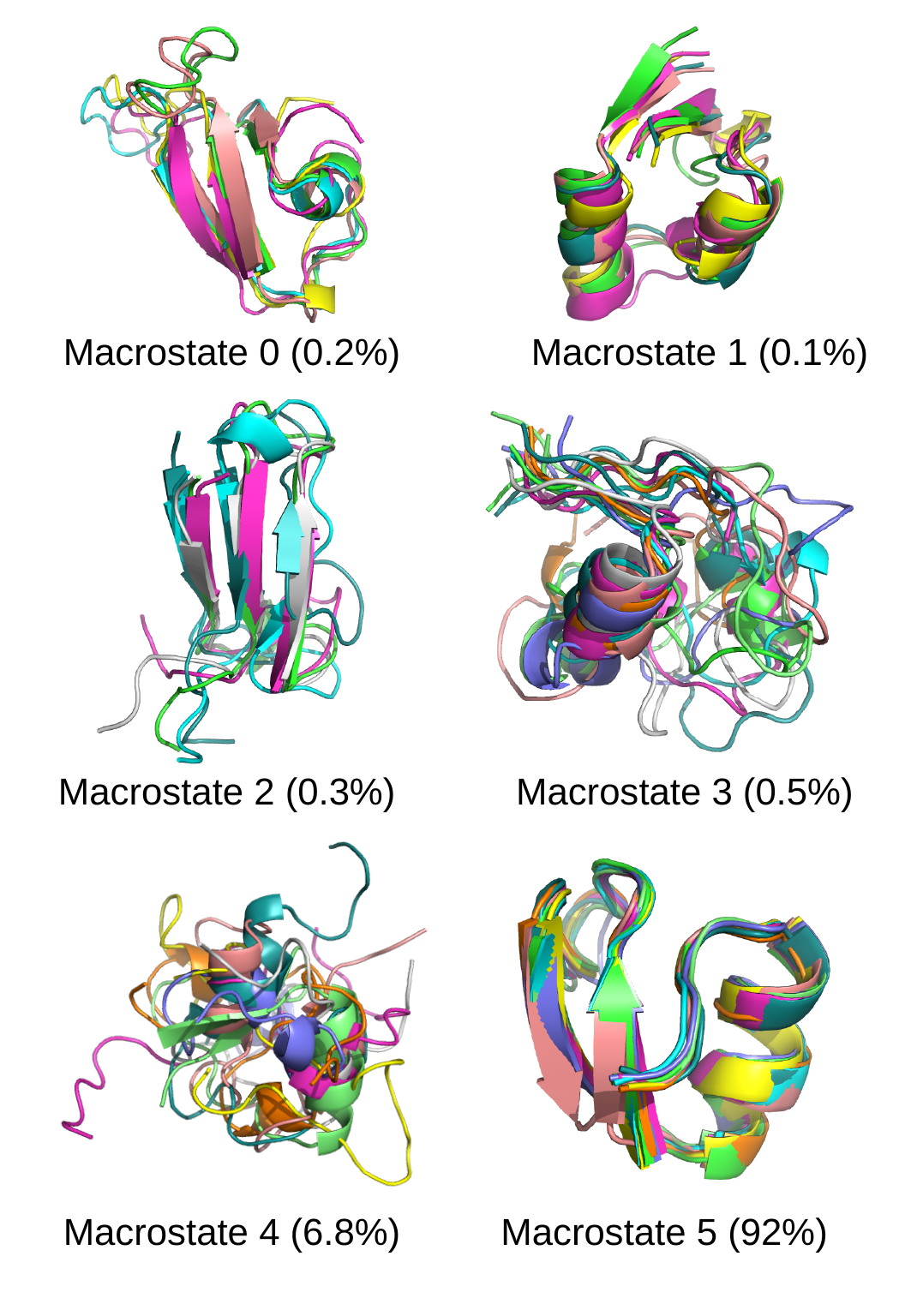}
    \caption{Representative conformations from each macrostate clustered by the PCCA+ algorithm for NTL-9. Macrostate population at equilibrium estimated from the PCCA+ coarse-grained transition matrix is shown below the conformational ensemble.}
    \label{fig:S32}
\end{figure}

\clearpage
\begin{figure}[htbp]
    \centering
    \includegraphics[scale=0.6]{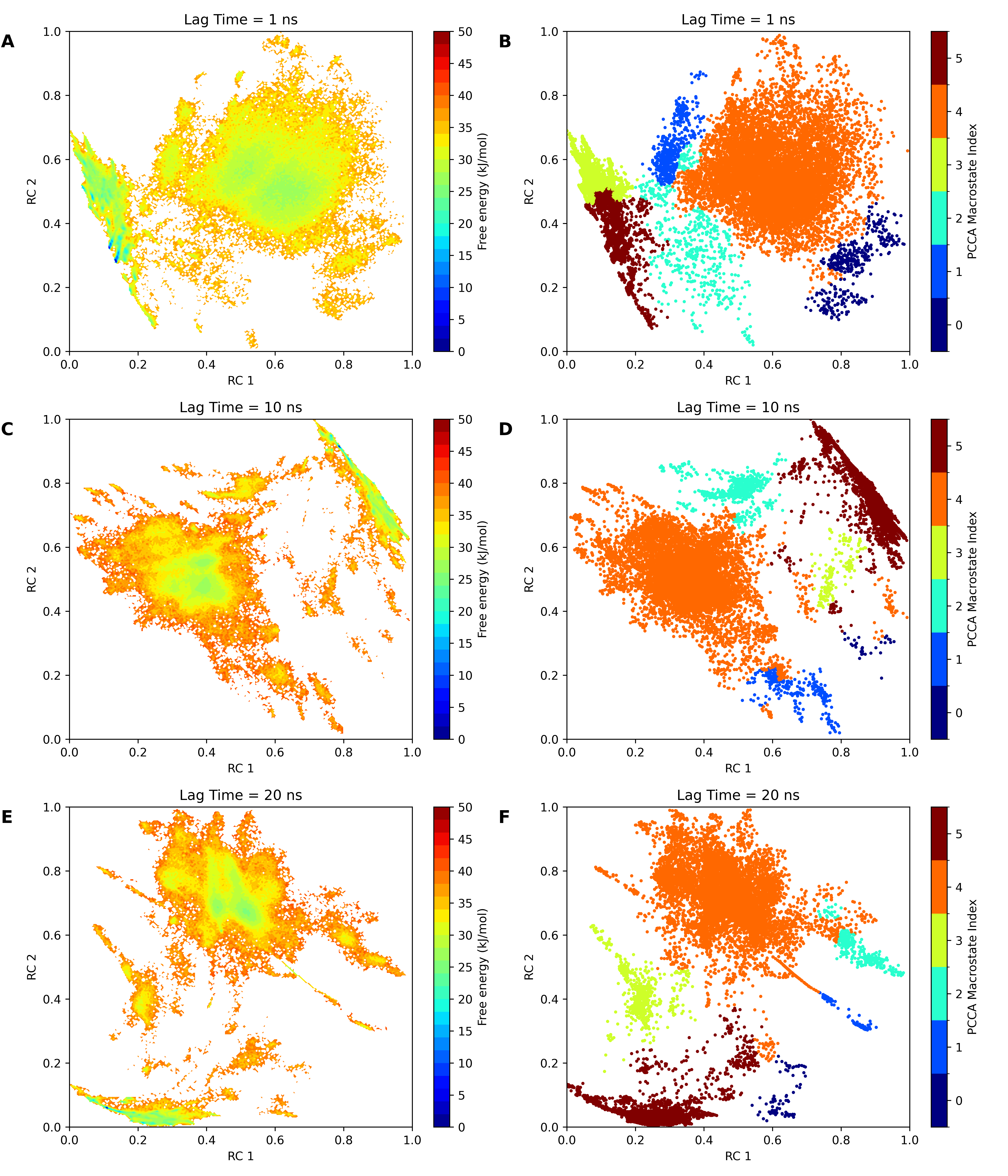}
    \caption{The NTL-9 2D FES projection and the NTL-9 PCCA+ macrostate assignment projection on the normalized $\mathbf r^{FMRC}$ space learned at $\tau=1$ ns (A,B), $\tau=10$ ns (C,D) and $\tau=20$ ns (E,F, same as main text Figure 9). Notice the macrostate assignment indices may not correspond to the same macrostate across plots for different $\tau$.}
    \label{fig:S33}
\end{figure}

\clearpage
\begin{figure}[htbp]
    \centering
    \includegraphics[scale=0.6]{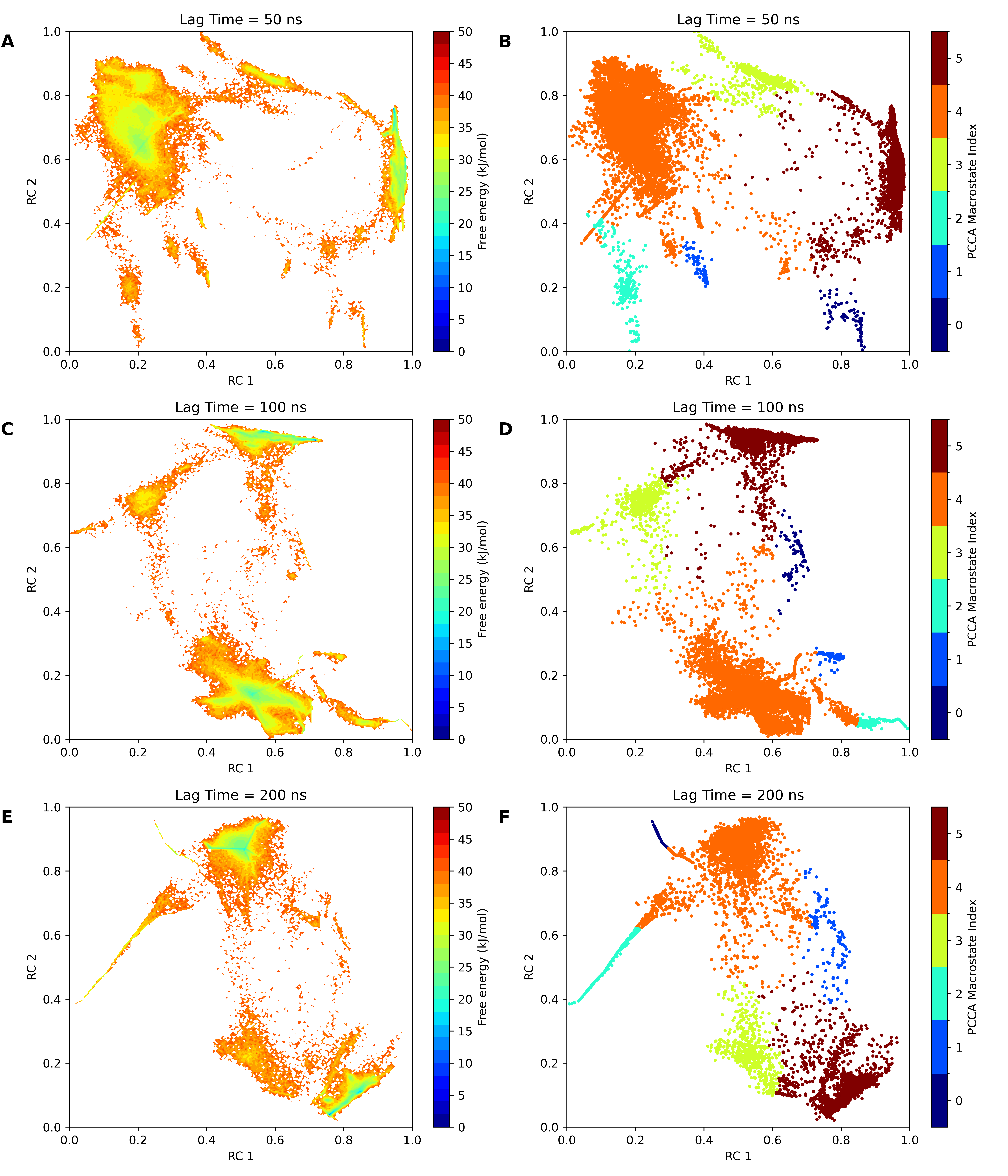}
    \caption{The NTL-9 2D FES projection and the NTL-9 PCCA+ macrostate assignment projection on the normalized $\mathbf r^{FMRC}$ space learned at $\tau=50$ ns (A,B), $\tau=100$ ns (C,D) and $\tau=200$ ns (E,F). Notice the macrostate assignment indices may not correspond to the same macrostate across plots for different $\tau$.}
    \label{fig:S34}
\end{figure}

\clearpage
\begin{figure}[htbp]
    \centering
    \includegraphics[scale=0.7]{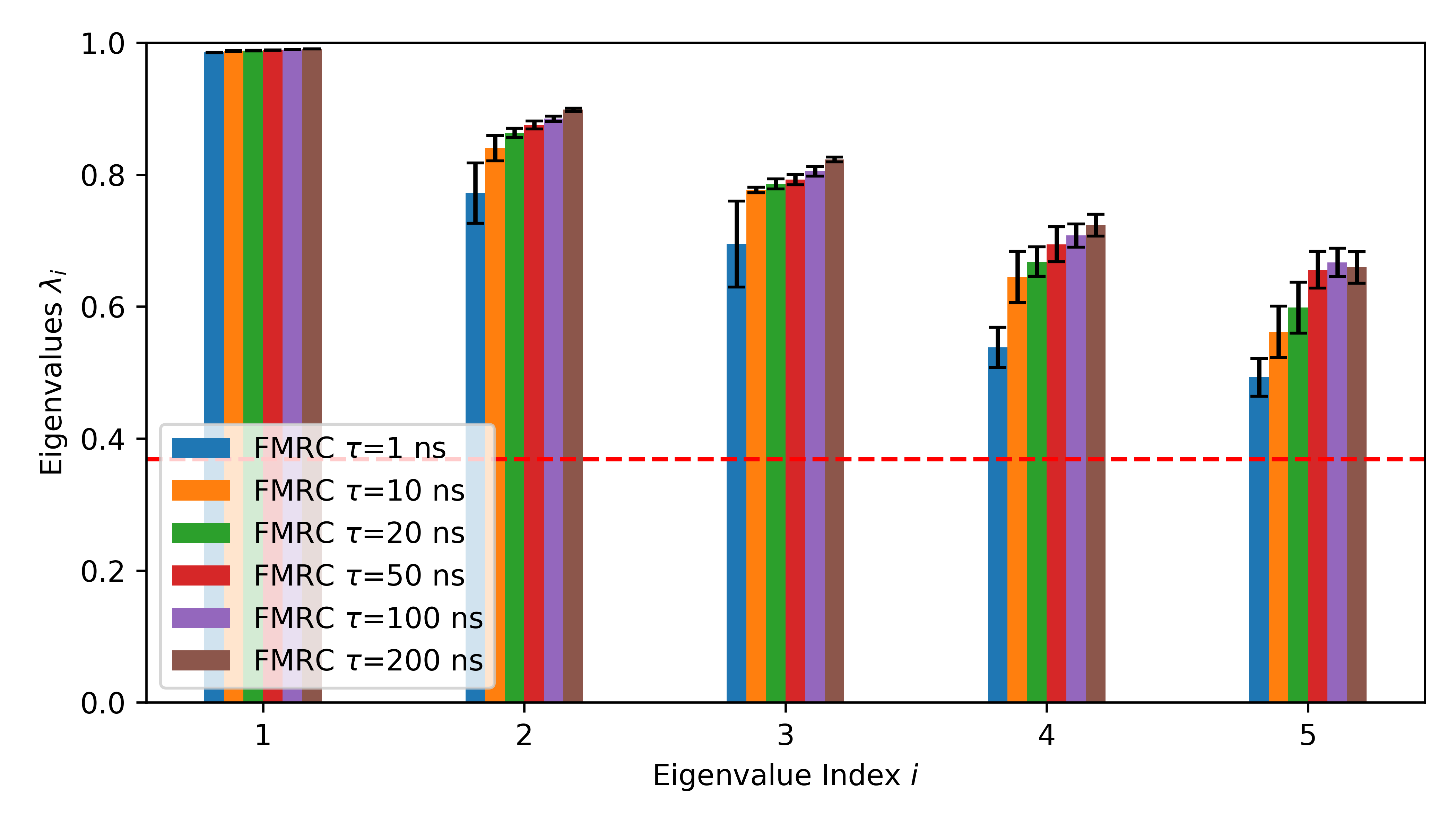}
    \caption{The comparison of $\hat \lambda_i^{MSM}$ of MSMs constructed in $\mathbf r^{FMRC}$ learned at different $\tau$ for NTL-9. A red strided line at $\lambda_i = 0.369$ has been drawn to denote a cutoff for the corresponding timescales lower than the $\tau^{MSM}$ for MSM construction. This indicates that the constructed MSM has failed to identify this slow process.}
    \label{fig:S35}
\end{figure}

\clearpage
\begin{figure}[htbp]
    \centering
    \includegraphics[scale=0.7]{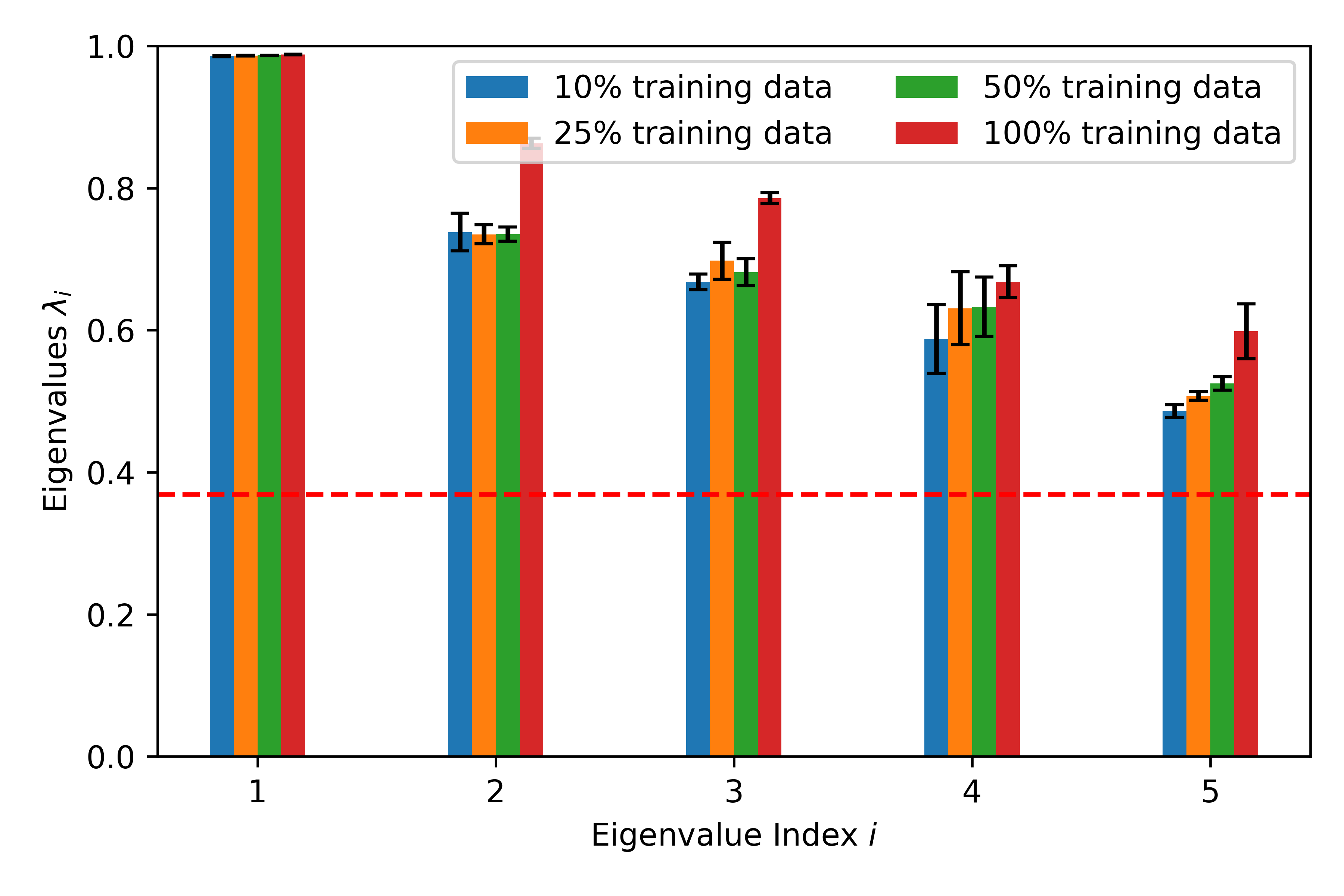}
    \caption{The comparison of $\hat \lambda_i^{MSM}$ of MSMs constructed in $\mathbf r^{FMRC}$ learned from a different amount of training data for NTL-9. A red strided line at $\lambda_i = 0.369$ has been drawn to denote a cutoff for the corresponding timescales lower than the $\tau^{MSM}$ for MSM construction. This indicates that the constructed MSM has failed to identify this slow process.}
    \label{fig:S36}
\end{figure}

\clearpage
\begin{figure}[htbp]
    \centering
    \includegraphics[scale=0.7]{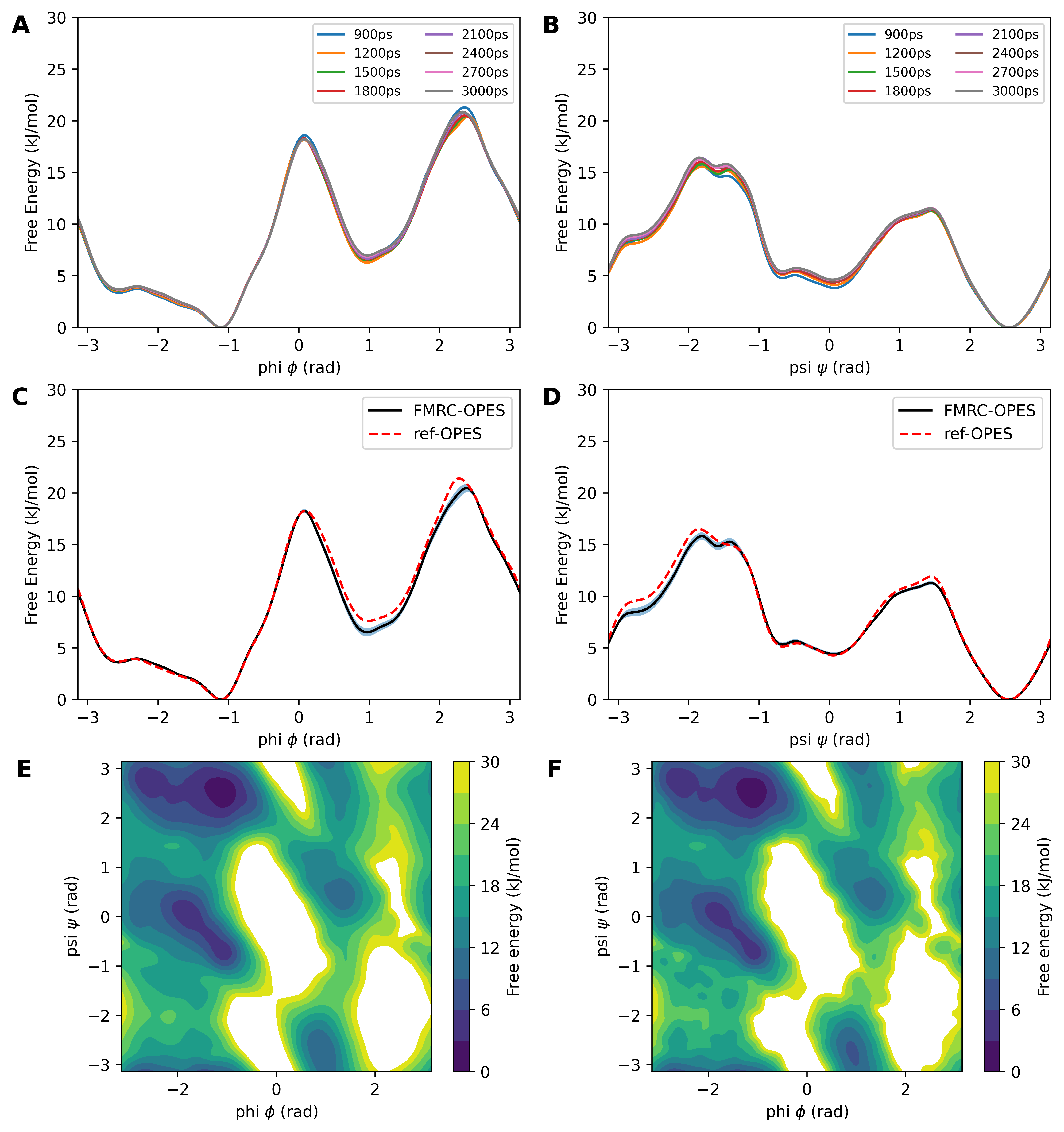}
    \caption{(A,B) Time evolution of the reweighted FES estimates along $\phi$ (A) and $\psi$ (B) from the Ala2 multiple-walkers OPES simulations. Notice that the simulation times shown in the legend represents the simulation time for each walker. (C,D) 1D FES estimates of Ala2 along $\phi$ (C), $\psi$ (D) from 16*3 ns multiple-walkers OPES simulation using the $\bold r^{FMRC}$ as biasing RC. A reference FES along $\phi$ and $\psi$ from a 250 ns OPES simulation biasing $\phi$ and $\psi$ is also shown in red dashed line. The error bars obtained from block analysis are shown as blue shades. (E) The reference $\phi$-$\psi$ 2D FES calculated from 250 ns OPES simulations using $\phi$ and $\psi$ as biasing RC. (F) The $\phi$-$\psi$ 2D FES calculated from 16*3ns OPES simulations using $\bold r^{FMRC}$ as biasing RC.}
    \label{fig:S37}
\end{figure}

\clearpage
\bibliography{reference}